\documentclass[10pt,twocolumn,letterpaper]{article}

\usepackage{3dv}

\usepackage{tikz}
\usetikzlibrary{positioning}

\usepackage{times}
\usepackage{epsfig}
\usepackage{graphicx}
\usepackage{amsmath}
\usepackage{amssymb}
\usepackage{xcolor}
\usepackage{tabularx}
\usepackage{booktabs}
\usepackage{cuted}
\usepackage{capt-of}
\usepackage{subcaption}
\usepackage{makecell}
\usepackage{authblk}

\newcommand\ignore[1]{}

\newcommand{\norm}[1]{\left\lVert#1\right\rVert}

\def \path{\bp C}

\usepackage[pagebackref=true,breaklinks=true,colorlinks,bookmarks=false]{hyperref}

\threedvfinalcopy 

\def\threedvPaperID{75} 

\ifthreedvfinal\fi
\begin{document}

\title{GO-Surf: Neural Feature Grid Optimization for \\Fast, High-Fidelity RGB-D Surface Reconstruction \vspace{-0.65cm}}

\author[*]{Jingwen Wang}
\author[*]{Tymoteusz Bleja}
\author[ ]{Lourdes Agapito}
\affil[ ]{Department of Computer Science, University College London}

\twocolumn[{%
    \renewcommand\twocolumn[1][]{#1}%
    \maketitle
    \centering
    \vspace{-0.8cm}
    \includegraphics[width=0.96\linewidth]{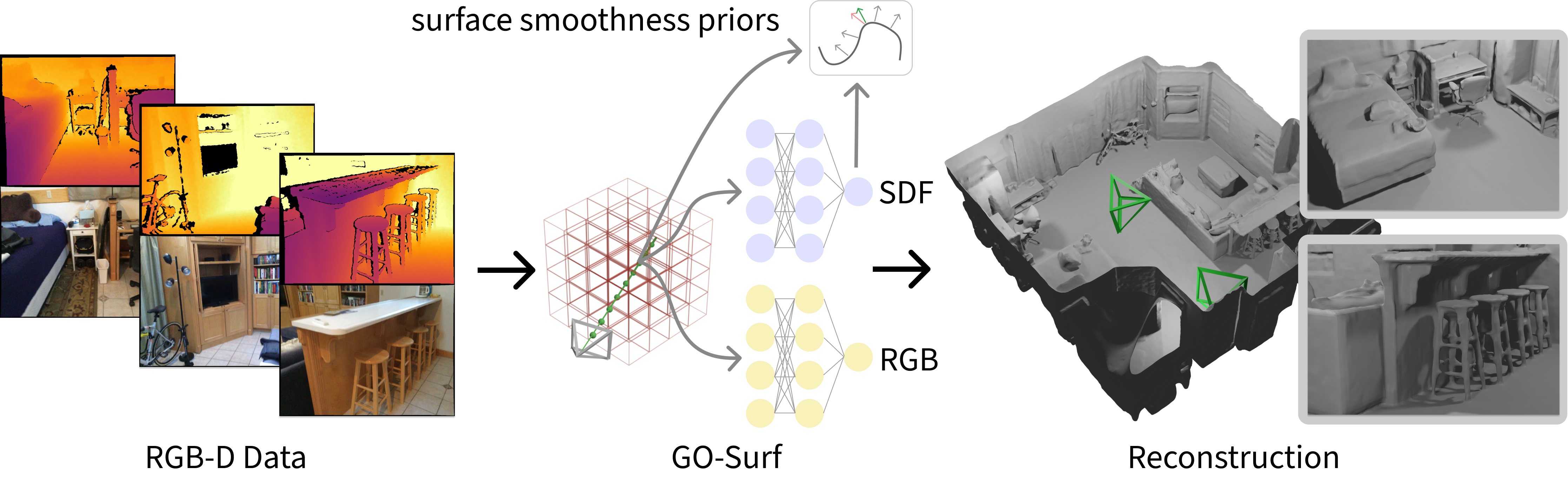}
\vspace{-3mm}
\captionof{figure}{Given an input RGB-D sequence, GO-Surf obtains a high quality 3D surface reconstruction by direct optimization of a multi-resolution feature grid and signed distance value and colour prediction. We formulate a new smoothness prior on the signed distance values that leads to improved hole filling and smoothness properties, while preserving details. Our optimization is $\times 60$ times faster than MLP-based methods. \label{fig:teaser}}
    \vspace{0.4cm}
}]

\maketitle
\thispagestyle{empty}
\renewcommand{\paragraph}[1]{\medskip\noindent\textbf{#1}}

\renewcommand*{\thefootnote}{\fnsymbol{footnote}}
\footnotetext{* The first two authors contributed equally.}
\renewcommand*{\thefootnote}{\arabic{footnote}}
\setcounter{footnote}{0}

\begin{abstract}
    \vspace{-0.8em}
    We present GO-Surf, a direct feature grid optimization method for accurate and fast surface reconstruction from RGB-D sequences. We model the underlying scene with a learned hierarchical feature voxel grid that encapsulates multi-level geometric and appearance local information. Feature vectors are directly optimized such that after being tri-linearly interpolated, decoded by two shallow MLPs into signed distance and radiance values, and rendered via volume rendering, the discrepancy between synthesized and observed RGB/depth values is minimized. Our supervision signals ---  RGB, depth and approximate SDF  --- can be obtained directly from input images without any need for fusion or post-processing. We formulate a novel SDF gradient regularization term that encourages surface smoothness and hole filling while maintaining high frequency details. GO-Surf can optimize sequences of $1$-$2$K frames in $15$-$45$ minutes, 
    a speedup of $\times60$ over NeuralRGB-D~\cite{azinovic2022neural}, the most related approach based on an MLP representation, while maintaining on par performance on standard benchmarks. Project page: \url{https://jingwenwang95.github.io/go_surf}.
\end{abstract}

\section{Introduction}
Recent years have seen impressive progress in learning-based methods for 3D scene modelling from video sequences, with a strong focus on coordinate-based scene representations~\cite{Park_2019_CVPR,occnet} and the application to highly photo-realistic novel view synthesis~\cite{srns,mildenhall2020nerf}. NeRF~\cite{mildenhall2020nerf} and its variants represent the scene in the weights of a multi-layer perceptron (MLP) that is trained to predict the volume density and colour of any 3D point location. Combined with classic differentiable volume renderering, NeRF learns to synthesize the input images and can generalize to render nearby unseen views. 

Although MLPs have shown an extraordinary ability to represent radiance fields of complex scenes with a low memory footprint, this comes at the cost of very long training and inference times. Direct optimization of trilinearly interpolated features on multi-resolution feature grids has become a powerful alternative representation due to its fast convergence that leads to orders of magnitude faster training and inference times~\cite{muller2022instant,karnewar2022relu,yu2021plenoxels,Zhu2022CVPR_niceslam,chen2022tensorf,sun2021direct}. Their higher memory requirement has also been addressed by using sparse feature grids, such as octrees~\cite{yu2021plenoxels}, or multi-resolution hash tables~\cite{muller2022instant}.

An added limitation of NeRF~\cite{mildenhall2020nerf} is its choice of geometric primitive: the volume density representation does not allow for accurate 3D surface extraction and can lead to blurry renderings or 3D reconstruction artefacts because of the limited capacity of the network to infer the exact surface location. A number of methods have addressed this~\cite{wang2021neus,niemeyer2020differentiable,yariv2020multiview} and proposed differentiable surface rendering pipelines that can be used in conjunction with neural implicit surface representations, thus avoiding representing the scene geometry via density.  
However, surface representations have mainly been proposed in the context of slow coordinate-based networks, while existing grid-based approaches struggle to achieve the level of smoothness comparable to MLP-based methods \cite{muller2022instant}.
Critically though, most AR/VR and robotics applications require both accurate and fast 3D surface reconstruction, therefore both limitations should be tackled for high fidelity surface inference at interactive runtimes. 

GO-Surf combines, for the first time, learnable feature volumes with SDF surface-based representations and rendering to improve both speed and accuracy. To further help overcome these limitations we  adopt the advantages of using depth measurements from consumer-level RGB-D cameras, which have become highly accessible, inexpensive, and are now found on many mobile-phones. Other approaches have recently also adopted RGB-D inputs in the context of neural scene representations~\cite{Sucar:etal:ICCV2021,Zhu2022CVPR_niceslam,Ortiz:etal:iSDF2022,azinovic2022neural} and use depth images to supervise scene reconstruction. Unlike GO-Surf, iMAP~\cite{Sucar:etal:ICCV2021} models volume density, while NICE-SLAM~\cite{Zhu2022CVPR_niceslam} uses occupancy, both failing to predict high-fidelity, accurate surface reconstructions. Although  NeuralRGB-D~\cite{azinovic2022neural} adopts an SDF representation, such as GO-Surf, it uses an MLP to encode the scene, leading to very long training times. Our experimental evaluation shows that GO-Surf achieves a speedup of $\times 60$ through use of learnable feature volumes. Although iSDF~\cite{Ortiz:etal:iSDF2022} also uses an SDF representation, we show that our novel SDF regularization term results in more accurate and higher fidelity reconstructions. 

\noindent{\bf Contribution:} GO-Surf brings learnable feature grids into the context of SDF  reconstruction from RGB-D sequences to achieve both: (\emph{i}) fast optimization at interactive runtimes, and (\emph{ii}) highly accurate surface reconstruction. We also apply for the first time Eikonal and smoothness regularisation terms in the context of voxel grids.

\begin{figure*}[t]
\centering
\includegraphics[width=0.95\linewidth]{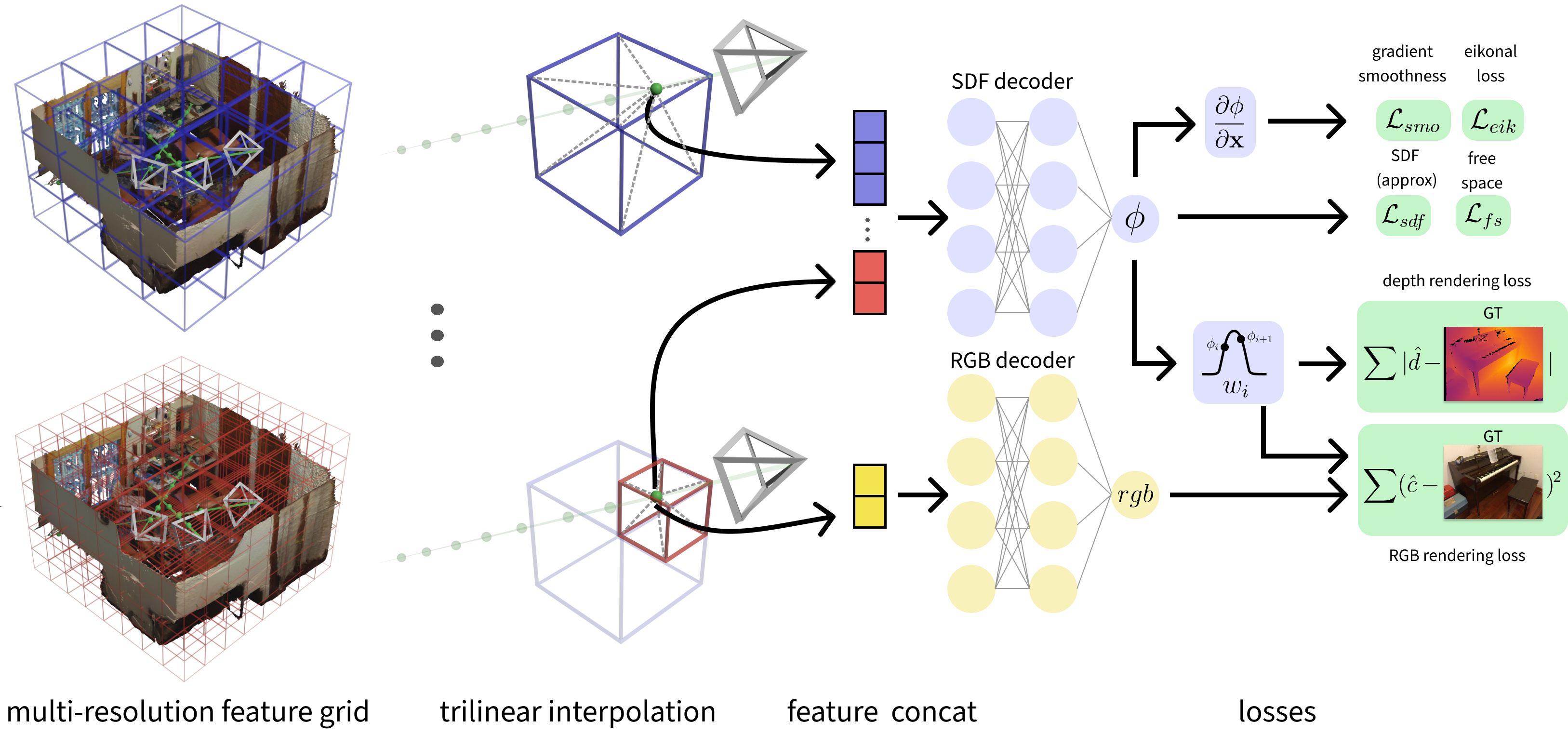}

\captionof{figure}{GO-Surf uses multi-level feature grids and two shallow MLP decoders. Given a sample point along a ray, each grid is queried via tri-linear interpolation. Multi-level features are concatenated and decoded into SDF, and used to compute the sample weight. Color is decoded separately from the finest grid. Loss terms are applied to SDF values, and rendered depth and color. The gradient of the SDF is calculated at each query point and used for Eikonal and smoothness regularization.\label{fig:overview}}
\vspace{-4mm}
\end{figure*}

\section{Related Work}
\noindent{\bf Classic RGB-D Reconstruction} Although this review  focuses on learning based approaches to RGB-D reconstruction we cite KinectFusion~\cite{newcombe2011kinectfusion}, VoxelHashing~\cite{niessner2013real} and BundleFusion~\cite{dai2017bundlefusion} as the best representatives of classic methods, with the last one being one of our baselines. 

\noindent{\bf Coordinate-based Networks}
Pioneered by Scene Representation Networks~\cite{srns}, and popularized by NeRF~\cite{mildenhall2020nerf}, coordinate-based methods encode the scene within the weights of a fully connected neural network that take as input a 3D location and predict its geometry and appearance information in the form of density and  radiance~\cite{mildenhall2020nerf,Sucar:etal:ICCV2021}, occupancy~\cite{occnet,chibane2020implicit}, signed distance fields~\cite{Park_2019_CVPR,icml2020_2086,azinovic2022neural,Ortiz:etal:iSDF2022}, colour~\cite{oechsle2019texture} or semantic labels~\cite{zhi2021place,kundu2022panoptic}. NeRF~\cite{mildenhall2020nerf} demonstrated photo-realistic novel view synthesis by combining an MLP to encode density and colour with classical volumetric rendering, only requiring a set of posed RGB images for training.  Using hybrid scene representations, NVSF~\cite{liu2020neural} and  Plenoctrees~\cite{yu2021plenoctrees} combine neural implicit fields with explicit sparse voxel structures to accelerate rendering.

Closely related to our work are coordinate-based networks trained with additional depth supervision for 3D reconstruction~\cite{Sucar:etal:ICCV2021,Ortiz:etal:iSDF2022,azinovic2022neural}. iMAP~\cite{Sucar:etal:ICCV2021} is the first to train an MLP in live, real-time operation with RGB-D inputs to learn a scene specific 3D model of occupancy and colour, using rendered depth and RGB as supervision signals, while also tracking camera pose. In iSDF~\cite{Ortiz:etal:iSDF2022} the focus is actual surface reconstruction. Taking as input a live stream of posed depth images, approximate signed distance values are predicted by minimising a loss on both the predicted signed distance and its spatial gradient. Both~\cite{Sucar:etal:ICCV2021,Ortiz:etal:iSDF2022} achieve live operation via active sampling and keyframe selection. Instead, Neural RGB-D~\cite{azinovic2022neural} operates in batch mode, taking as input a large set of RGB-D frames to train MLPs to map coordinate inputs to approximate SDF values and radiance via differentiable volumetric rendering. While they achieve extremely accurate reconstructions that preserve high frequency details, their training time is on the order of $10$ hours for typical scenes.
The huge advantage of these MLP-based representations is their compactness, given that they can be sampled continuously without increasing memory footprint, they either suffer from extremely long training times~\cite{azinovic2022neural} or sacrifice accuracy/details for online operation~\cite{Sucar:etal:ICCV2021,Ortiz:etal:iSDF2022}. 




\noindent{\bf Neural 3D Scene Reconstruction} 
A recent trend includes architectures trained end-to-end on multiple sequences to aggregate image features and perform fusion over multiple frames using a volumetric representation and decoders to predict signed distance~\cite{sun2021neuralrecon,murez2020atlas,bozic2021transformerfusion} or radiance~\cite{zhang2022nerfusion} fields. NeuralRecon~\cite{sun2021neuralrecon} reconstructs surfaces sequentially from video fragments as TSDF volumes by performing feature fusion from previous fragments via recurrent units. Atlas~\cite{murez2020atlas} aggregates image features over an entire sequence to predict a globally consistent TSDF volume and semantic labels with a 3DNN, while others~\cite{bozic2021transformerfusion,stier2021vortx} rely on  transformers for feature fusion. While the above methods require direct SDF supervision at training time, inference only requires posed RGB video. NeRFusion~\cite{zhang2022nerfusion} uses a 3D CNN for feature fusion but drops the need for 3D supervision by including a volume rendering module and can be trained only from RGB images.

\noindent{\bf Neural Implicit Surface Reconstruction}
Following a similar observation to ours --- that the volume density estimated by NeRF~\cite{mildenhall2020nerf} does not enable high-quality surface extraction --- a recent family of methods focus instead on neural surface representations and formulate compatible differentiable renderers. 
DVR~\cite{niemeyer2020differentiable} propose the first differentiable renderer for implicit shape and texture representations needing only multiview RGB images and object masks as supervision. IDR~\cite{yariv2020multiview} trains an end-to-end architecture with a learned neural renderer that approximates light reflected from the surface and can implicitly model a variety of lighting conditions and materials. NeuS~\cite{wang2021neus} instead formulates a new volume rendering method to train a bias-free neural SDF representation, while VolSDF~\cite{yariv2021volume} formulate a novel parameterization for volume density, both leading to more accurate surface reconstruction even without mask supervision. 
UniSurf~\cite{sun2021direct} unifies neural volume and surface rendering enabling both within the same model. 

\noindent{\bf Direct Grid Optimization}
Recent, mostly concurrent, work has also explored the use of directly optimizing neural features on an explicit volumetric grid. NGLoD~\cite{takikawa2021nglod} represents implicit surfaces using a sparse octree feature volume which adaptively fits shapes with multiple levels of detail using 3D supervision. Instant NGP~\cite{muller2022instant} proposes a multi-resolution hash table of trainable feature vectors achieving a speedup of various orders of magnitude, demonstrating instant training on a variety of tasks. DirectVoxGO~\cite{sun2021direct} also improves NeRF's training time by two orders of magnitude by adopting an explicit density voxel-grid with post-activation interpolation --- that enables to model sharp boundaries --- and a feature voxel-grid with a shallow MLP for view-dependent appearance. Plenoxels~\cite{yu2021plenoxels} and ReLU Fields~\cite{karnewar2022relu} take this idea further dropping the reliance on any neural networks. While Plenoxels~\cite{yu2021plenoxels} relies on a spherical harmonics basis, ReLU Fields~\cite{karnewar2022relu} apply post-activation interpolation. TensoRF~\cite{chen2022tensorf} tackles the issue of memory footprint by modelling the volume field as a 4D tensor that is factorized into multiple compact low-rank tensor components. Finally, NiceSLAM~\cite{Zhu2022CVPR_niceslam} is an RGB-D SLAM system that encodes the scene as a multi-level feature grid that is optimized at run-time using pre-trained geometric priors, enabling detailed reconstruction on large indoor scenes.


\section{Method}

We formulate scene reconstruction as a direct optimisation problem, representing scene geometry and color using multi-resolution feature grids, which are decoded into SDF and RGB values with two MLP decoders shared across grids. Given a sequence of RGB-D images $\{I_t, D_t\}$, and initial camera poses $\{\xi_t\} \in \mathbb{SE}(3)$ (provided by SLAM or SfM) we jointly optimize feature volumes, decoders and camera poses.

\subsection{Hybrid Scene Representation\label{sec:scene_representation}}

Our hybrid scene representation combines multi-level feature grids and shared MLPs for geometry and color prediction. We encode the scene geometry into a four-level feature grid $\mathcal{V}_{\theta} = \{V^{l}\}$, where $l \in \{0, 1, 2, 3\}$  encode multi-level local detail from coarse to fine. Finer  features capture high-frequency detailed geometry while the coarser features encapsulate larger structures and are crucial for hole-filling. 

Given a scene point $\mathbf{x} \in \mathbb{R}^3$, its geometry feature is obtained by concatenating the trilinearly interpolated features at each level. The feature is decoded into an SDF value $\phi(\mathbf{x})$ via the geometry MLP $f_{\omega}(\cdot)$:
\begin{align} \label{SDF_MLP}
    \mathcal{V}_{\theta}(\mathbf{x}) &= [V^0(\mathbf{x}), V^1(\mathbf{x}), V^2(\mathbf{x}), V^3(\mathbf{x})] \\
    \phi(\mathbf{x}) &= f_{\omega}(\mathcal{V}_{\theta}(\mathbf{x}))
\end{align}
We encode color information only at the finest level, as~\cite{Zhu2022CVPR_niceslam}, using a separate feature grid $\mathcal{W}_{\beta}$ and decoder $g_{\gamma}(\cdot)$:
\begin{equation} \label{COLOR_MLP}
    \mathbf{c} = g_{\gamma}(\mathcal{W}_{\beta}(\mathbf{x}), \mathbf{d})
\end{equation}
where $\mathbf{d}$ is the viewing direction. Here $\theta$, $\beta$, $\omega$, $\gamma$ represent the optimizable parameters, i.e. geometry and color local features and decoder parameters.

\paragraph{Architecture Details.} The multi-resolution feature grids, use geometry features of dimension $h_g = 4$ per-level, giving a final geometry feature vector of dimension $16$. Color features have dimension $h_c = 6$. For both geometry and color decoders we use light-weight MLPs with $2$ hidden layers, $32$ neurons each.

\subsection{Depth and Color Rendering\label{method:rendering}}

Inspired by the recent success of volume rendering~\cite{mildenhall2020nerf}, we render depth and color by integrating predicted colors and sampled depth along camera rays.
Specifically, for each back-projected ray parameterised by camera centre $\mathbf{o}$ and ray direction $\mathbf{r}$ we sample $N$ points $\mathbf{x}_i = \mathbf{o} + d_i\mathbf{r}, i \in \{1, \dots, N\}$ and take the expected values of predicted colors $\mathbf{c}_i$ and sampled depth $d_i$:%
\begin{equation} \label{volume_rendering}
    \hat{\mathbf{c}} = \sum_{i=1}^{N}w_i \mathbf{c}_i, \quad \hat{d} = \sum_{i=1}^{N}w_i d_i
\end{equation}
where $\{w_i\}$ are unbiased and occlusion-aware weights \cite{wang2021neus} given by  $w_i = T_i\alpha_i$, where $T_i = \prod_{j=1}^{i-1}(1-\alpha_j)$ represents the \textit{accumulated transmittance} at point $\mathbf{x}_i$, and $\alpha_i$ is the \textit{opacity value} computed by:
\begin{equation} \label{volume_rendering_alpha}
    \alpha_i = \max \bigg(\frac{\sigma_s(\phi(\mathbf{x}_{i})) - \sigma_s(\phi(\mathbf{x}_{i+1}))}{\sigma_s(\phi(\mathbf{x}_{i}))}, 0 \bigg)
\end{equation}
where $\sigma_s(x) = (1 + e^{-sx})^{-1}$ is a Sigmoid function modulated by a learnable parameter $s$ which controls the smoothness of the transition at the surface. The RGB and depth per-pixel rendering losses are:
\begin{equation} \label{rendering terms}
    \ell_{rgb} = \lVert I[u, v] - \hat{\mathbf{c}} \rVert, \quad \ell_{d} = \lvert D[u, v] - \hat{d} \rvert
\end{equation}
To obtain sampling points along the ray we perform 3 steps of importance sampling, as~\cite{wang2021neus}: starting from $N_c=96$ coarse samples we iteratively add $12$ samples each time based on weights computed with previously sampled points. 
\subsection{Approximate SDF Supervision}
Similar to \cite{azinovic2022neural, Ortiz:etal:iSDF2022}, for each sampled point along the ray we also approximate ground-truth SDF supervision based on the distance to observed depth values along the ray direction: $b(\mathbf{x}) = D[u, v] - d$.
With this bound we have $\lvert \phi(\mathbf{x}) \rvert \leq \lvert b(\mathbf{x}) \rvert, \forall \mathbf{x}$, which is expected to be tighter near the surface, so for near-surface points ($|D[u, v] - d| <= t$) we apply the following SDF loss:
\begin{equation} \label{near_loss}
    \ell_{sdf}(\mathbf{x}) = |\phi(\mathbf{x}) - b(\mathbf{x})|
\end{equation}
%
The truncation threshold $t$ is a hyper-parameter (we set $t = 16cm$). Unlike~\cite{azinovic2022neural} which models TSDF for points far from the surface ($|D[u, v] - d| > t$), encouraging the MLP to predict a fixed truncation value; we apply a relaxed loss to encourage free space prediction as~\cite{Ortiz:etal:iSDF2022}:
\begin{equation} \label{far_loss}
    \ell_{fs}(\mathbf{x}) = \max \Big(0, e^{-\alpha \phi(\mathbf{x})} - 1, \phi(\mathbf{x}) - b(\mathbf{x})\Big)
\end{equation}
We apply no penalty if the SDF prediction is positive and smaller than the bound, a linear loss if it is larger, and exponential penalty ($\alpha=5$) if it is negative. Despite being approximated SDF values, these terms provide more direct supervision than the rendering terms in Sec.~\ref{method:rendering}. Empirically we observed that they add robustness to the optimisation.

\subsection{Grid-based SDF Regularisation}
\begin{figure}
    \centering
    \includegraphics[width=0.32\linewidth]{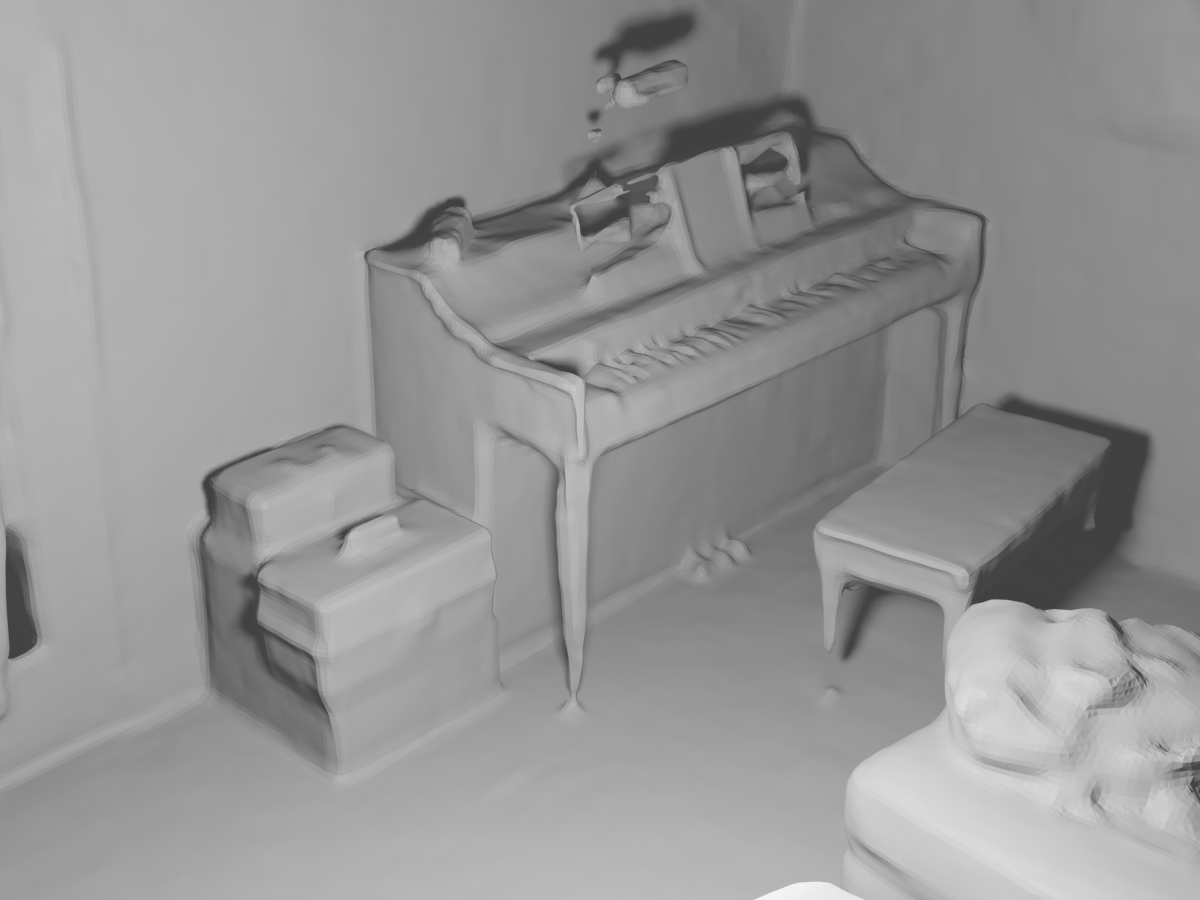}
    \includegraphics[width=0.32\linewidth]{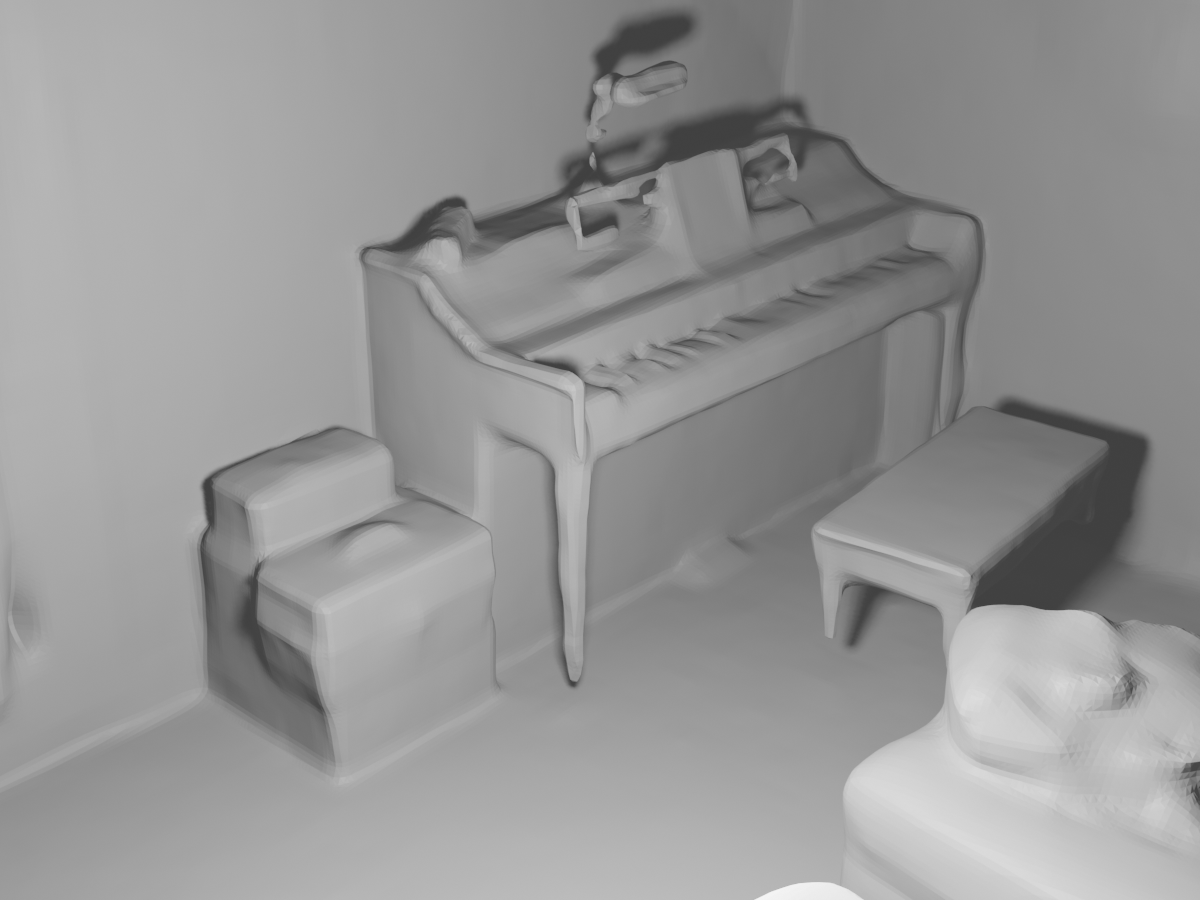}
    \includegraphics[width=0.32\linewidth]{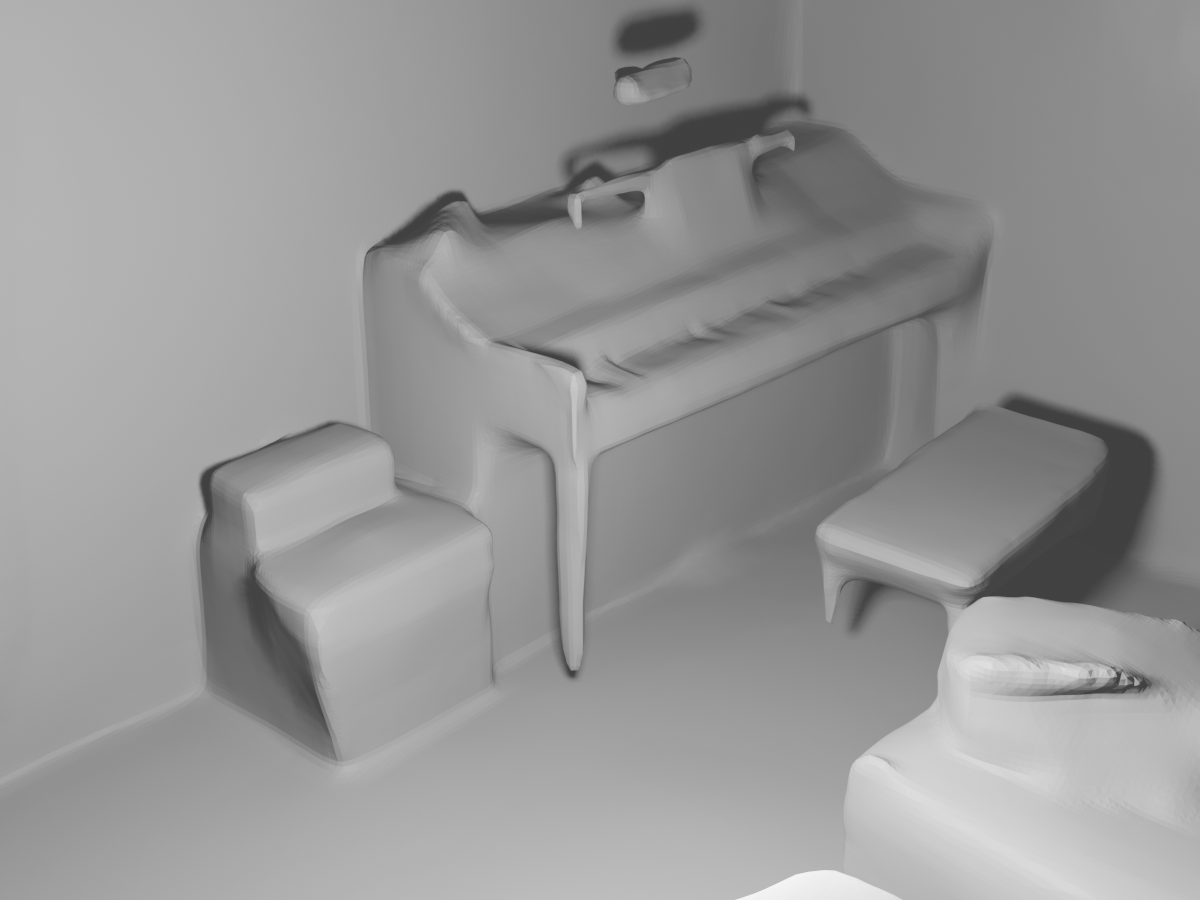}
    \caption{The distance between sampled points for gradient regularisation ($\delta_s$) controls surface smoothness. Left: $\delta_s=1\text{ mm}$, Middle: $\delta_s=4\text{ mm} $, Right: $\delta_s=2\text{ cm}$.}
    \vspace{-4mm}
    \label{fig:smoothness_std}
\end{figure}

Directly optimizing the feature volume from RGB-D observations is not sufficiently  constrained and can lead to  artifacts and invalid predictions. We employ two regularization terms on the SDF values: Eikonal term $\ell_{eik}$ and smoothness term $\ell_{smooth}$. Eikonal regularisation has been used in~\cite{icml2020_2086, wang2021neus, Ortiz:etal:iSDF2022} to encourage valid SDF values, especially in areas without direct supervision. Specifically, at any query point, the gradient of the SDF value w.r.t. 3D query coordinates should have unit length:
\begin{equation} \label{eikonal_loss}
    \ell_{eik}(\mathbf{x}) = \big(1 - \norm{\nabla \phi(\mathbf{x})}\big)^2
\end{equation}
Intuitively, this term encourages the absolute value of SDF to increase uniformly as we move away from the surface. 
Unlike~\cite{azinovic2022neural, wang2021neus, Sucar:etal:ICCV2021, Zhu2022CVPR_niceslam} that use an MLP scene  representation with built-in smoothness priors, we must explicitly add a smoothness term. Our loss minimises the difference between the gradients of nearby points. 
\begin{equation} \label{smoothness_loss}
    \ell_{smooth}(\mathbf{x}) = \norm{\nabla \phi(\mathbf{x}) - \nabla \phi(\mathbf{x + \epsilon})}^2
\end{equation}
Here $\mathbf{x}$ is taken from near-surface points sampled randomly across the whole model, $\mathbf{\epsilon}$ is a small perturbation of length $\delta_s$ and random direction (we set $\delta_s$ between 4mm and 1mm, see Fig.\ref{fig:smoothness_std}). 

%
%

\noindent{\bf Customised Grid Sampler.} Adding the above terms to the final loss requires second order derivatives of the SDF to be computed. The SDF value is a function of 3D coordinates $\mathbf{x}$, feature vectors $\theta$, and geometry network parameters $\omega$:
\begin{equation}  \label{sdf_full}
    \phi(\mathbf{x}, \theta, \omega) = f_{\omega}(\mathcal{V}_{\theta}(\mathbf{x}))
\end{equation}
The gradient of the SDF w.r.t. 3D query coordinates is:
\begin{equation} \label{sdf_grad}
    \frac{\partial \phi}{\partial \mathbf{x}} = \frac{\partial f_{\omega}(\mathcal{V}_{\theta}(\mathbf{x}))}{\partial \mathcal{V}_{\theta}(\mathbf{x})} \frac{\partial \mathcal{V}_{\theta}(\mathbf{x})}{\partial \mathbf{x}}
\end{equation}
To regularise this gradient we need to compute the first three rows (since $\mathbf{x}$ is 3-dimensional) of the Hessian of the SDF (Eq.~\ref{sdf_full}). In practice, Pytorch's automatic differentiation package supports double back-propagation through an MLP in the computation graph, but we still need to compute second order derivatives of the trilinear interpolation w.r.t. features $\theta$ and query coordinates $\mathbf{x}$. We implement a drop-in replacement of Pytorch's \texttt{grid\_sampler} that supports double back-propagation. Specifically, we reuse Pytorch's implementation for forward and backward passes, and add a new CUDA kernel for the second backward pass. Our implementation will be open-sourced. For derivation of the second order derivatives of trilinear interpolation, please refer to the  supplementary material.

\subsection{Optimization}

To optimize our scene representation described in Sec~\ref{sec:scene_representation}, at each iteration we sample a batch of $R$ rays from all pixels across all images (including pixels that have missing depth values). Our global objective function $\mathcal{L}(\mathcal{P})$ is: 
\begin{eqnarray}
    \mathcal{L}(\mathcal{P}) = \lambda_{rgb}\mathcal{L}_{rgb} + \lambda_{d}\mathcal{L}_{d} + \lambda_{sdf}\mathcal{L}_{sdf} + \nonumber \\ 
     \lambda_{fs}\mathcal{L}_{fs} + \lambda_{eik}\mathcal{L}_{eik} + \lambda_{smooth}\mathcal{L}_{smooth}
\end{eqnarray}
where $\mathcal{P} = \{\theta, \omega, \beta, \gamma, \{\xi_t\}\}$ is the list of parameters being optimised, including features, decoders and camera poses. $\mathcal{L}_{rgb}$ and $\mathcal{L}_{d}$ measure the difference between observed pixel color/depth and rendered values. $\mathcal{L}_{rgb}$ is computed over all the sampled rays, while for $\mathcal{L}_d$ only rays with valid depth values $R_d$ are considered.
\begin{equation}
    \mathcal{L}_{rgb} = \frac{1}{M} \sum_{m=1}^{M} \ell_{rgb}^m, \quad \mathcal{L}_{d} = \frac{1}{|R_d|} \sum_{r\in R_d} \ell_{d}^m
\end{equation}
SDF and free space losses, $\mathcal{L}_{sdf}$ and $\mathcal{L}_{fs}$ are applied on two disjoint subsets $S_{tr}$ and $S_{fs}$ of sampled ray points:
\begin{eqnarray}
    \mathcal{L}_{sdf} = \frac{1}{M} \sum_{m=1}^{M} \frac{1}{|S_{tr}|} \sum_{s \in S_{tr}} \ell_{sdf}(\mathbf{x}_s) \\
    \mathcal{L}_{fs} = \frac{1}{M} \sum_{m=1}^{M} \frac{1}{|S_{fs}|} \sum_{s \in S_{fs}} \ell_{fs}(\mathbf{x}_s)
\end{eqnarray}
$\mathcal{L}_{eik}$ encourages points far from the surface to have valid SDF predictions and is applied on $S_{fs}$:
\begin{equation}
    \mathcal{L}_{eik} = \frac{1}{M} \sum_{m=1}^{M} \frac{1}{|S_{fs}|} \sum_{s \in S_{fs}} \ell_{eik}(\mathbf{x}_s)
\end{equation}
$\mathcal{L}_{smooth}$ encourages surface smoothness and is applied to randomly sampled near-surface points $S_g$ over the whole voxel grid:
\begin{equation}
    \mathcal{L}_{smooth} = \frac{1}{|S_{g}|} \sum_{s \in S_{g}} \ell_{smooth}(\mathbf{x}_s)
\end{equation}

\paragraph{Geometric Initialisation.}
To improve convergence, we initialize our feature grid and geometry decoder such that the initial surface is a sphere~\cite{icml2020_2086} centred at volume origin and with a radius half the size of the smallest  dimension.

\begin{figure*}[tb]
    \centering
    \newcolumntype{Y}{>{\centering\arraybackslash}X}
    \begin{tabularx}{0.95\linewidth}{@{}Y@{\,}Y@{\,}Y@{\,}Y@{}}
        \includegraphics[width=\linewidth]{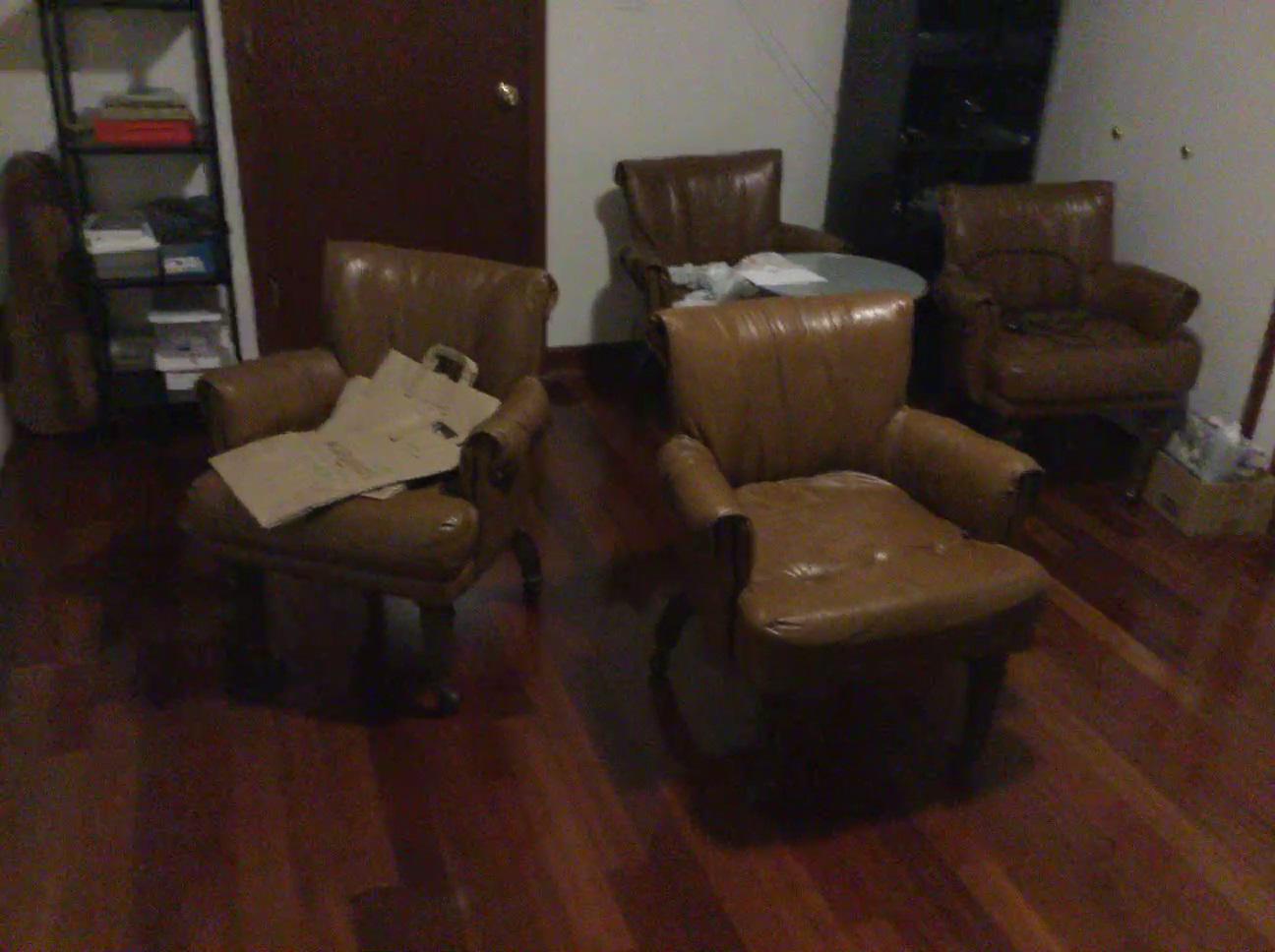}&
        \includegraphics[width=\linewidth]{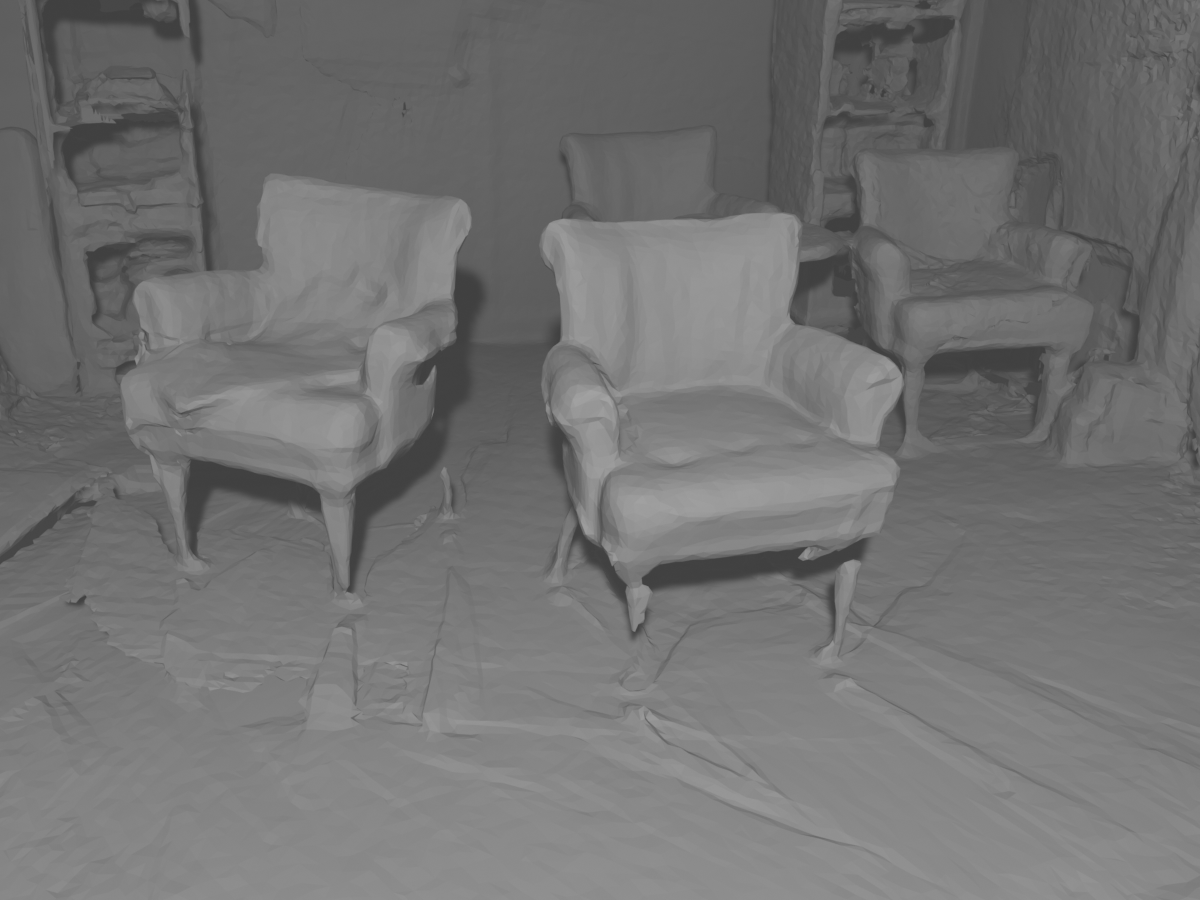}&
        \includegraphics[width=\linewidth]{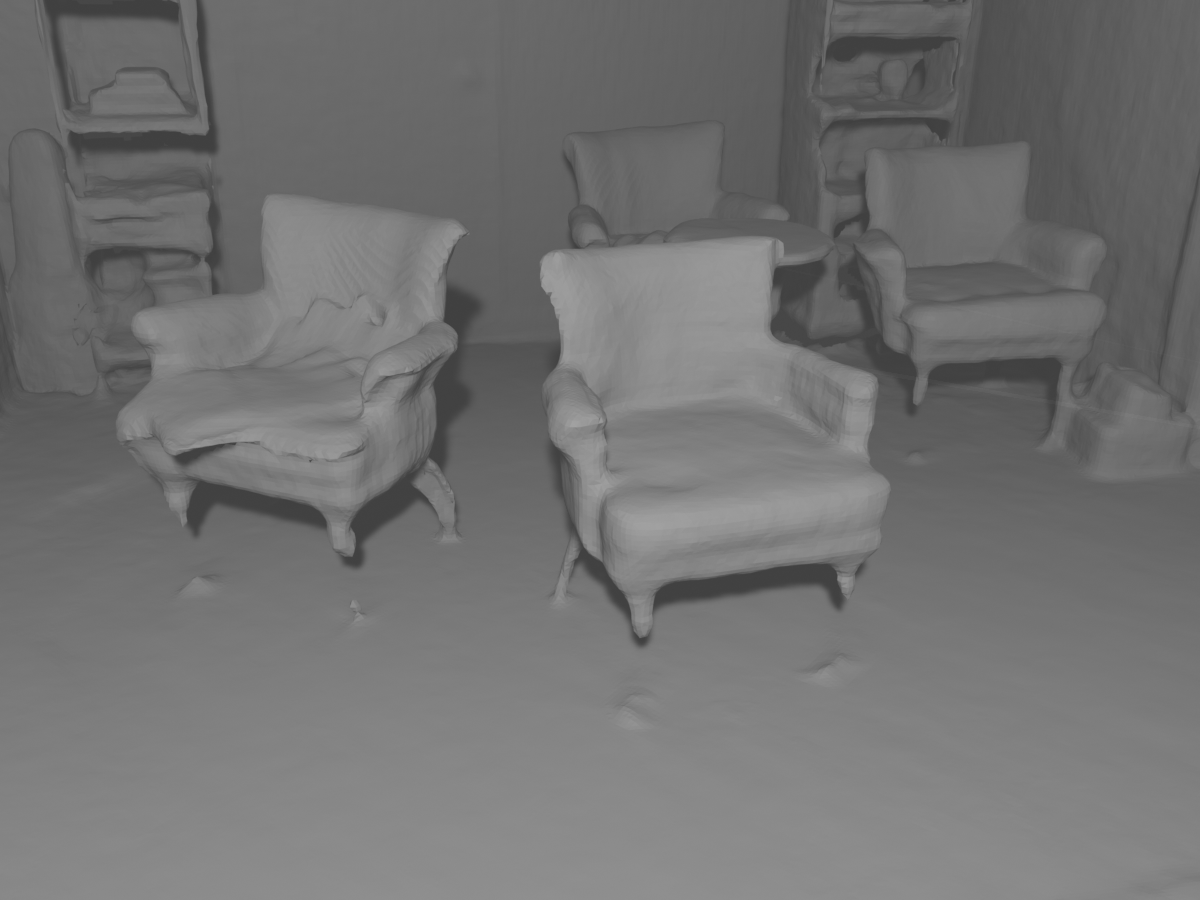}&
        \includegraphics[width=\linewidth]{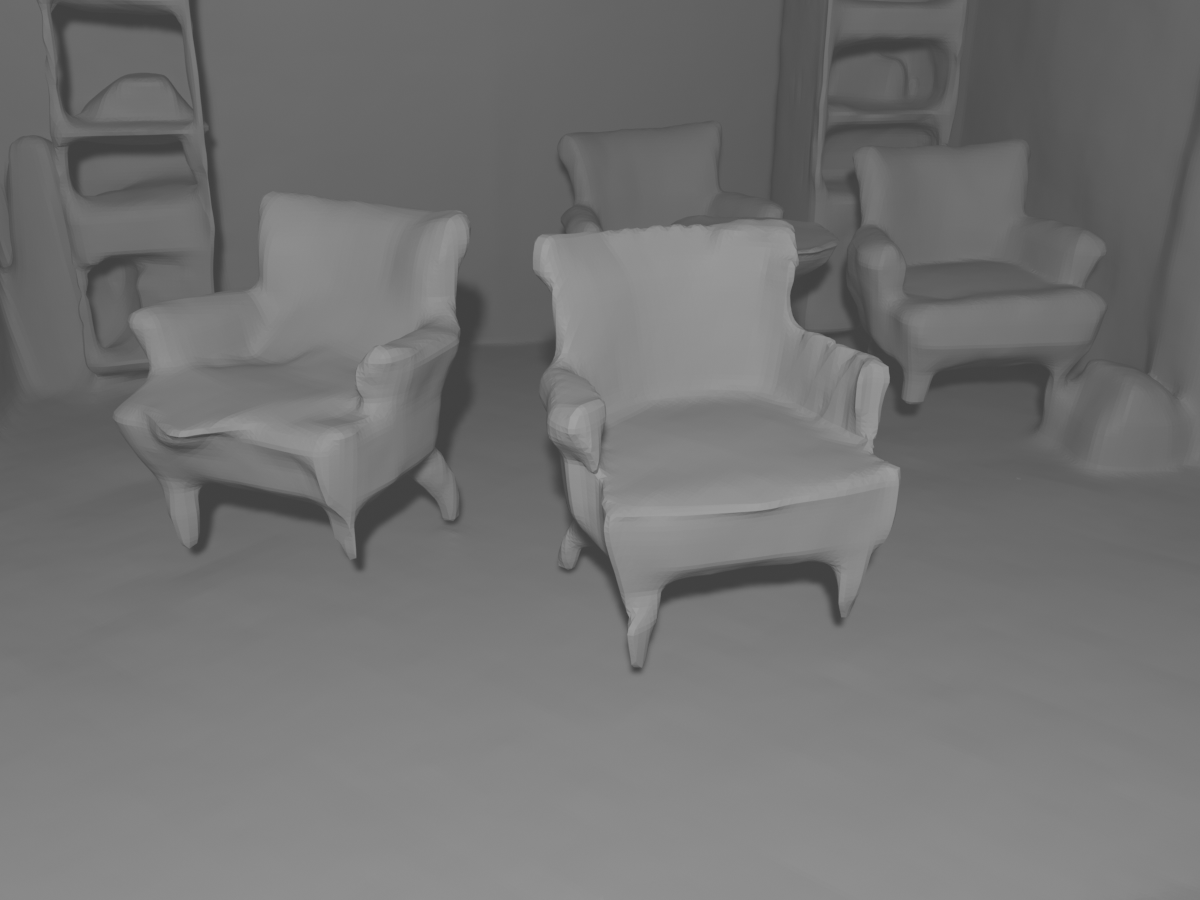}\\
        \includegraphics[width=\linewidth]{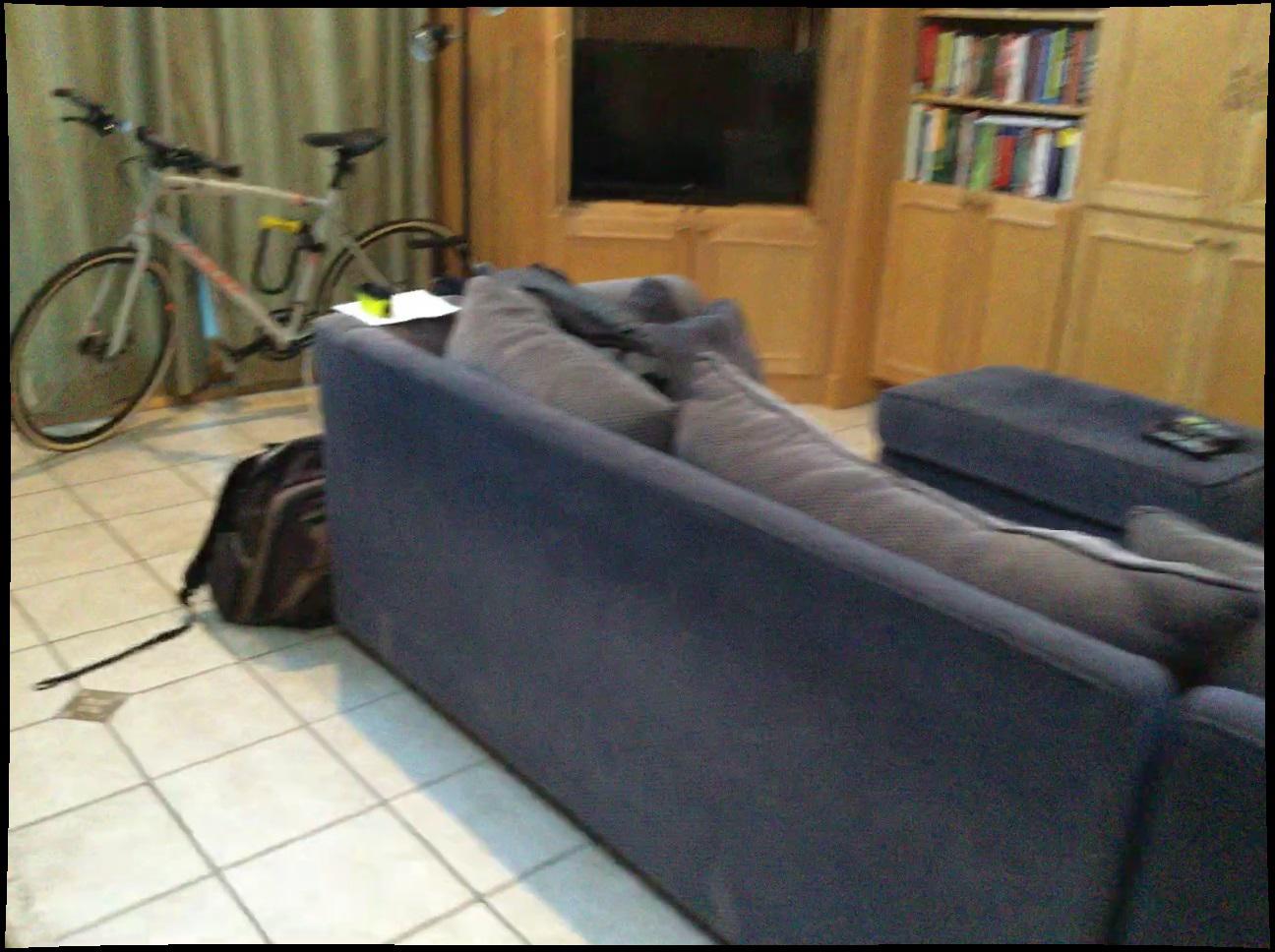}&
        \includegraphics[width=\linewidth]{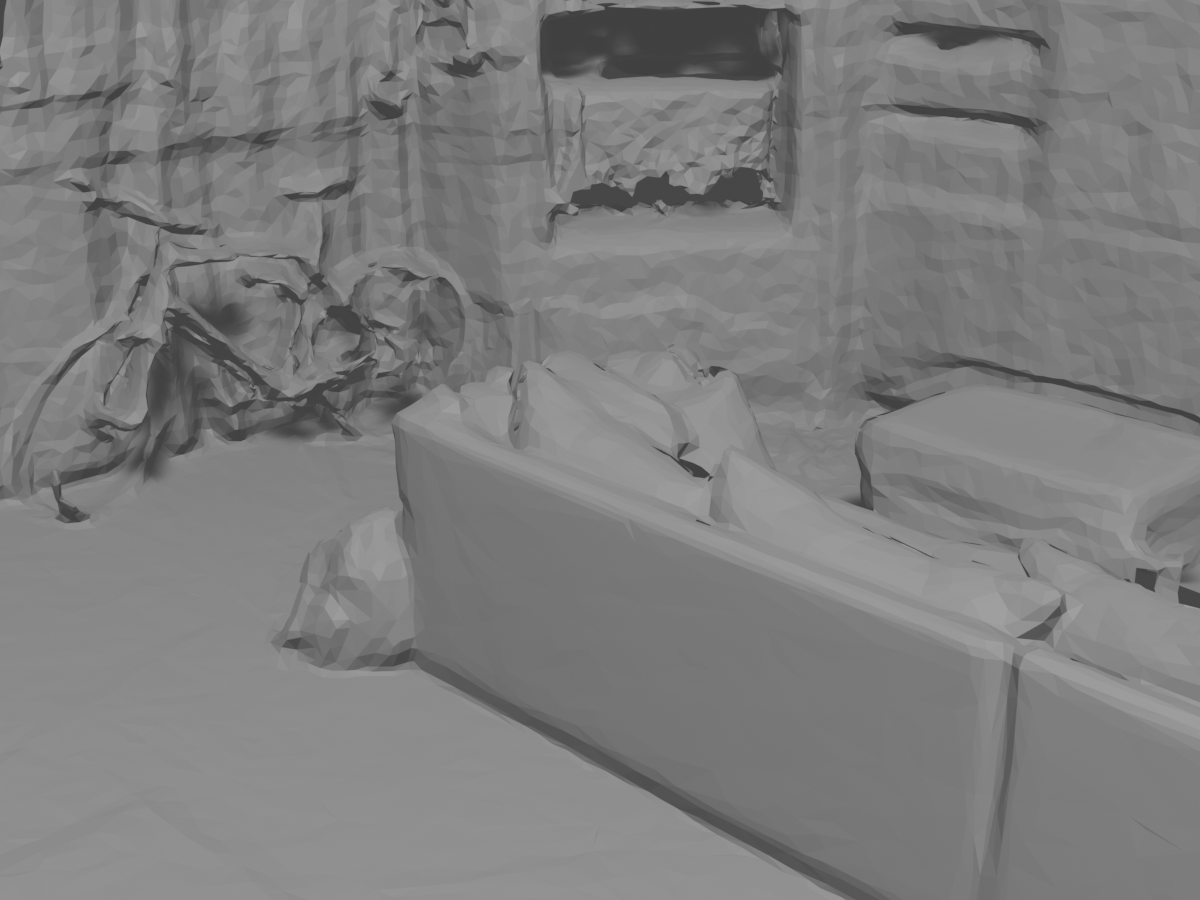}&
        \includegraphics[width=\linewidth]{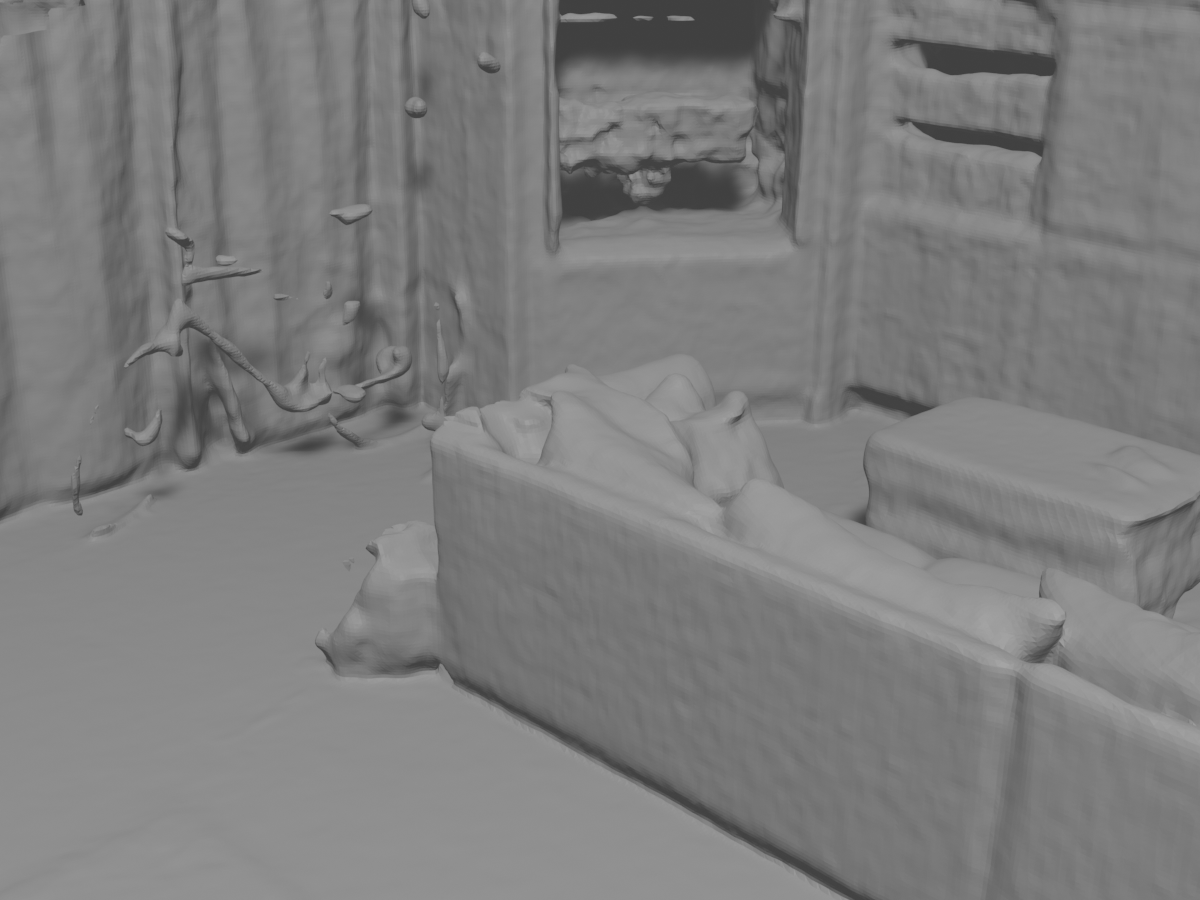}&
        \includegraphics[width=\linewidth]{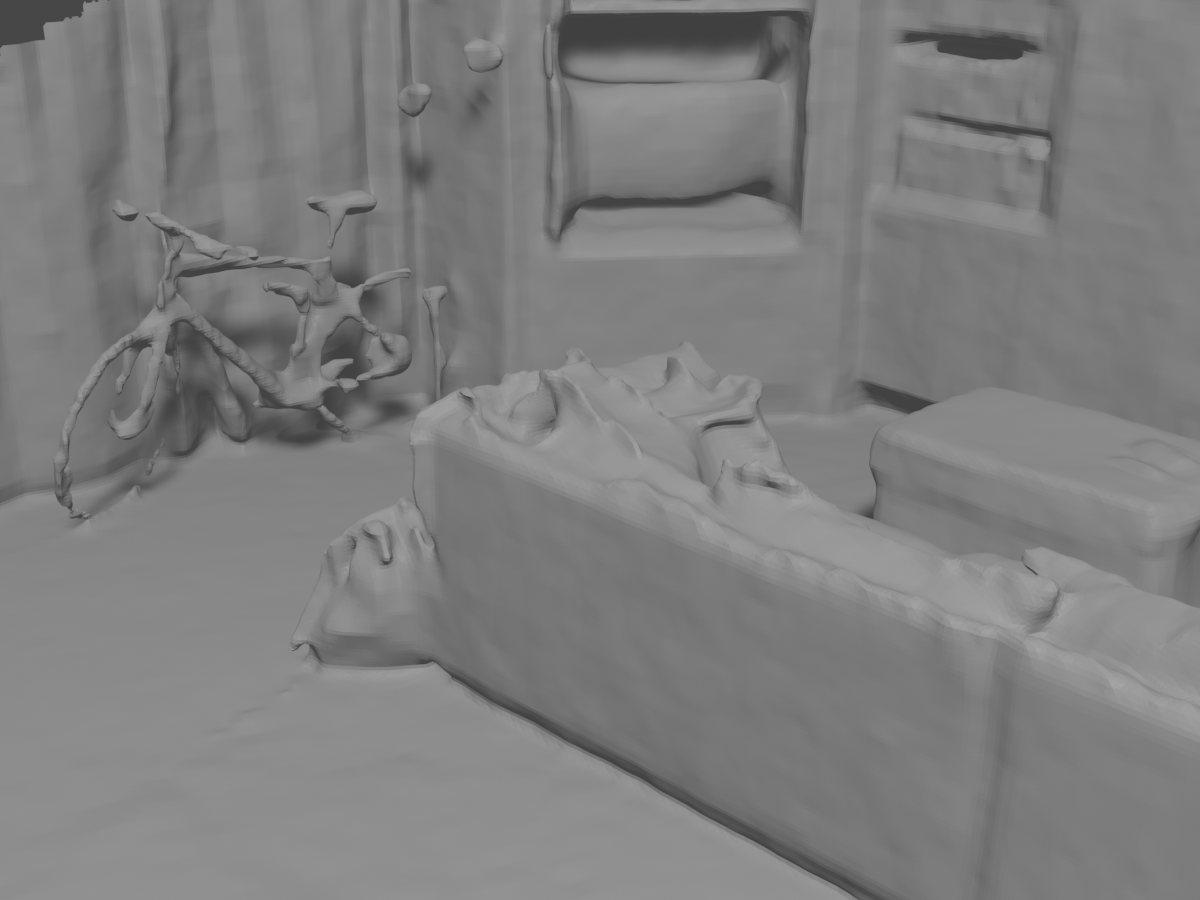}\\
        \includegraphics[width=\linewidth]{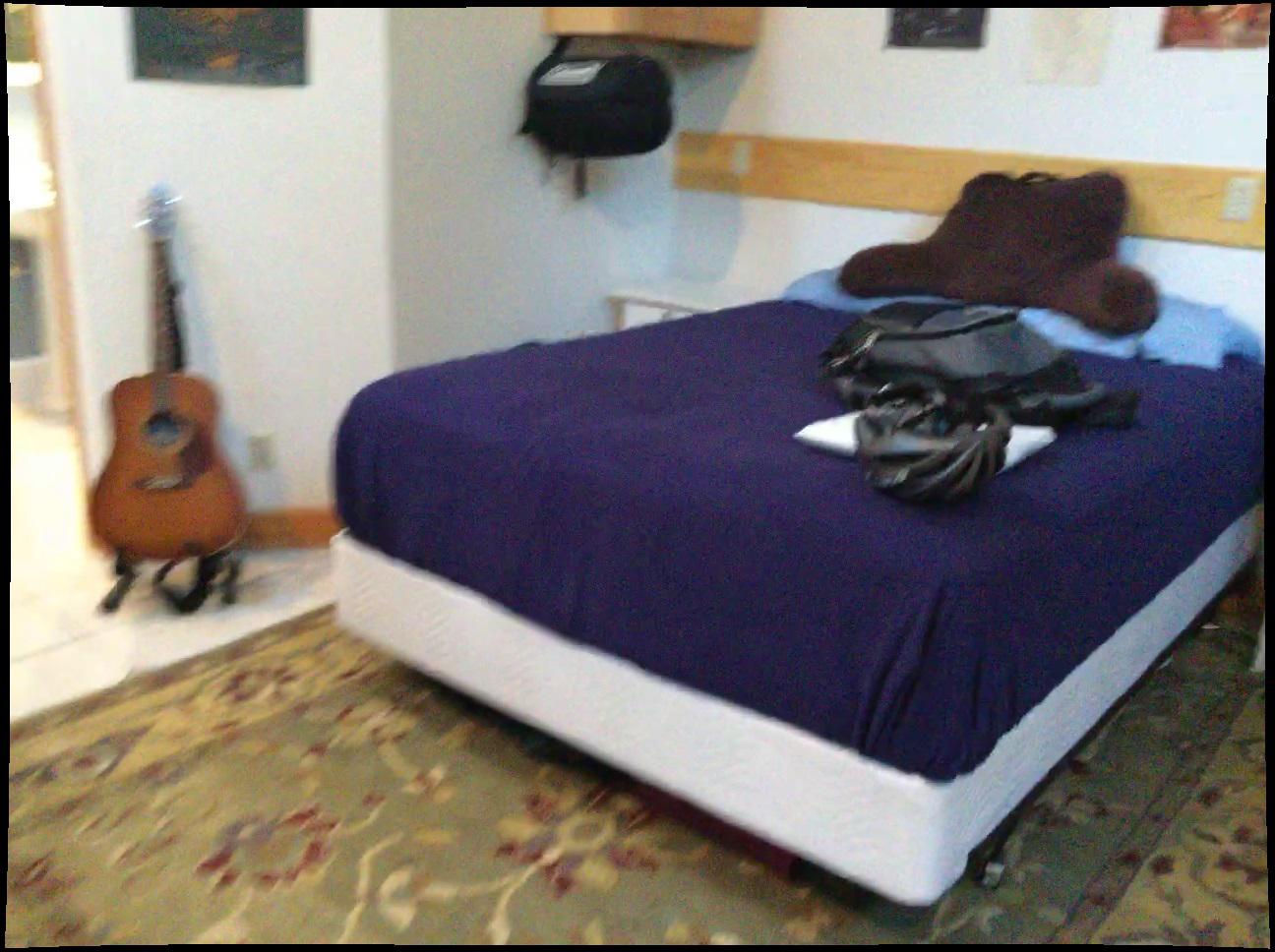}&
        \includegraphics[width=\linewidth]{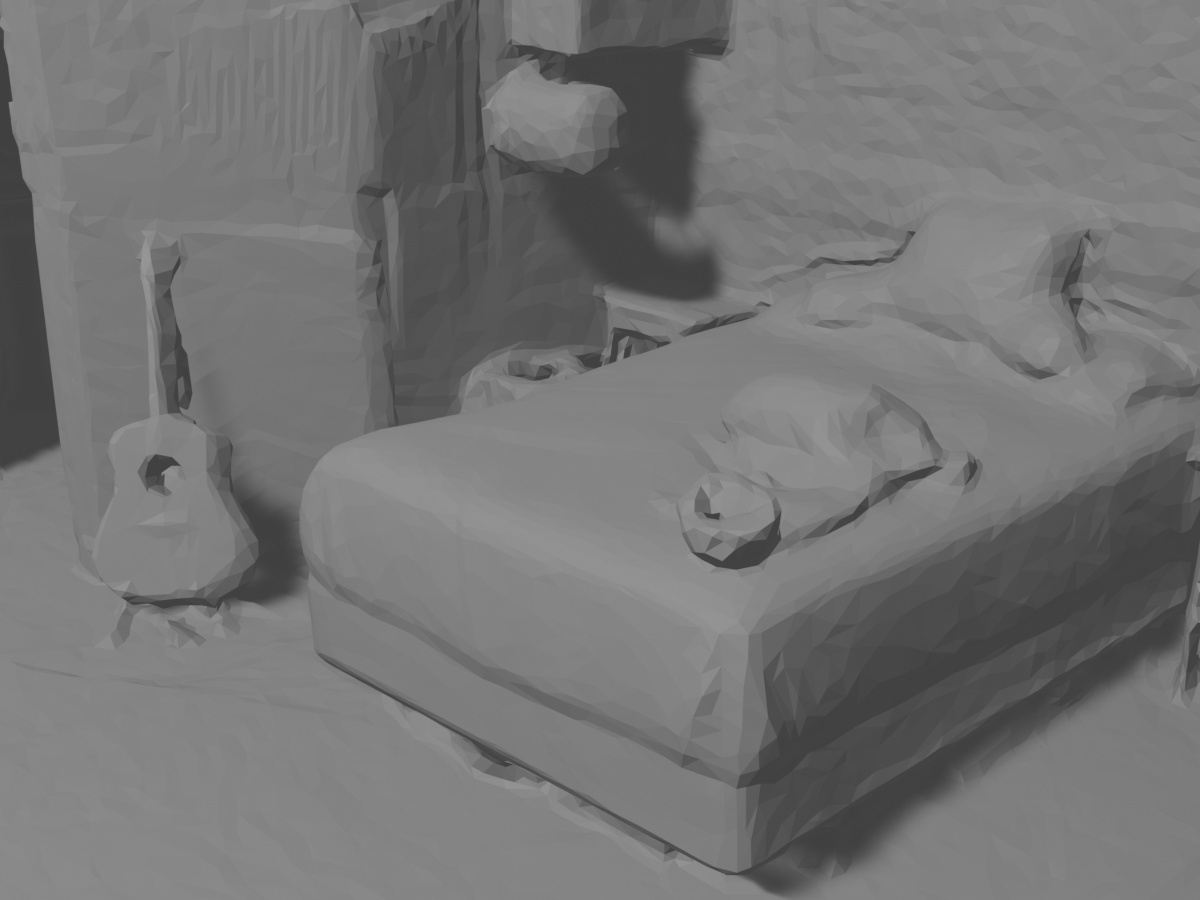}&
        \includegraphics[width=\linewidth]{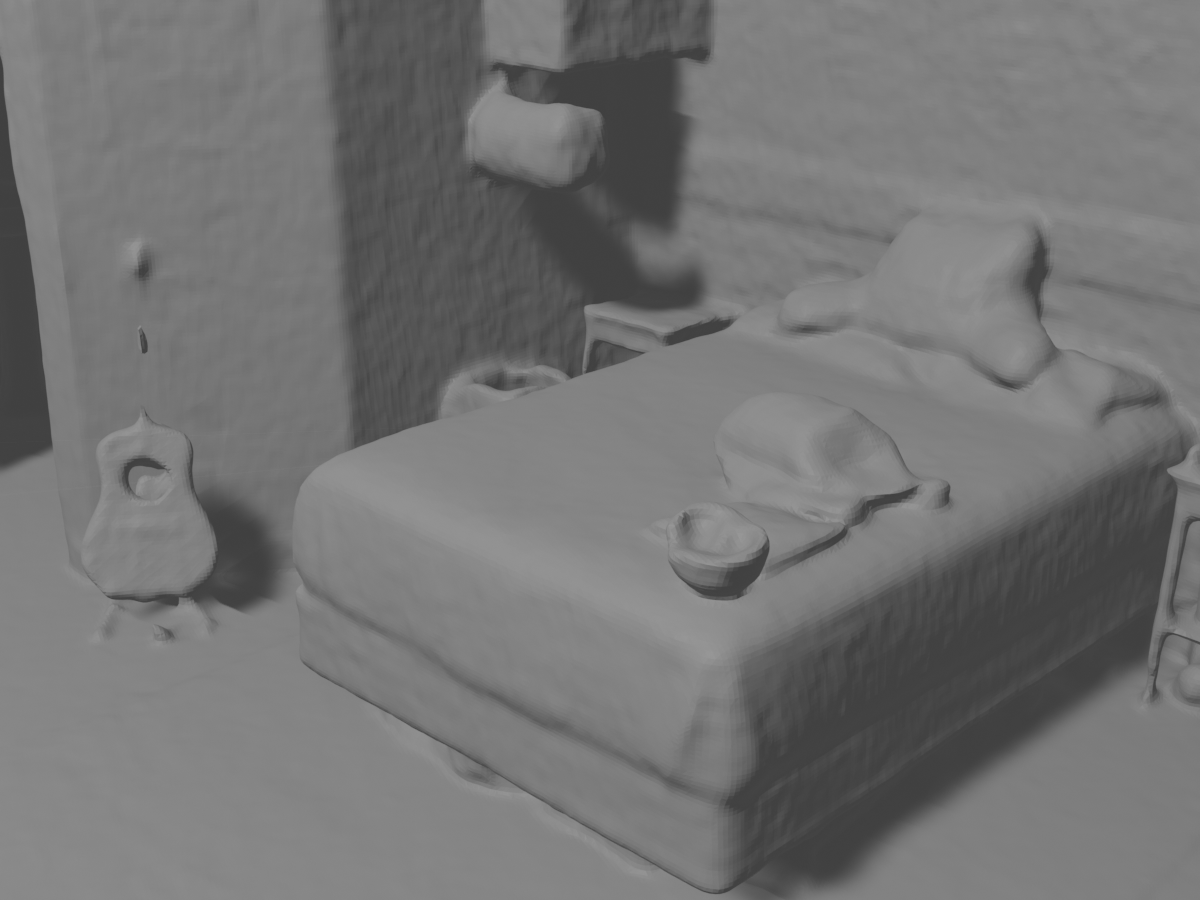}&
        \includegraphics[width=\linewidth]{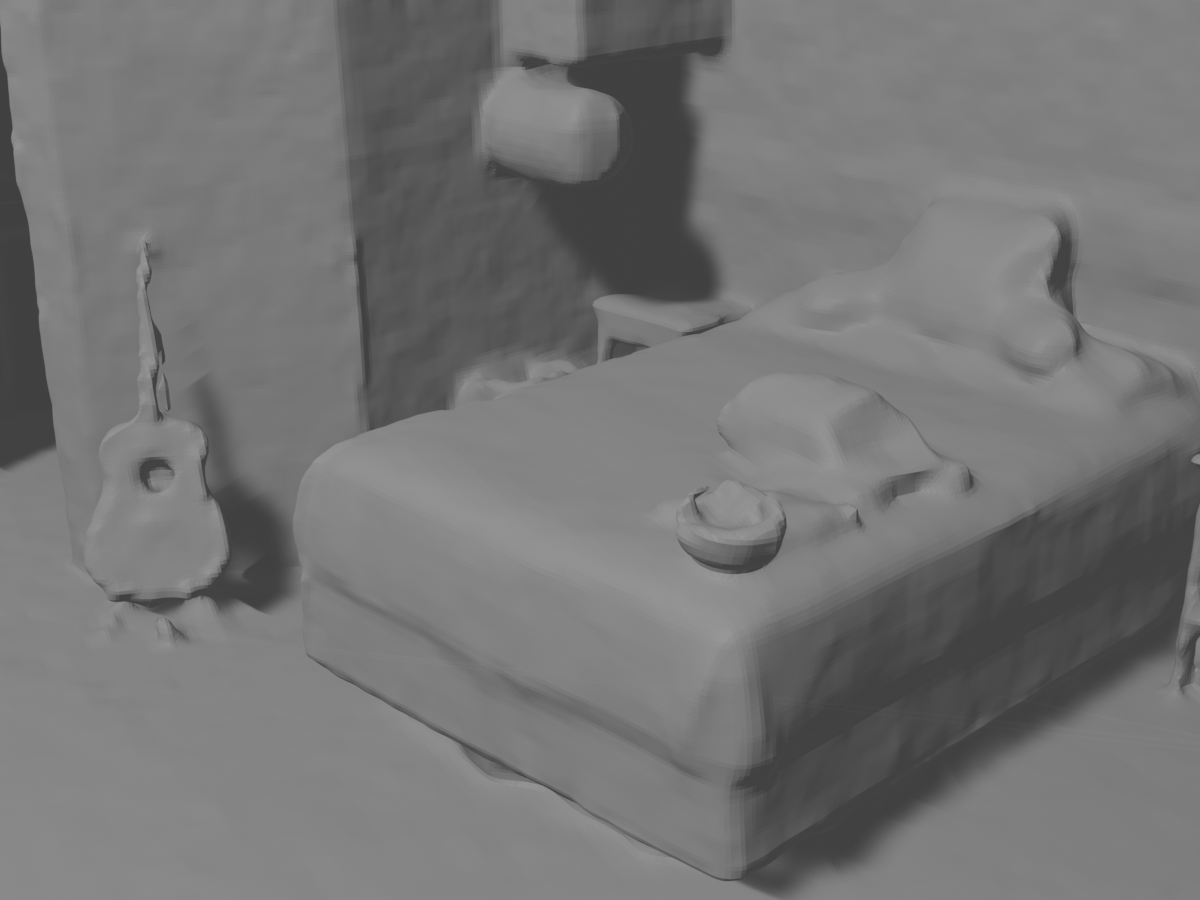}\\
        \includegraphics[width=\linewidth]{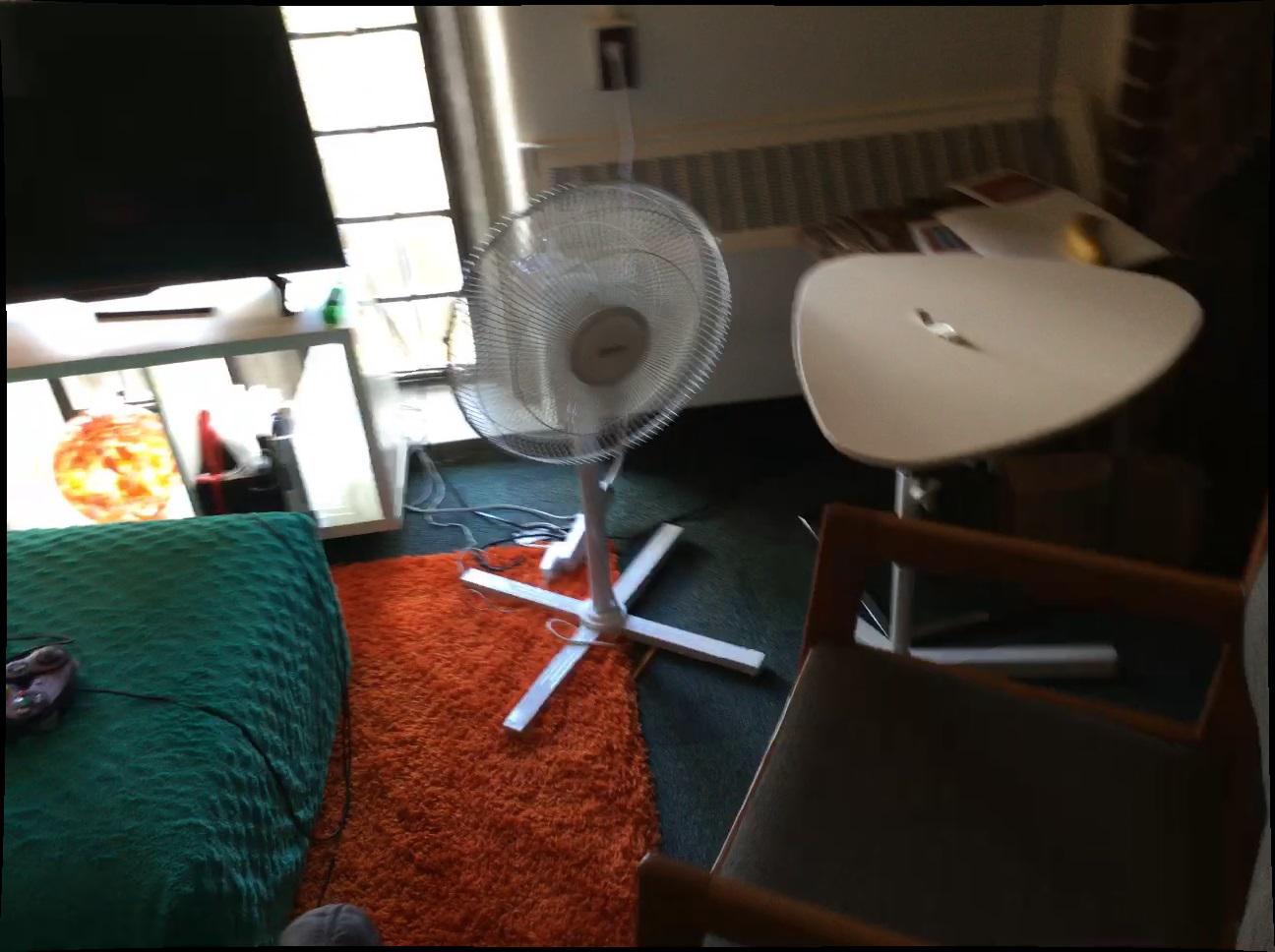}&
        \includegraphics[width=\linewidth]{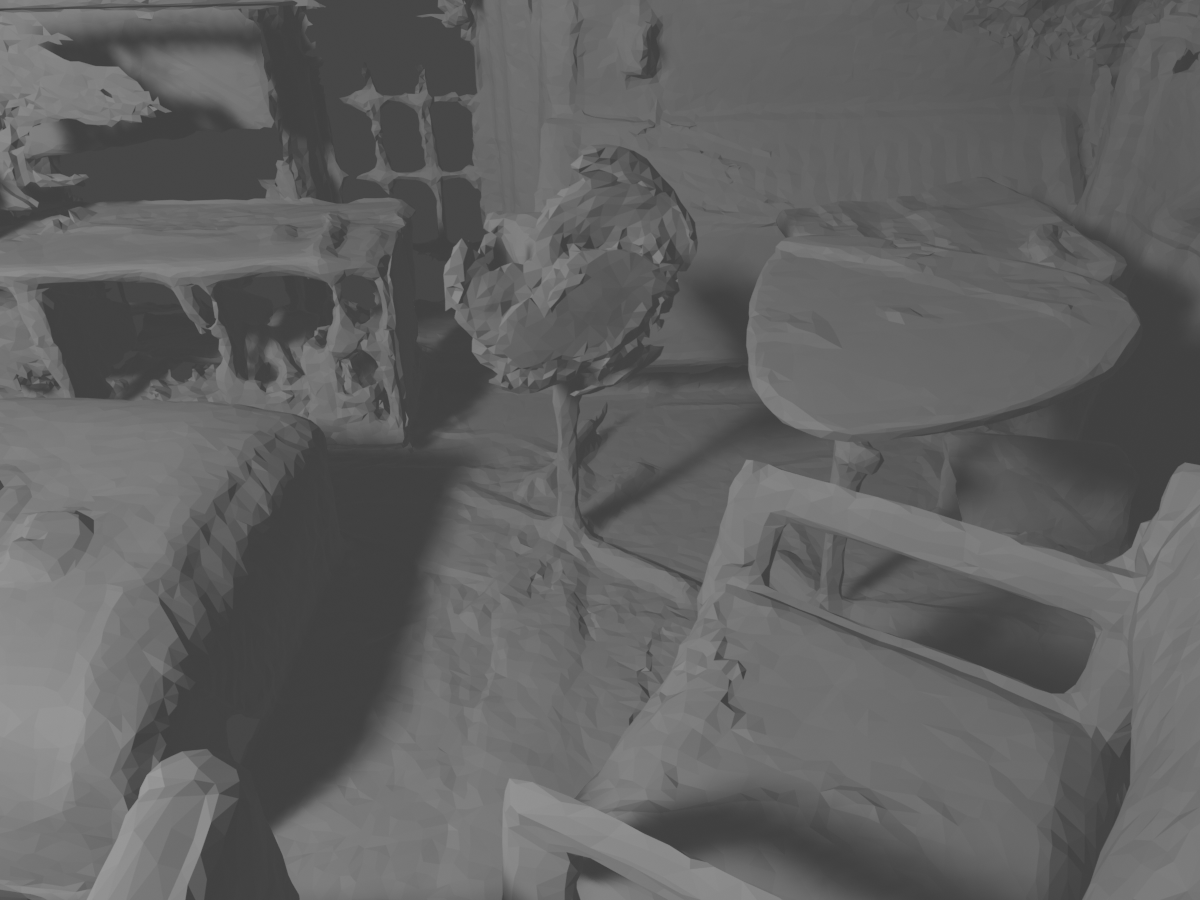}&
        \includegraphics[width=\linewidth]{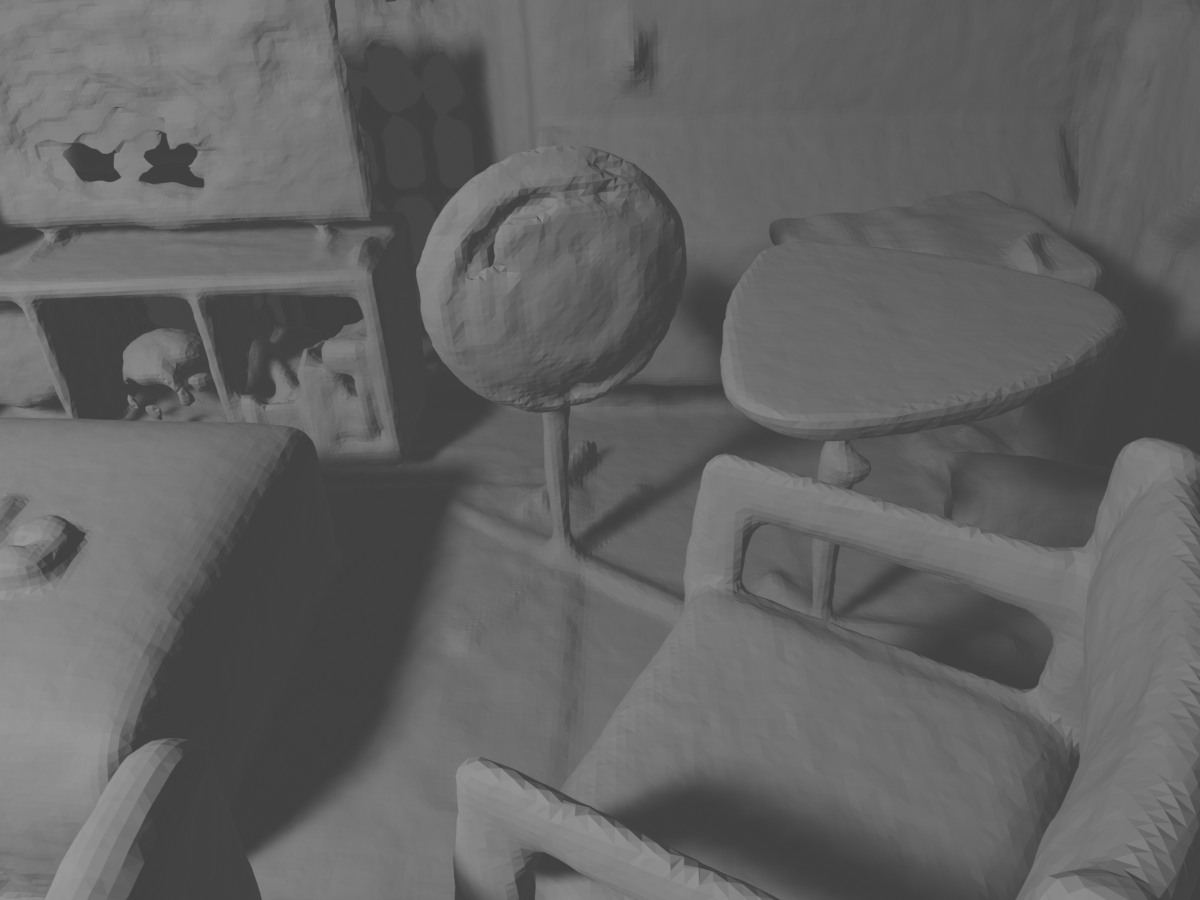}&
        \includegraphics[width=\linewidth]{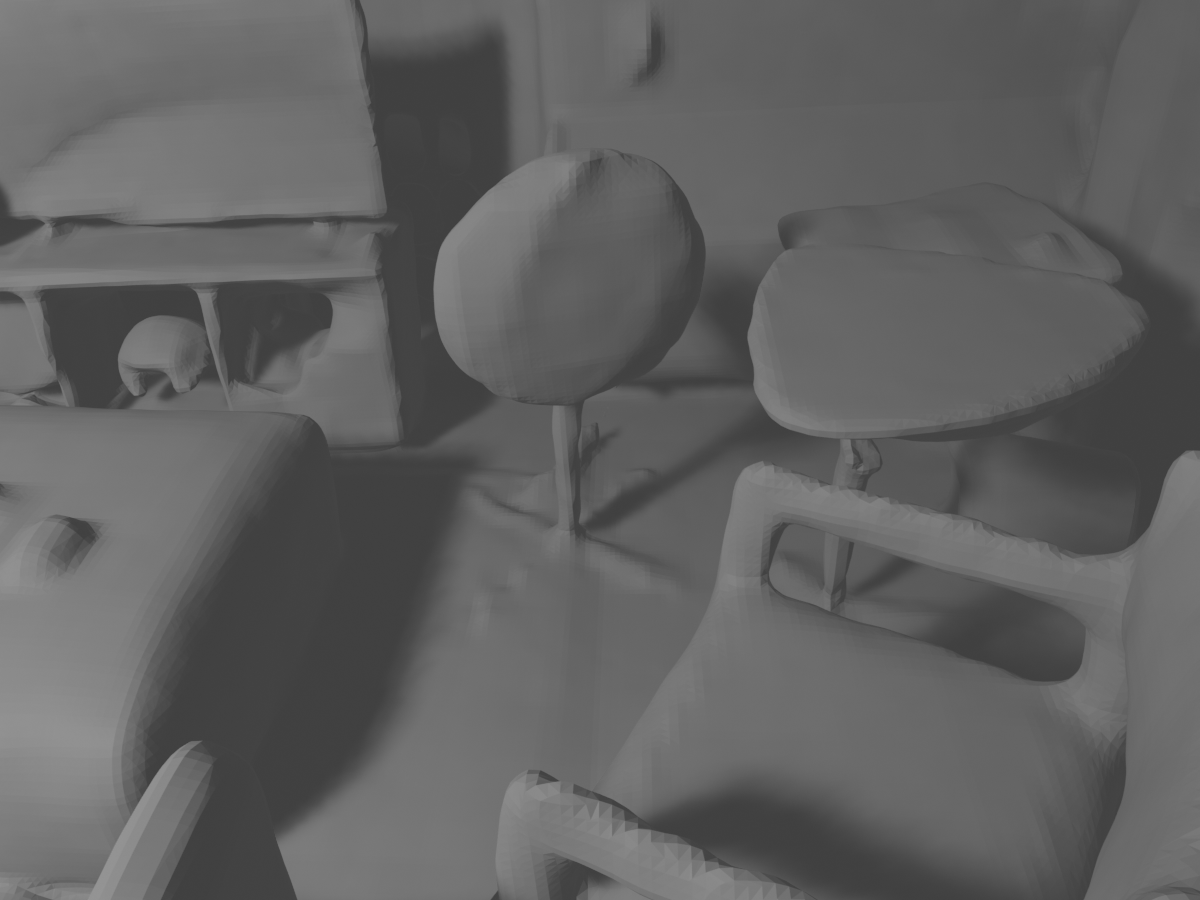}\\
        \includegraphics[width=\linewidth]{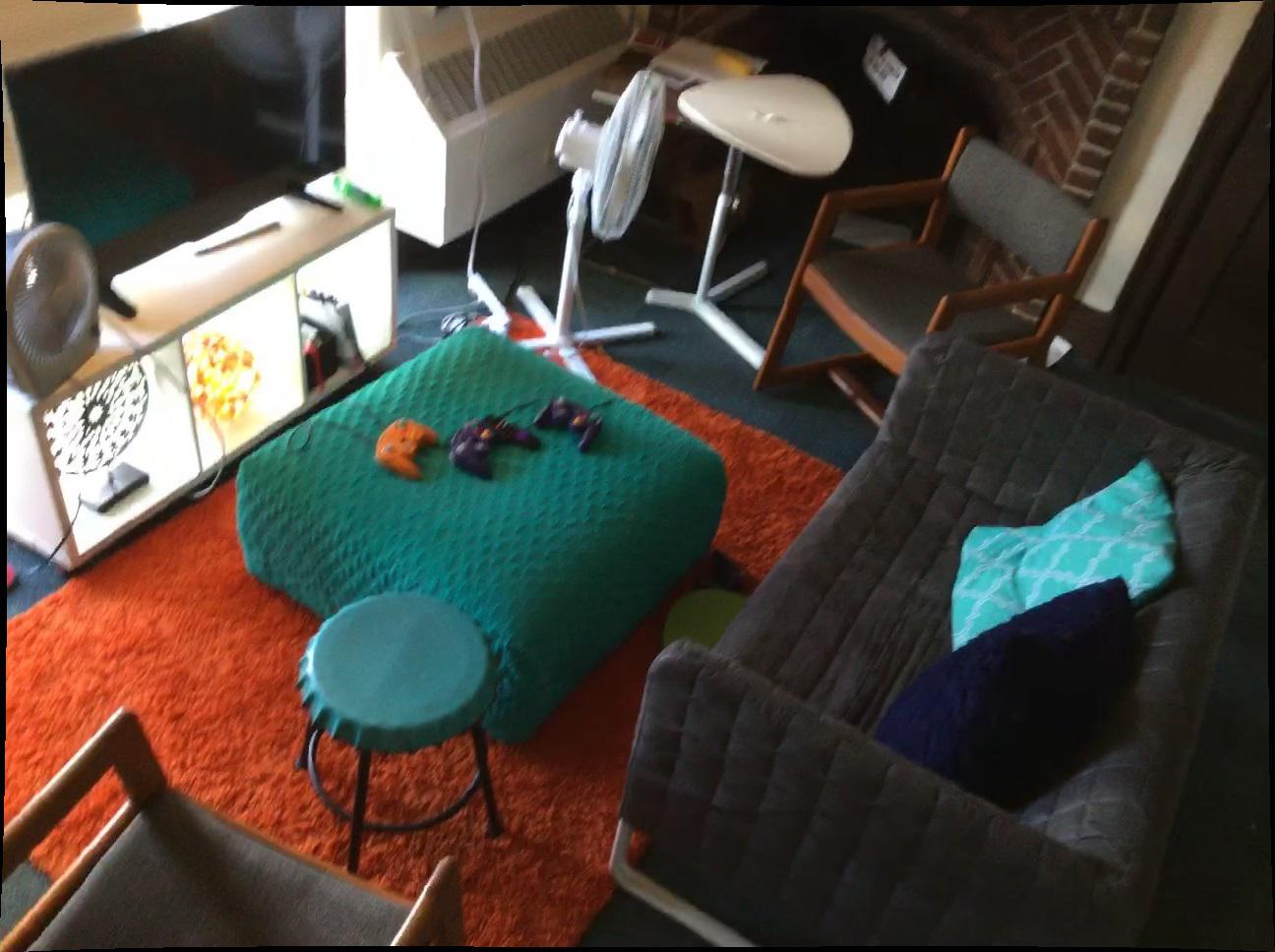}&
        \includegraphics[width=\linewidth]{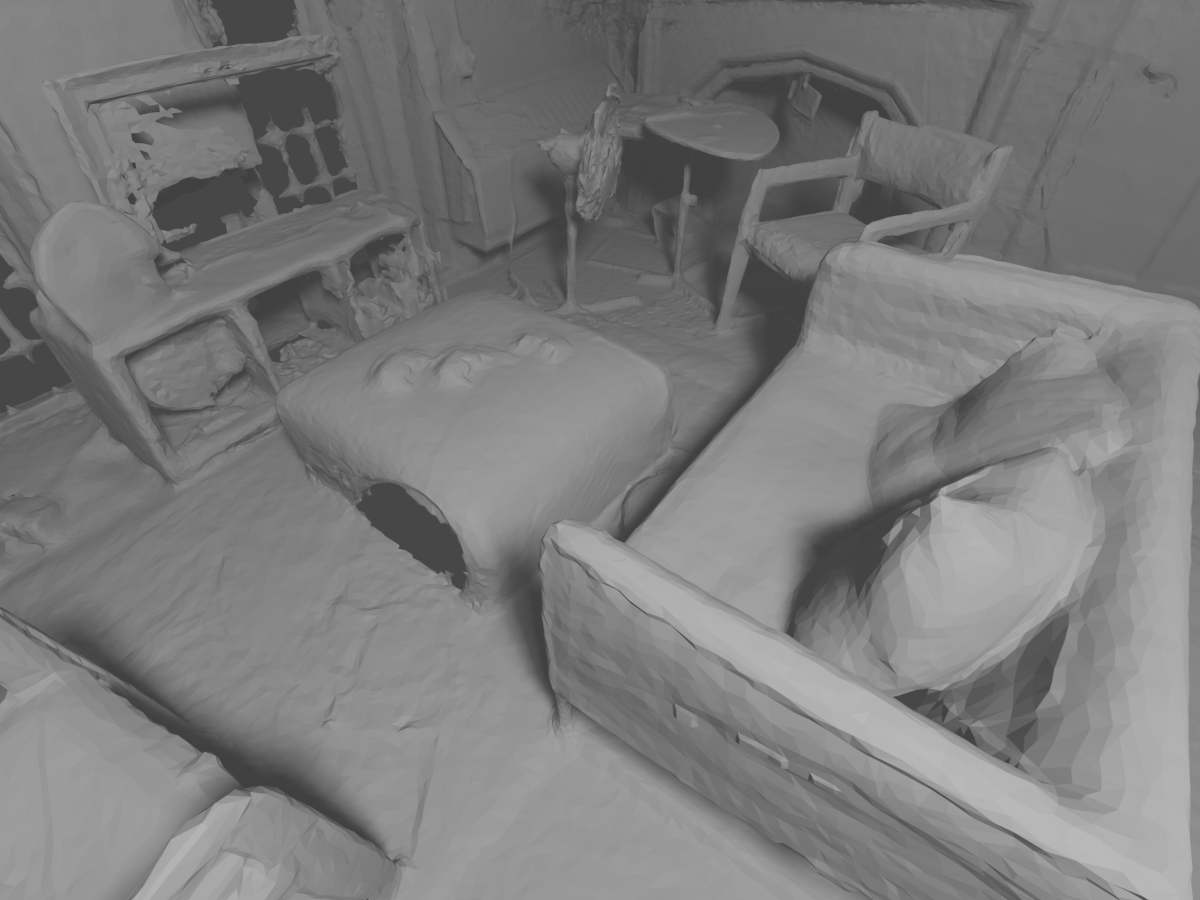}&
        \includegraphics[width=\linewidth]{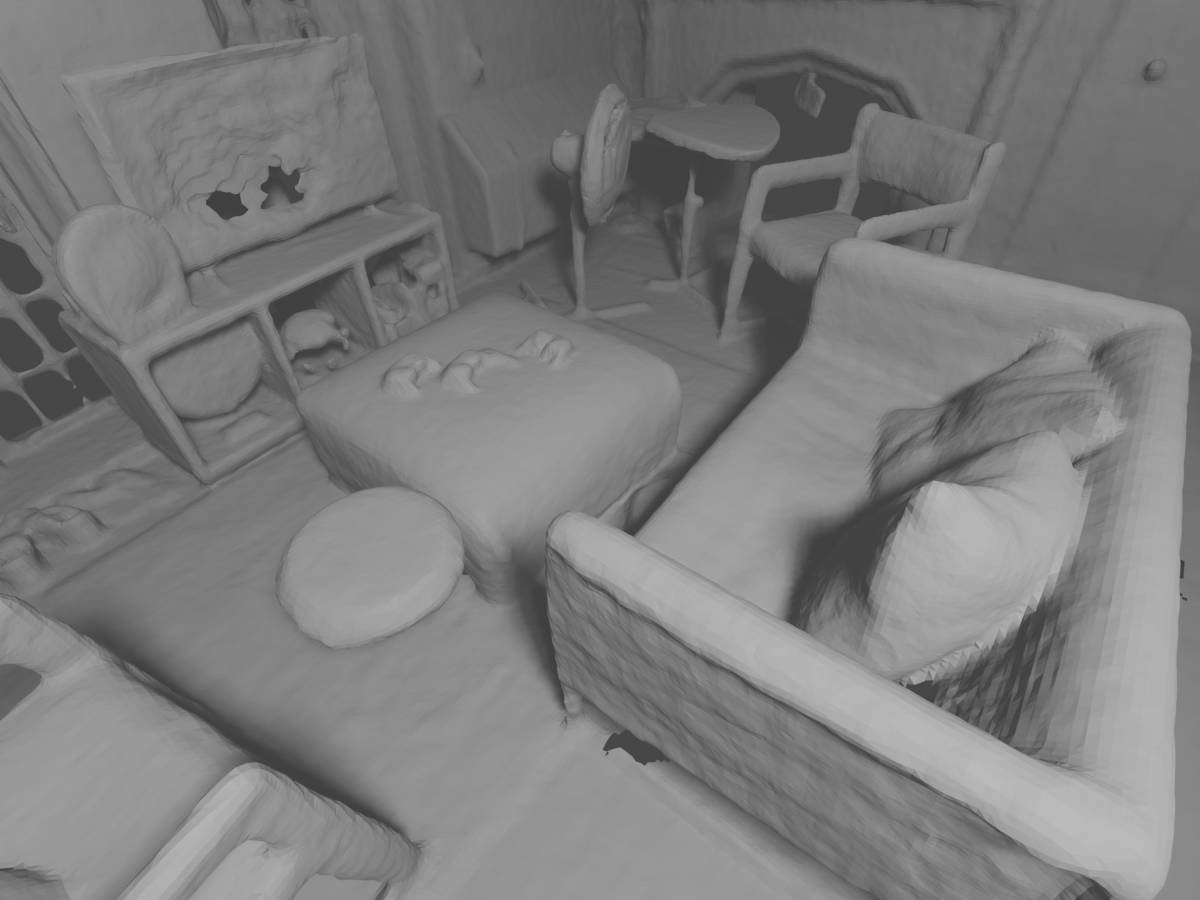}&
        \includegraphics[width=\linewidth]{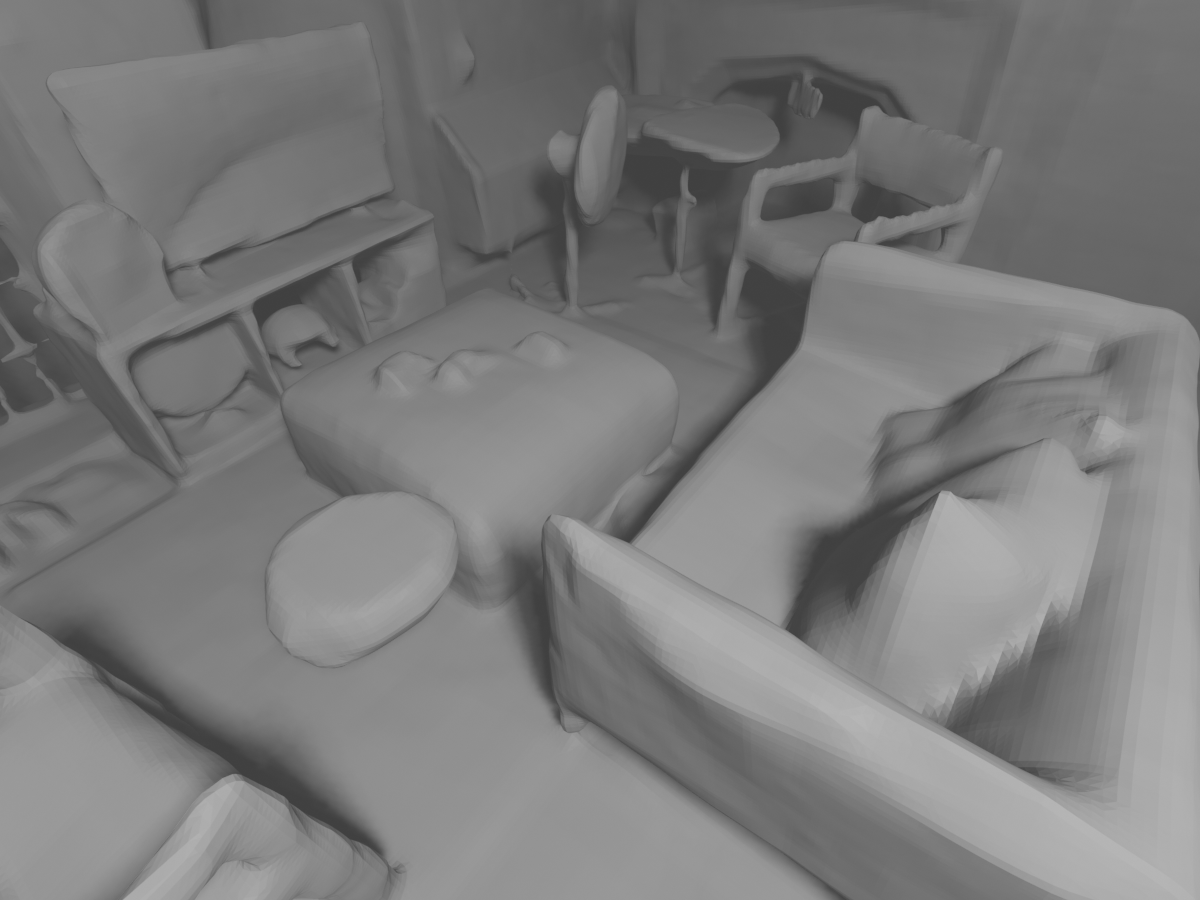}\\
        \includegraphics[width=\linewidth]{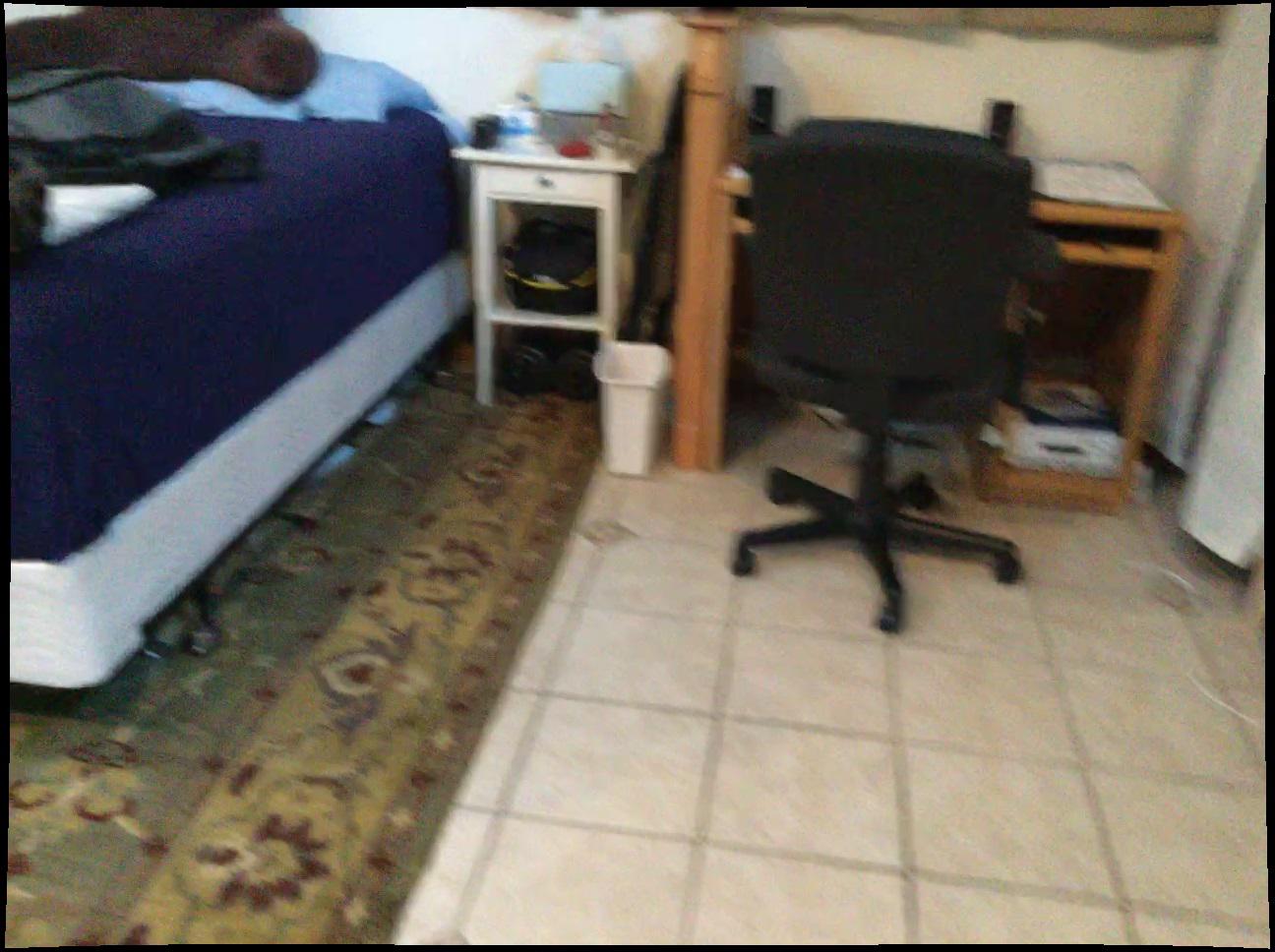}&
        \includegraphics[width=\linewidth]{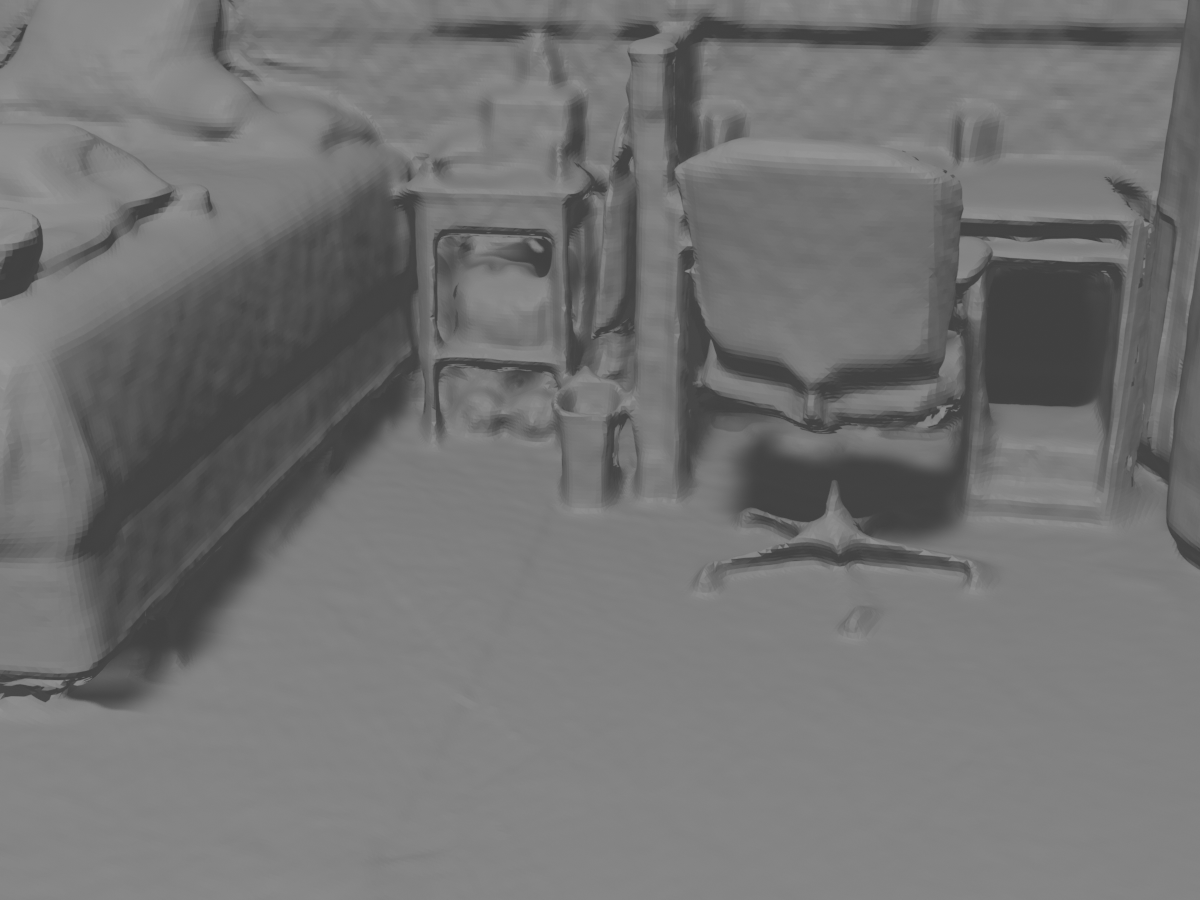}&
        \includegraphics[width=\linewidth]{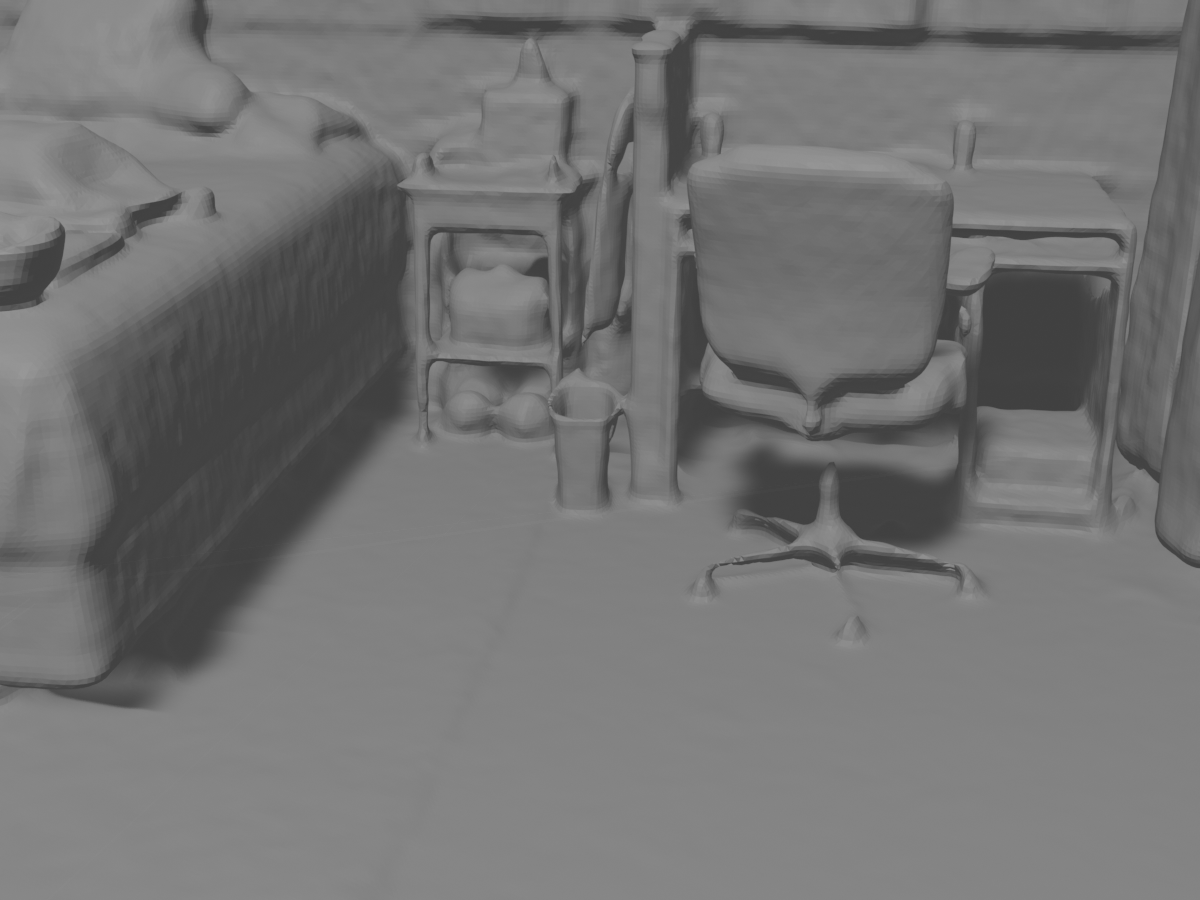}&
        \includegraphics[width=\linewidth]{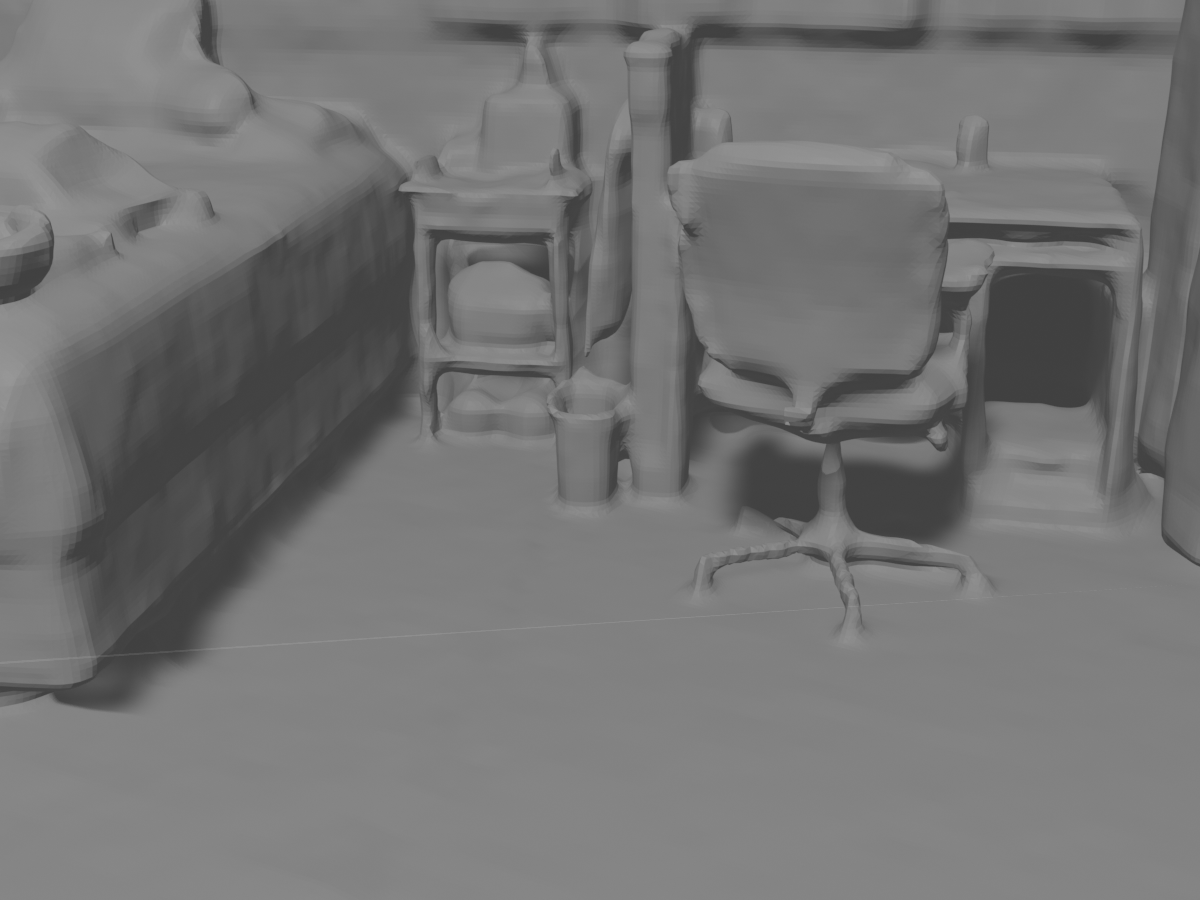}\\
         RGB &
         BundleFusion~\cite{dai2017bundlefusion} &
         Neural RGBD~\cite{azinovic2022neural} &
         GO-Surf (Ours)
    \end{tabularx}
    \vspace{-1mm}
    \caption{We compare our method to Neural RGB-D Surface Reconstruction~\cite{azinovic2022neural} and BundleFusion~\cite{dai2017bundlefusion} on scenes 0, 2, and 12 of the ScanNet dataset. Qualitative results show that our method displays higher completion and smoothness than the baselines. Our method is also able to produce high-quality reconstructions for missing depth regions and thin structures.  \label{fig:scannet_result}}
    \vspace{-4mm}
\end{figure*}
\begin{figure*}[tp]
    \centering
    \newcolumntype{Y}{>{\centering\arraybackslash}X}
    \begin{tabularx}{0.95\linewidth}{@{}Y@{\,}Y@{\,}Y@{\,}Y@{}}
        \includegraphics[width=\linewidth]{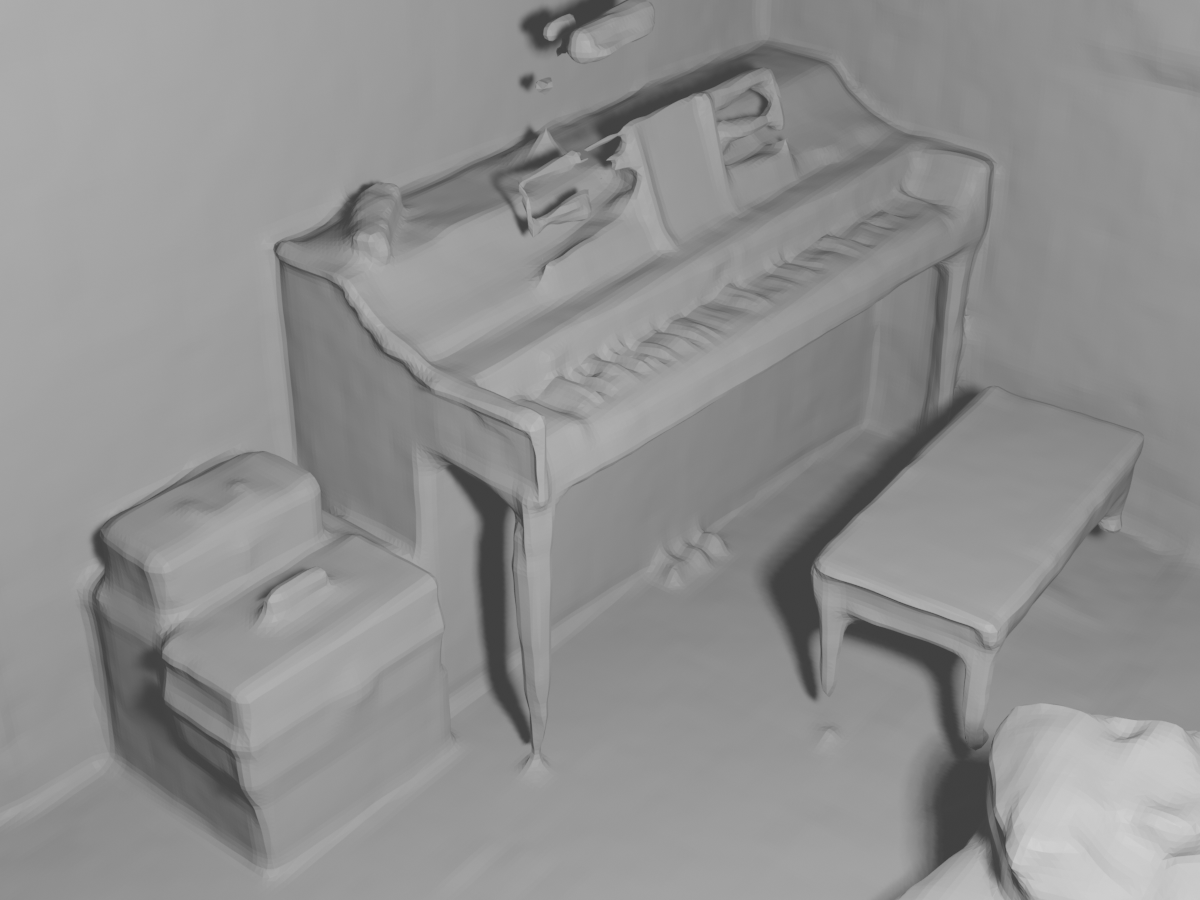}&
        \includegraphics[width=\linewidth]{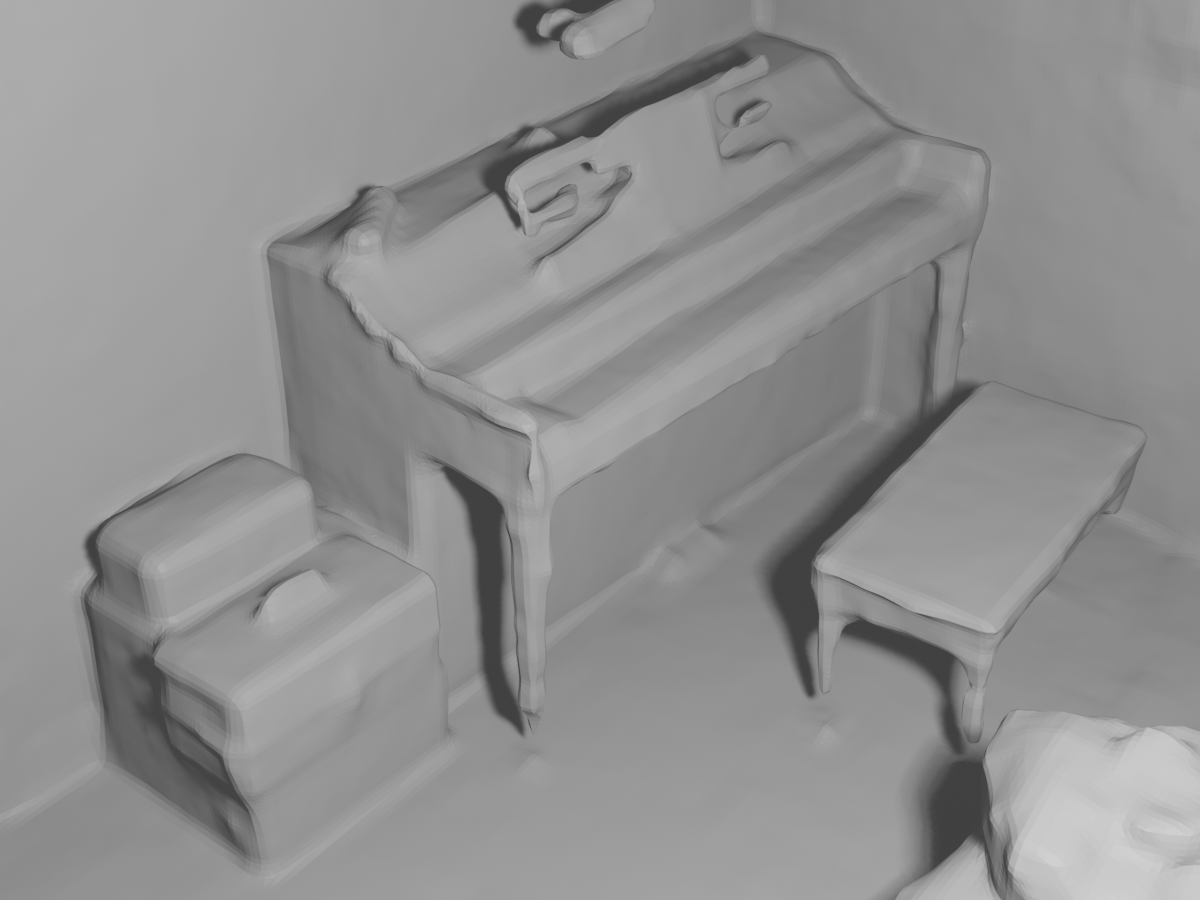}&
        \includegraphics[width=\linewidth]{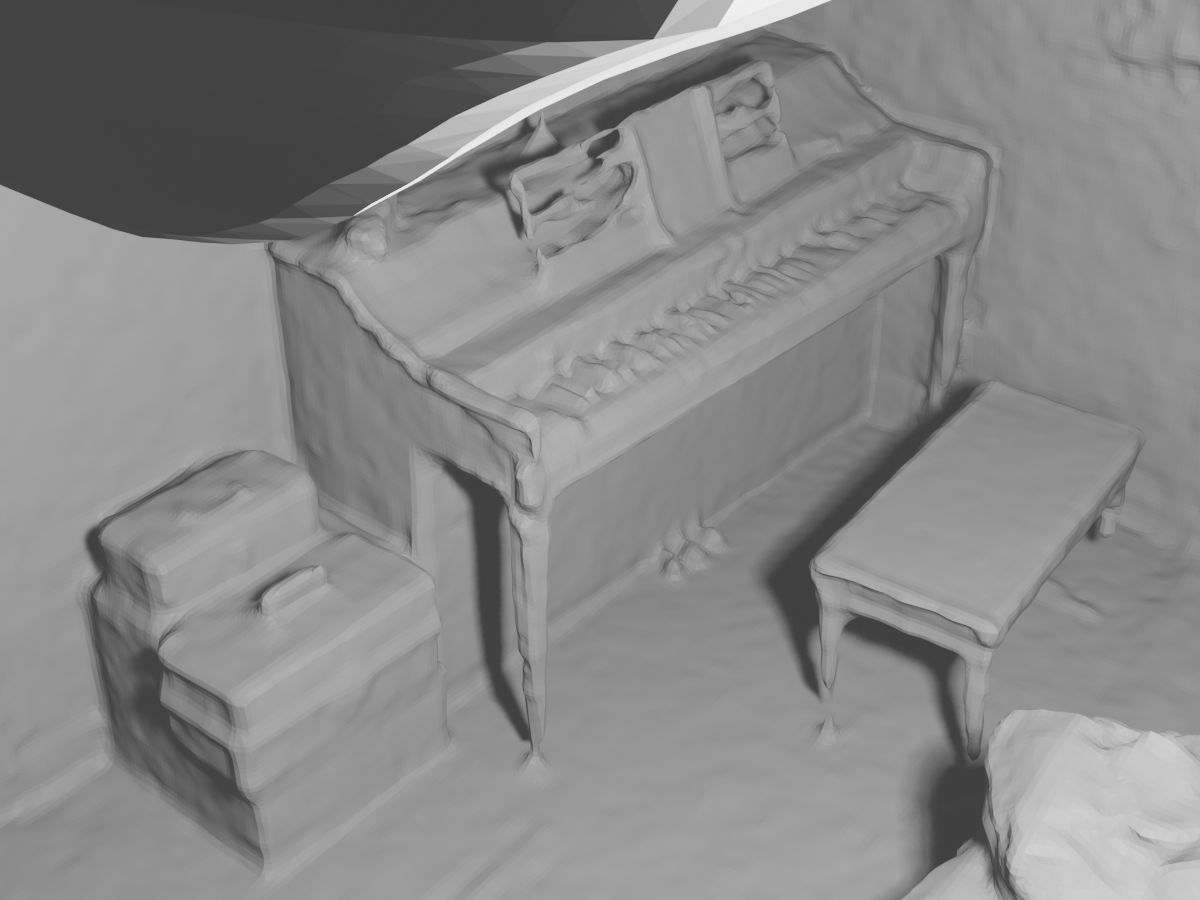}&
        \includegraphics[width=\linewidth]{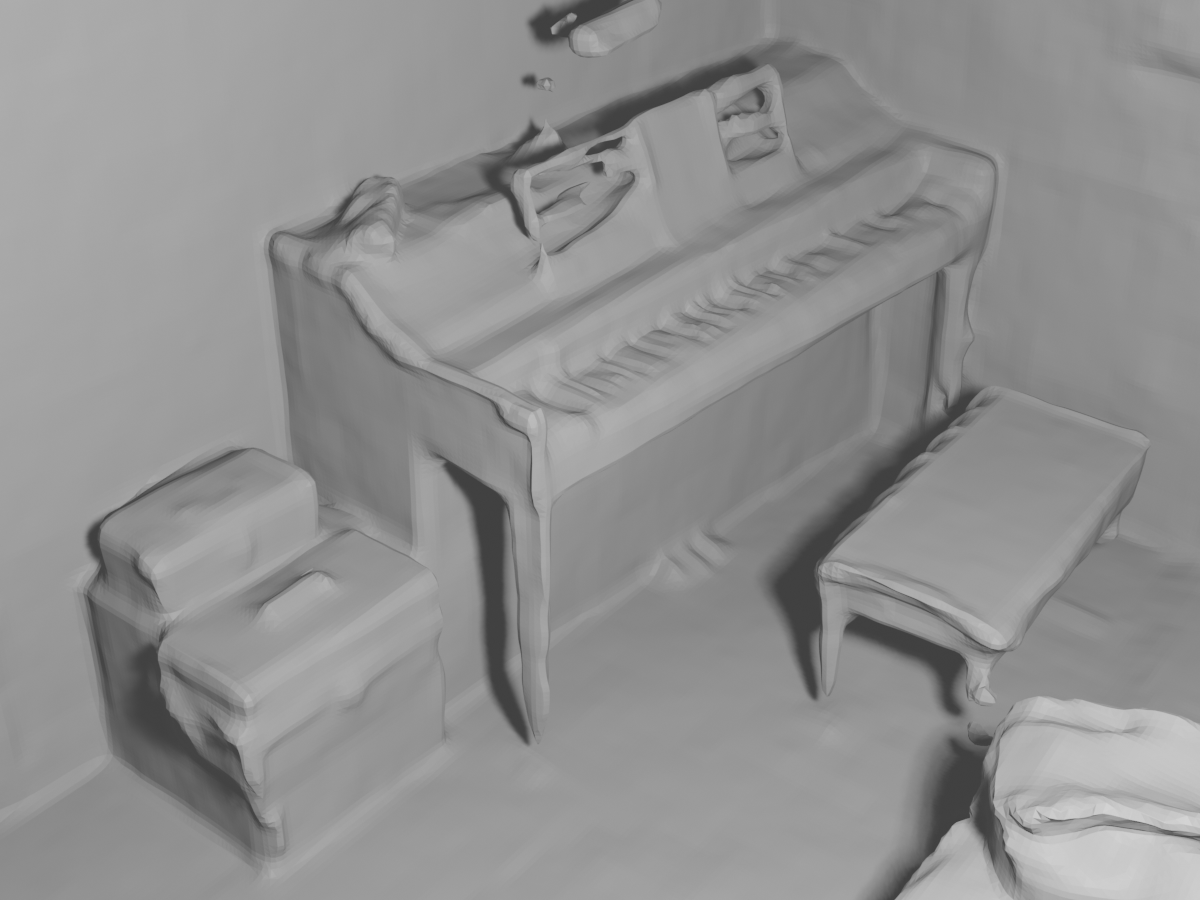}\\
        \includegraphics[width=\linewidth]{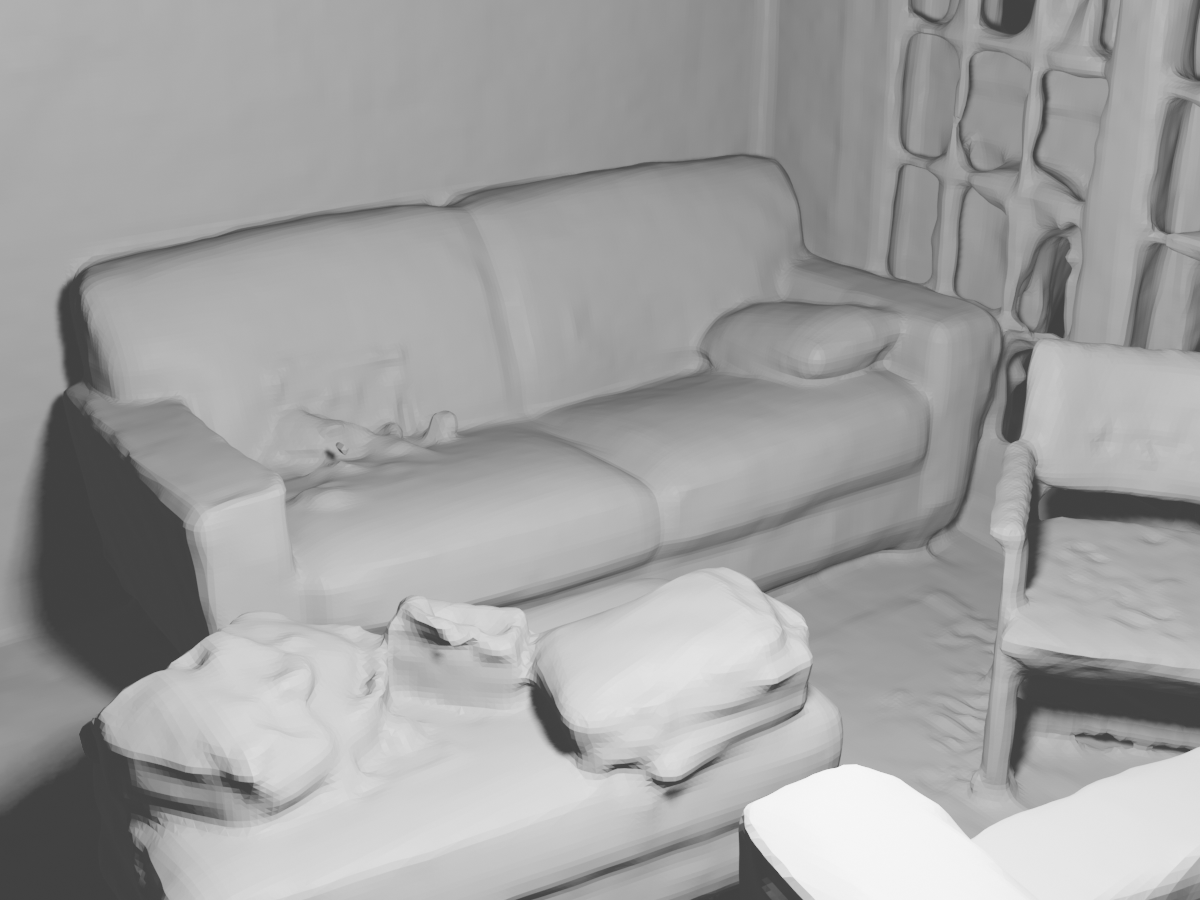}&
        \includegraphics[width=\linewidth]{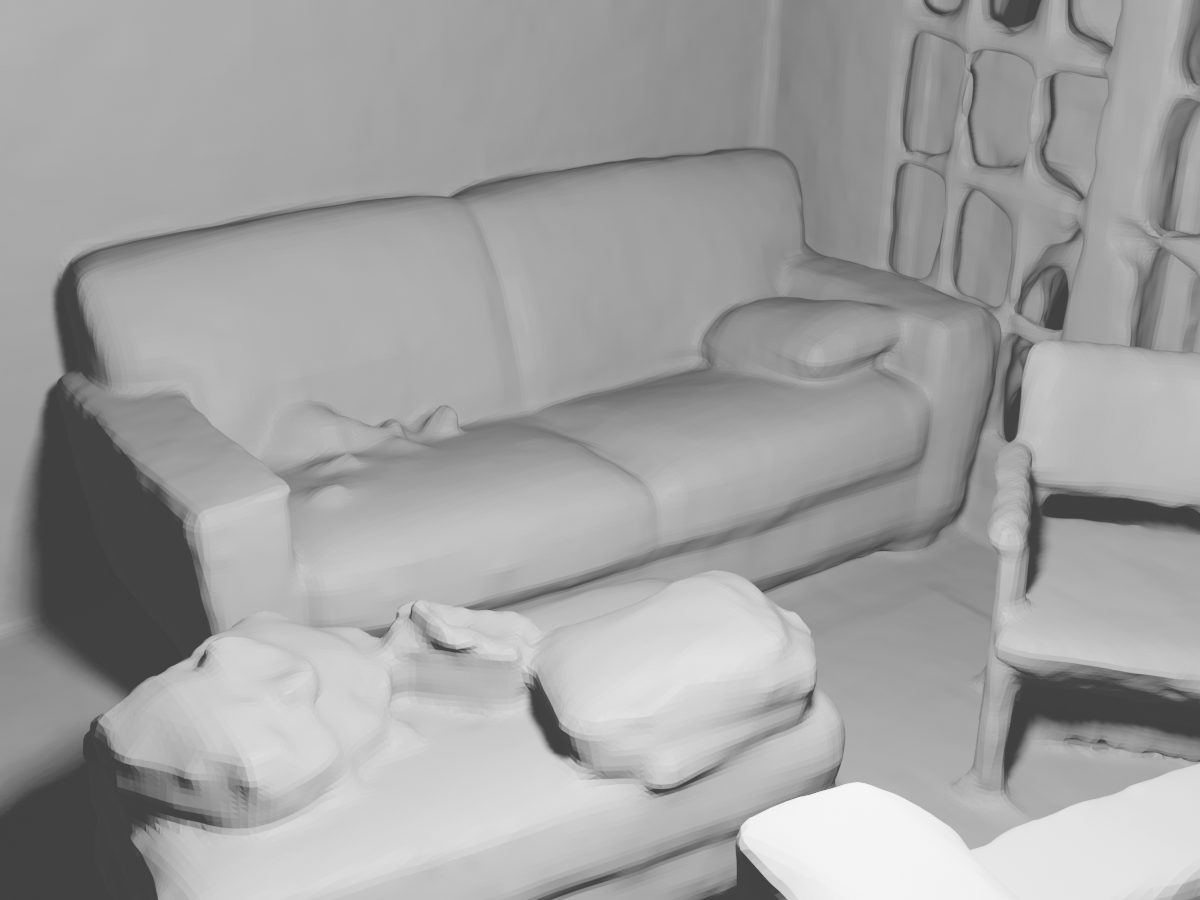}&
        \includegraphics[width=\linewidth]{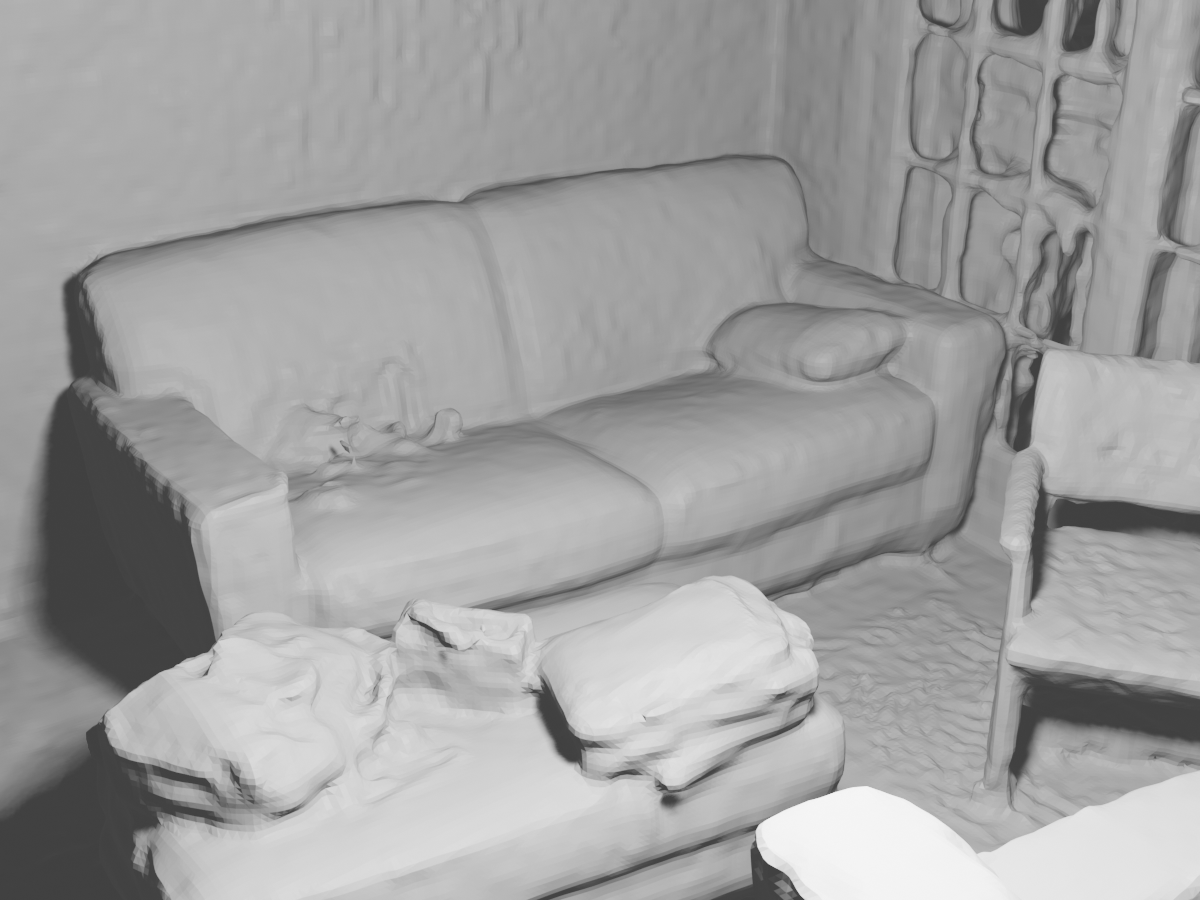}&
        \includegraphics[width=\linewidth]{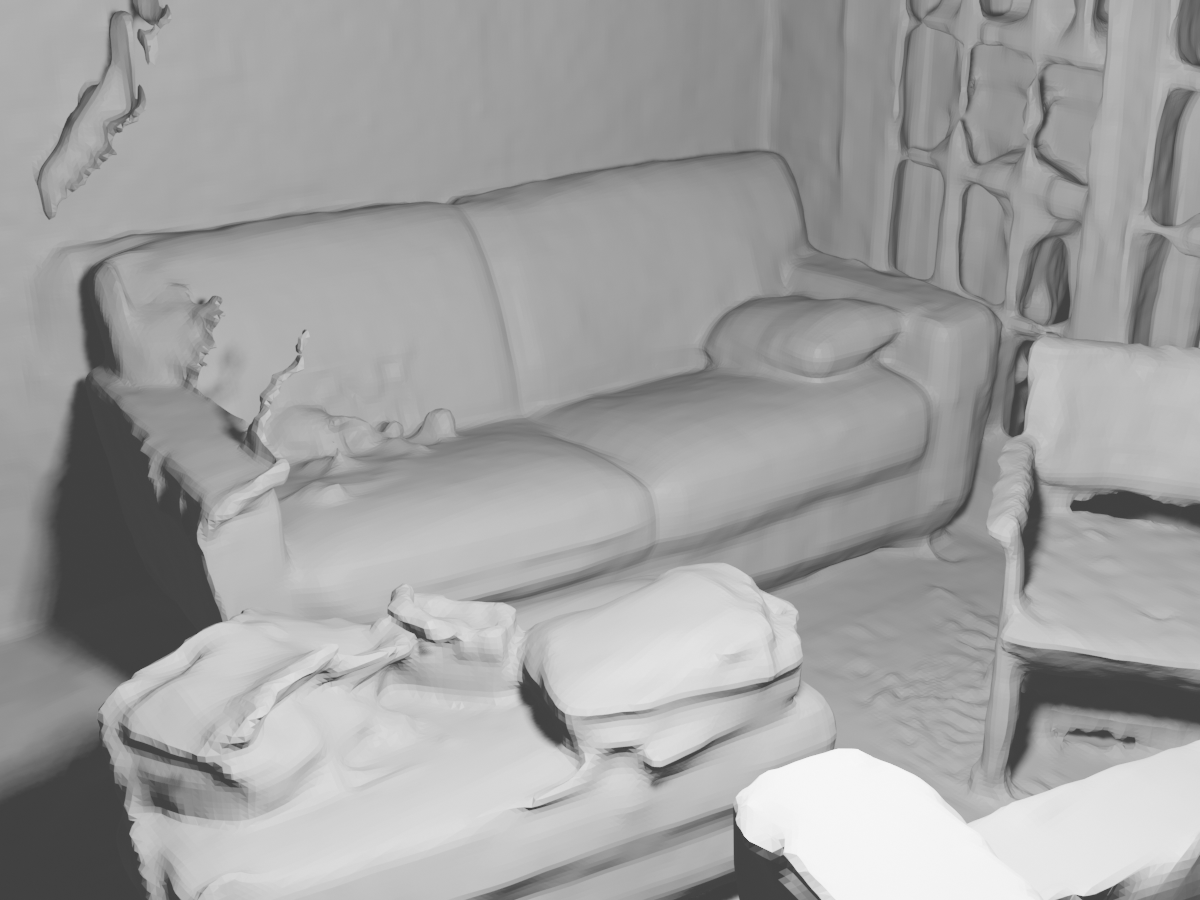}\\
        \includegraphics[width=\linewidth]{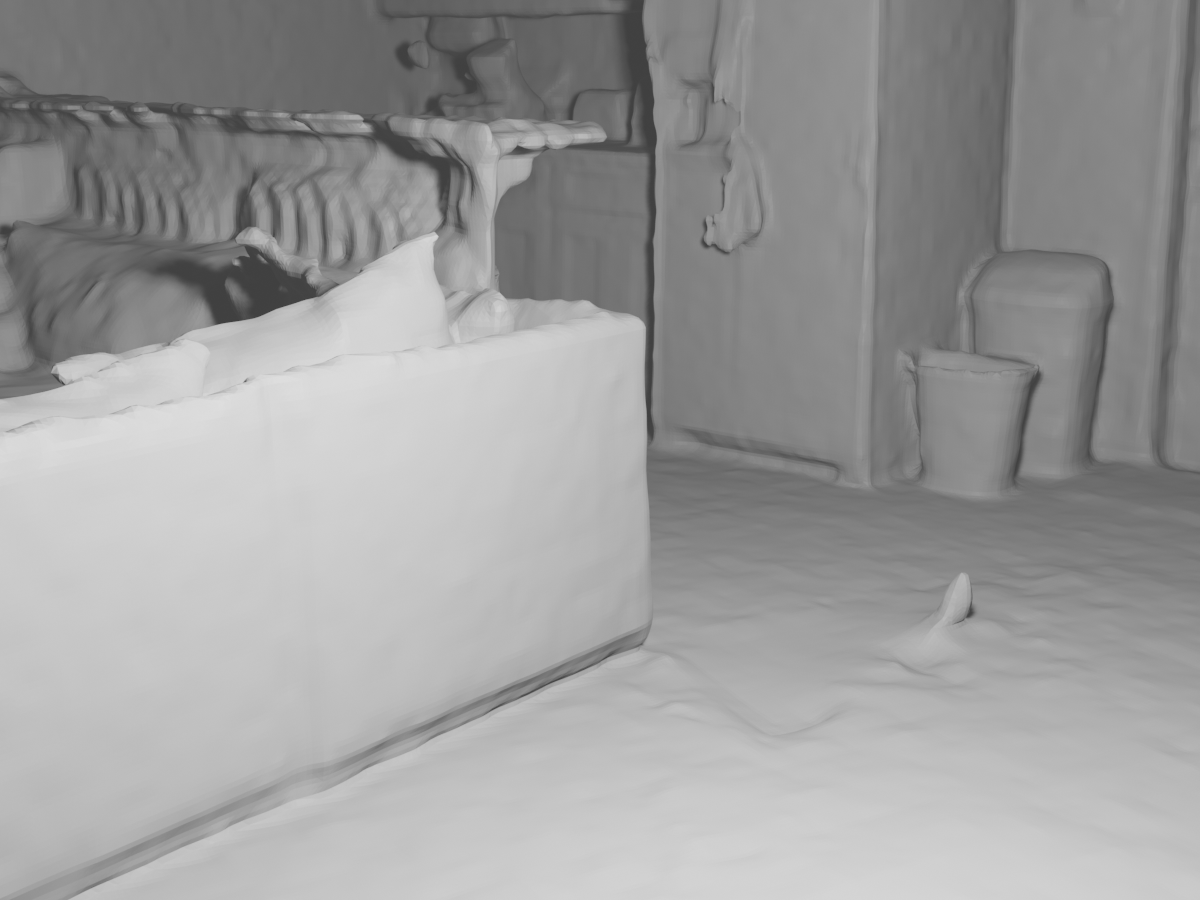}&
        \includegraphics[width=\linewidth]{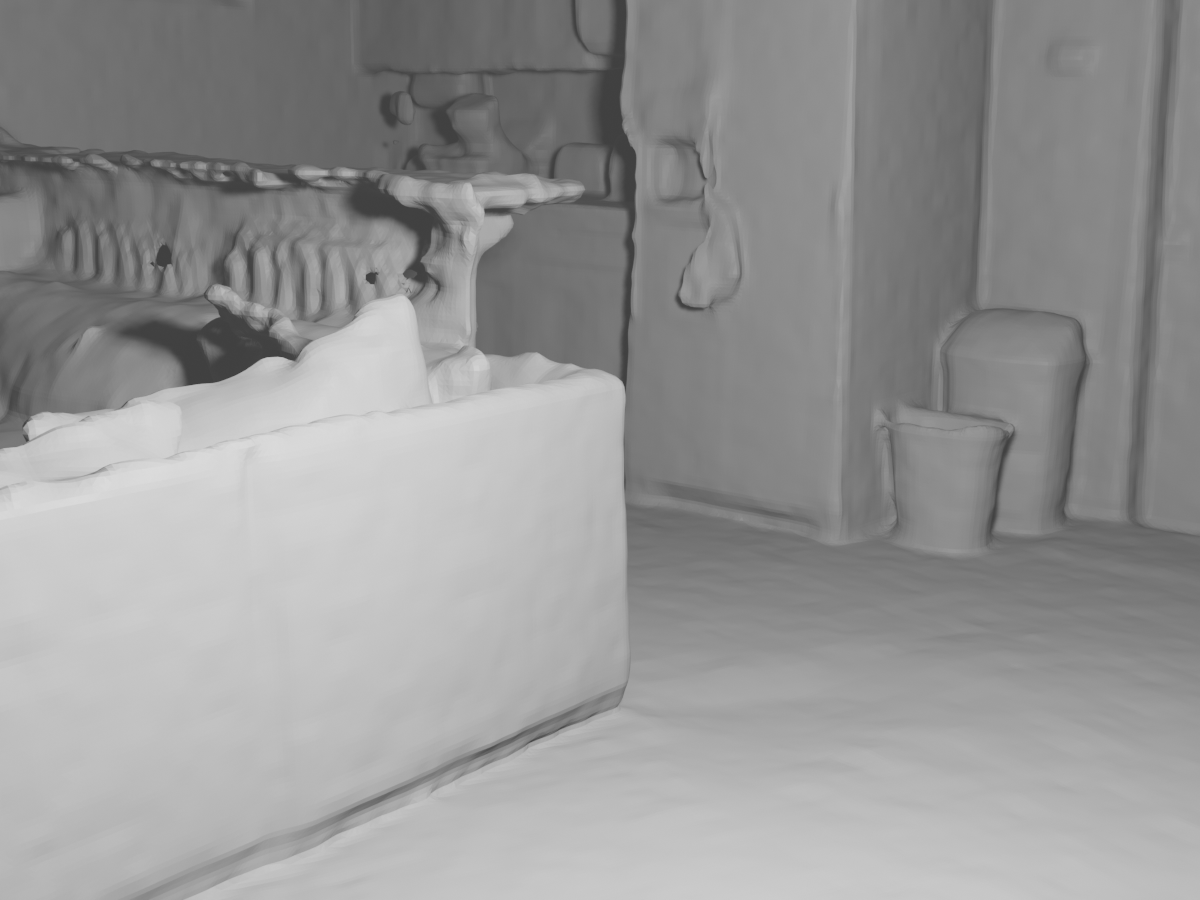}&
        \includegraphics[width=\linewidth]{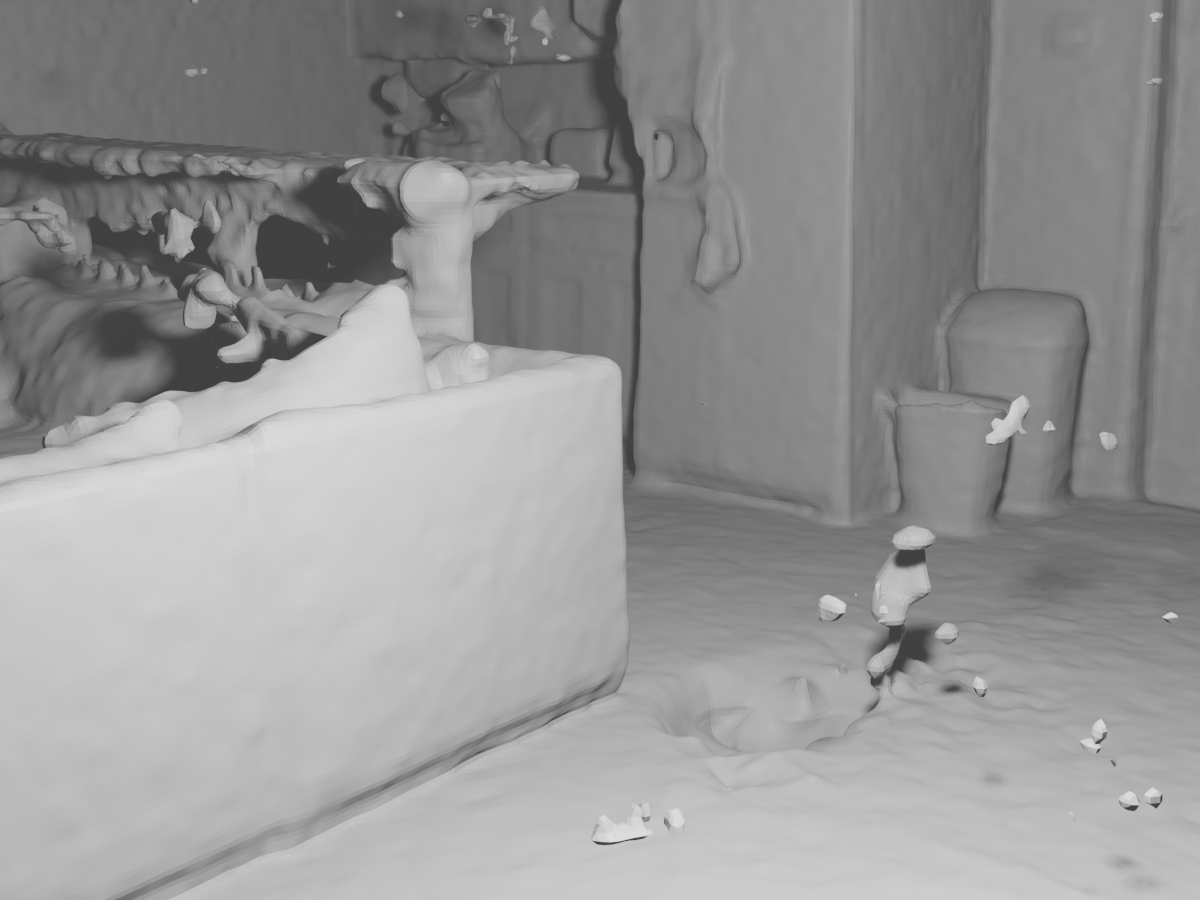}&
        \includegraphics[width=\linewidth]{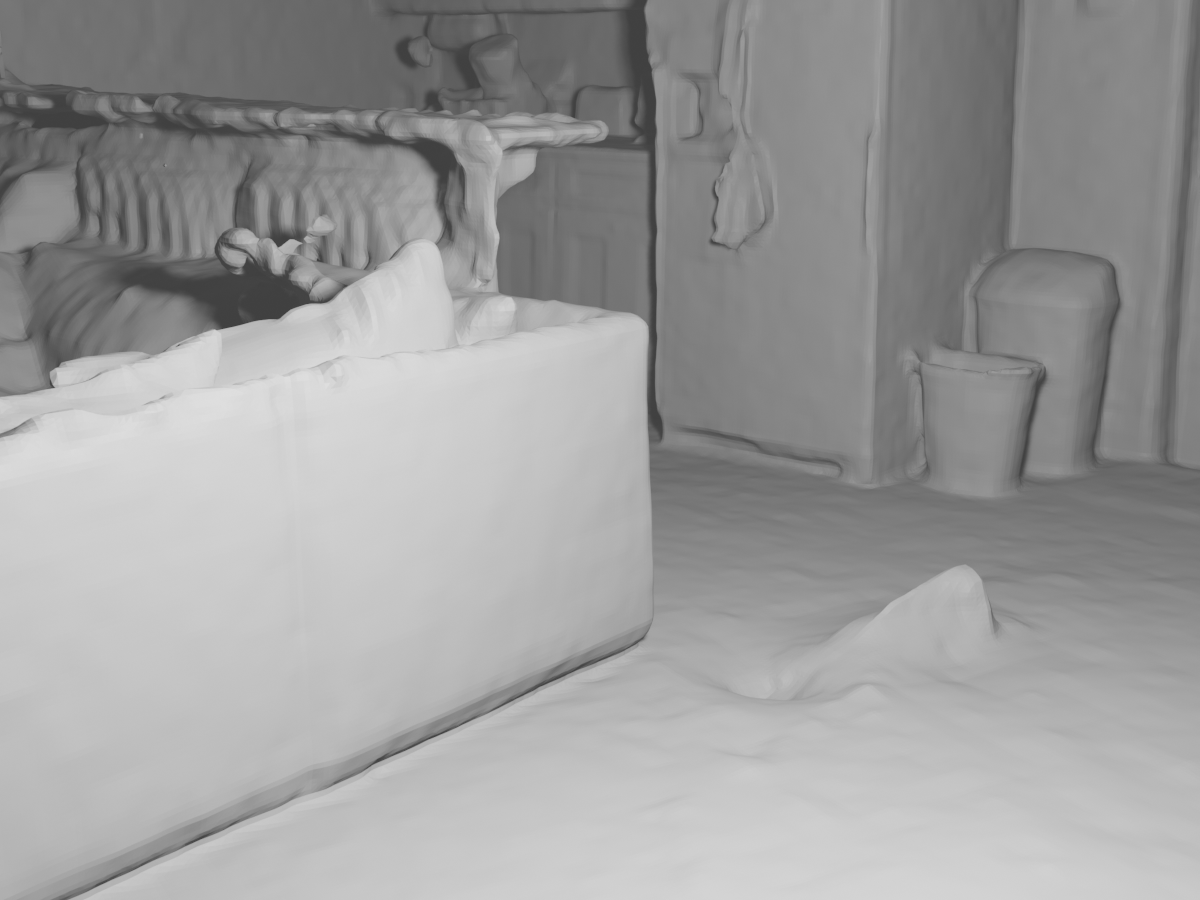}\\
         Ours &
         Ours w/o RGB loss &
         Ours w/o smoothness prior &
         Ours w/o pose optim.
    \end{tabularx}
    \vspace{-2mm}
    \caption{We evaluate our method without different loss components. We observe better reconstruction of fine details with RGB loss which, however, without smoothness prior results in noisy surfaces. Similarly to \cite{azinovic2022neural}, we observe improved surface consistency with pose refinement. \label{fig:ablation}}
    \vspace{-3mm}
\end{figure*}

\section{Results}


\subsection{Experimental Setup}

\noindent{\bf Datasets.} We consider two main datasets for evaluation. We qualitatively evaluate our reconstruction quality on real sequences from ScanNet~\cite{dai2017scannet} showing our approach is able to achieve complete and smooth reconstructions. We also quantitatively evaluate our method on 10 synthetic datasets following the same evaluation protocol as in~\cite{azinovic2022neural}.

\noindent{\bf Metrics.} We measure accuracy, completion, chamfer $\ell_1$ distance, normal consistency and F-score for reconstruction quality. As for pose accuracy we measure translation and rotation error. The metrics are computed between point clouds sampled at a density of 1 point per $\text{cm}^2$. F-score is computed using athreshold of $5\text{cm}$.

\noindent{\bf Baselines.} We consider NeuralRGB-D Surface Reconstruction~\cite{azinovic2022neural} (NeuralRGB-D) as our main baseline, and also compare quantitatively against some other state-of-the-art traditional as well as learning-based RGB-D reconstruction methods: BundleFusion~\cite{dai2017bundlefusion}, COLMAP with Poisson surface reconstruction~\cite{schoenberger2016mvs, schoenberger2016vote, Schonberger_2016_CVPR},  RoutedFusion~\cite{Weder_2020_CVPR}, Convolutional Occupancy Networks~\cite{Peng2020ECCV} and SIREN~\cite{sitzmann2019siren}.

\begin{table}[tp]
    \centering
	\resizebox{\linewidth}{!}{
    \begin{tabular}{lccccc}
        \toprule
        \textbf{Method}  & \textbf{Acc} $\downarrow$ & \textbf{Com} $\downarrow$ & \textbf{C-$\ell_1$} $\downarrow$ & \textbf{NC} $\uparrow$  & \textbf{F-score} $\uparrow$ \\
        \midrule
        BundleFusion   & 0.0191   & 0.0581  & 0.0386 & 0.9027 & 0.8439 \\
        RoutedFusion   & 0.0223   & 0.0364  & 0.0293 & 0.8765 & 0.8736 \\
        COLMAP   & 0.0271   & 0.0322  & 0.0296 & 0.9134 & 0.8744 \\
        ConvOccNets   & 0.0498   & 0.0524  & 0.0511 & 0.8607 & 0.6822 \\
        SIREN   & 0.0229   & 0.0412  & 0.0320 & 0.9049 & 0.8515 \\
        \midrule
        Neural RGBD   &  \textbf{0.0151}  & 0.0197  &	\textbf{0.0174} & 0.9316 & \textbf{0.9635} \\
        Ours  & 0.0158   & \textbf{0.0195}  &  0.0177 & \textbf{0.9317} & 0.9591 \\
        \bottomrule
    \end{tabular}
    }
    \vspace{-2mm}
    \caption{Quantitative evaluation of the reconstruction quality on a dataset of 10 synthetic scenes using the protocol established in~\cite{azinovic2022neural}. We compare with a number of baselines and with NeuralRGB-D~\cite{azinovic2022neural}. GO-Surf is on par in terms of performance but is significantly faster.
    }
    \vspace{-2mm}
    \label{tab:quantitative}
\end{table}

\noindent{\bf Implementation Details.} We run our reconstruction method on a desktop PC with a 3.60GHz Intel i7-9700K CPU and an NVIDIA RTX-2080Ti GPU. For all our experiments, we sample $M=6144$ rays per batch, and $N_c=96$ coarse samples and $N_f=36$ fine samples along each ray. Our method is implemented in Pytorch using the ADAM optimizer with learning rate of $1 \times 10^{-3}$, $1 \times 10^{-2}$ and $5 \times 10^{-4}$ for MLP decoders, feature grids and camera poses. We set the loss weights to $\lambda_{rgb} = 10.0$, $\lambda_{depth} = 1.0$, $\lambda_{sdf} = 10.0$, $\lambda_{fs} = 1.0$, $\lambda_{eik} = 1.0$ and $\lambda_{smooth} = 1.0$. We use $3\text{cm}$, $6\text{cm}$, $24\text{cm}$ and $96\text{cm}$ voxel sizes for each level. Camera pose is parameterised as translation vector $\mathbf{t} \in \mathbb{R}^3$ and Lie Algebra $\mathbf{\nu} \in \mathfrak{so}(3)$. We run the optimisation for $10\text{k}$ for all scenes. The optimisation process takes $15$ to $45$ minutes depending on the scene size. 

\begin{table}[tp]
	\resizebox{\linewidth}{!}{%
    \centering
    \begin{tabular}{lcccc}
        \toprule
        \textbf{Method}  & \textbf{Trans. error (meters)} $\downarrow$  & \textbf{Rot. error (degrees)} $\downarrow$   \\
        \midrule
        BundleFusion     & 0.033  & 0.571 \\
        COLMAP           & 0.038  & 0.692 \\
        Neural RGBD      & 0.023  & 0.146 \\
        \midrule
        Ours             & \textbf{0.014} & \textbf{0.143} \\
        \bottomrule
    \end{tabular}
    }
    \vspace{-2mm}
    \caption{
    We evaluate the average translation and rotation errors of the estimated camera poses. Our method further improves camera poses from BundleFusion initialization and outperforms NeuralRGB-D which adopts similar pose refinement scheme.
    }
\vspace{-2mm}
\label{tab:quantitative_poses}
\end{table}

\subsection{Reconstruction Quality}

\noindent{\bf Qualitative Evaluation on ScanNet~\cite{dai2017scannet}.}
We test our method on real-world sequences from the ScanNet RGB-D dataset. Depth image measurements are often noisy and miss valid depth values in areas such as glass and thin structures like chair legs. We show that our RGB loss and SDF regularisation can mitigate these effects significantly. We also jointly refine the camera poses provided in ScanNet.

Fig.~\ref{fig:scannet_result} shows comparisons of our method with  reconstructions obtained via BundleFusion~\cite{dai2017bundlefusion}, and NeuralRGB-D~\cite{azinovic2022neural}. Our method produces more complete and smoother reconstruction results. For missing depth on  areas such as TV screens ($2\text{nd}$, $4\text{th}$ and $5\text{th}$ row), our method demonstrates strong  hole-filling ability and produces high-quality reconstruction. Our method also captures thin structures (bike in the $2\text{nd}$ row and chair leg in the $5\text{th}$ row) better than our baselines. More reconstruction results can be found in the supplementary material.

\noindent{\bf Quantitative Evaluation on Synthetic Sequences.} We perform a quantatitive evaluation of our method on a synthetic dataset of 10 sequences~\cite{azinovic2022neural}. Noise and artifacts are added to the rendered ground-truth depth images to simulate a real depth sensor. The camera poses obtained with BundleFusion~\cite{dai2017bundlefusion} are used as initialization for reconstruction, and are optimized along with the feature grids and decoders. We run marching cubes at the resolution of 1cm to extract meshes. For fair comparison, we also run BundleFusion~\cite{dai2017bundlefusion} and RoutedFusion at the same resolution. Regions that are not observed in any camera views are culled before evaluation. 

As shown in Tab.~\ref{tab:quantitative}, our method performs on par with NeuralRGB-D, and outperforms all the other baselines by a large margin. Specifically, we achieve better Completion and Normal Consistency which shows the benefit of our proposed smoothness prior in terms of hole-filling and smoothness. In Tab.~\ref{tab:quantitative_poses} we evaluate the pose estimation accuracy of our method. We take BundleFusion as initial camera poses and refine them along with other optimisable parameters. Results show that our pose-refinement  significantly improved the initial poses and also outperforms NeuralRGB-D in terms of both translation and rotation error.

\begin{table}[tp]
	\resizebox{\linewidth}{!}{%
    \centering
    \begin{tabular}{lccc}
        \toprule
         & scene 0000  & scene 0012 & morning apartment\\
        \midrule
        dimension & $8.8 \times 9.1 \times 3.4$ & $5.8 \times 5.7 \times 2.9$ & $3.5 \times 4.0 \times 2.3$ \\
        voxel dim & $321 \times 321 \times 129$ & $225 \times 225 \times 129$ & $129 \times 161 \times 97$ \\
        runtime  & $44 \text{ min}$  & $31 \text{ min}$ & $19 \text{ min}$ \\
        model size      & $268 \text{ MB}$ & $132 \text{ MB}$ &  $41 \text{ MB}$ \\
        num params & $66.5$ M & $32.7$ M & $10.1$ M \\
        \bottomrule
    \end{tabular}
    }
    \vspace{-1.5mm}
    \caption{
    Performance Analysis on 3 scenes from ScanNet~\cite{dai2017scannet} and synthetic dataset. Our method requires much less runtime than NeuralRGB-D~\cite{azinovic2022neural} at the cost of larger model size.
    }
    \vspace{-3mm}
\label{tab:performance}
\end{table}

\subsection{Performance Analysis}
\noindent{\bf Runtime.} GO-Surf achieves high-quality reconstructions in $15$ to $45$ minutes, while NeuralRGB-D~\cite{azinovic2022neural} takes $15$ to $25$ hours on the same hardware. Tab.~\ref{tab:performance} shows our method's runtime on $3$ example scenes with various scales from ScanNet and synthetic datasets. 

\noindent{\bf Memory Footprint.} We report memory usage for storing our model in Tab.~\ref{tab:performance}. Our model requires hundreds of MB to store the full feature grid and scales up rapidly as the scene size increases. This is one of our major limitations and we leave it as  future work to optimize memory footprint. In contrast, our method runs much faster than the low-memory-footprint implicit representations while  achieving on par reconstruction quality and still requires fewer parameters than traditional methods that use explicit voxel grid.

\subsection{Ablation Studies}
We conduct ablation studies to show the effect of individual building blocks and justify our design choices.

\noindent{\bf Effect of RGB Term.}
We compare our full model against our model without the RGB loss term. Results show that the RGB loss term enables reconstruction of high frequency details that could not be recovered using solely the depth supervision while still preserving overall smoothness (Fig.~\ref{fig:ablation}).

\noindent{\bf Effect of Smoothness Term}
We also verify the effectiveness of the smoothness prior term, which has a twofold role. Firstly, it completes the unobserved parts of the scene, for example holes on the floor. Secondly, it regularises the RGB and depth losses that would otherwise overfit and produce noisy surfaces (Fig.~\ref{fig:ablation}).

\noindent{\bf Effect of Pose Refinement}
Similarly to \cite{azinovic2022neural}, we observe that joint optimisation of the model and camera poses corrects the imperfect trajectory estimate from SLAM system and leads to more consistent surface reconstruction (Fig.~\ref{fig:ablation}).


\section{Conclusion}
We presented GO-Surf, a novel approach to surface reconstruction from a sequence of RGB-D images. We achieved a level of smoothness and hole filling on-par with MLP-based approaches while reducing the training time by an order of magnitude. Our system produces accurate and complete meshes of indoor scenes.

\paragraph{Limitations and Future Work.}
The biggest limitation of our system is the memory consumption which at the moment scales cubically with the scene dimensions. This is a well-known problem of voxel-like architectures and various solutions have been proposed to mitigate it. Voxel hashing \cite{niessner2013hashing} or octree-based sparsification would significantly reduce the memory footprint of our system and we intend to explore this direction in future work. 
We also recognize the high complexity of our loss term which makes it difficult to fully explore the hyperparameter space. Specifically, we would like to better understand the relation between SDF, free-space and depth loss and preferably merge them into one loss term. 
Finally, our method currently overfits to a single scene. We are interested in learning the geometry priors on large datasets to use them at inference time.

\vspace{-0.1cm}

\section*{Acknowledgements}

\vspace{-0.1cm}

Research presented here has been supported by the UCL Centre for Doctoral Training in Foundational AI under UKRI grant number EP/S021566/1. TB was supported from a sponsored research award by Cisco Research. We also thank Dejan Azinovic for providing additional details and results of NeuralRGBD.

{\small
\bibliographystyle{ieee_fullname}
\bibliography{main}

\begin{thebibliography}{10}\itemsep=-1pt

\bibitem{azinovic2022neural}
Dejan Azinovi{\'c}, Ricardo Martin-Brualla, Dan~B Goldman, Matthias
  Nie{\ss}ner, and Justus Thies.
\newblock Neural rgb-d surface reconstruction.
\newblock In {\em Proceedings of the IEEE/CVF Conference on Computer Vision and
  Pattern Recognition (CVPR)}, June 2022.

\bibitem{bozic2021transformerfusion}
Aljaz Bozic, Pablo Palafox, Justus Thies, Angela Dai, and Matthias Nie{\ss}ner.
\newblock Transformerfusion: Monocular rgb scene reconstruction using
  transformers.
\newblock {\em Advances in Neural Information Processing Systems}, 34, 2021.

\bibitem{chen2022tensorf}
Anpei Chen, Zexiang Xu, Andreas Geiger, Jingyi Yu, and Hao Su.
\newblock Tensorf: Tensorial radiance fields.
\newblock {\em arXiv preprint arXiv:2203.09517}, 2022.

\bibitem{chibane2020implicit}
Julian Chibane, Thiemo Alldieck, and Gerard Pons-Moll.
\newblock Implicit functions in feature space for 3d shape reconstruction and
  completion.
\newblock In {\em Proceedings of the IEEE/CVF Conference on Computer Vision and
  Pattern Recognition}, pages 6970--6981, 2020.

\bibitem{dai2017scannet}
Angela Dai, Angel~X. Chang, Manolis Savva, Maciej Halber, Thomas Funkhouser,
  and Matthias Nie{\ss}ner.
\newblock Scannet: Richly-annotated 3d reconstructions of indoor scenes.
\newblock In {\em Proc. Computer Vision and Pattern Recognition (CVPR), IEEE},
  2017.

\bibitem{dai2017bundlefusion}
Angela Dai, Matthias Nie{\ss}ner, Michael Zollh{\"o}fer, Shahram Izadi, and
  Christian Theobalt.
\newblock Bundlefusion: Real-time globally consistent 3d reconstruction using
  on-the-fly surface reintegration.
\newblock {\em ACM Transactions on Graphics (TOG)}, 36(4):76a, 2017.

\bibitem{icml2020_2086}
Amos Gropp, Lior Yariv, Niv Haim, Matan Atzmon, and Yaron Lipman.
\newblock Implicit geometric regularization for learning shapes.
\newblock In {\em International Conference on Machine Learning}, pages
  3789--3799. PMLR, 2020.

\bibitem{karnewar2022relu}
Animesh Karnewar, Tobias Ritschel, Oliver Wang, and Niloy~J Mitra.
\newblock Relu fields: The little non-linearity that could.
\newblock {\em arXiv preprint arXiv:2205.10824}, 2022.

\bibitem{kundu2022panoptic}
Abhijit Kundu, Kyle Genova, Xiaoqi Yin, Alireza Fathi, Caroline Pantofaru,
  Leonidas Guibas, Andrea Tagliasacchi, Frank Dellaert, and Thomas Funkhouser.
\newblock Panoptic neural fields: A semantic object-aware neural scene
  representation.
\newblock {\em arXiv preprint arXiv:2205.04334}, 2022.

\bibitem{liu2020neural}
Lingjie Liu, Jiatao Gu, Kyaw Zaw~Lin, Tat-Seng Chua, and Christian Theobalt.
\newblock Neural sparse voxel fields.
\newblock {\em Advances in Neural Information Processing Systems},
  33:15651--15663, 2020.

\bibitem{occnet}
Lars Mescheder, Michael Oechsle, Michael Niemeyer, Sebastian Nowozin, and
  Andreas Geiger.
\newblock Occupancy networks: Learning 3d reconstruction in function space.
\newblock In {\em Proceedings IEEE Conf. on Computer Vision and Pattern
  Recognition (CVPR)}, 2019.

\bibitem{mildenhall2020nerf}
Ben Mildenhall, Pratul~P. Srinivasan, Matthew Tancik, Jonathan~T. Barron, Ravi
  Ramamoorthi, and Ren Ng.
\newblock Nerf: Representing scenes as neural radiance fields for view
  synthesis.
\newblock In {\em ECCV}, 2020.

\bibitem{muller2022instant}
Thomas M{\"u}ller, Alex Evans, Christoph Schied, and Alexander Keller.
\newblock Instant neural graphics primitives with a multiresolution hash
  encoding.
\newblock {\em arXiv preprint arXiv:2201.05989}, 2022.

\bibitem{murez2020atlas}
Zak Murez, Tarrence~van As, James Bartolozzi, Ayan Sinha, Vijay Badrinarayanan,
  and Andrew Rabinovich.
\newblock Atlas: End-to-end 3d scene reconstruction from posed images.
\newblock In {\em European Conference on Computer Vision}, pages 414--431.
  Springer, 2020.

\bibitem{newcombe2011kinectfusion}
Richard~A Newcombe, Shahram Izadi, Otmar Hilliges, David Molyneaux, David Kim,
  Andrew~J Davison, Pushmeet Kohi, Jamie Shotton, Steve Hodges, and Andrew
  Fitzgibbon.
\newblock Kinectfusion: Real-time dense surface mapping and tracking.
\newblock In {\em 2011 10th IEEE International Symposium on Mixed and Augmented
  Reality (ISMAR)}, pages 127--136. IEEE, 2011.

\bibitem{niemeyer2020differentiable}
Michael Niemeyer, Lars Mescheder, Michael Oechsle, and Andreas Geiger.
\newblock Differentiable volumetric rendering: Learning implicit 3d
  representations without 3d supervision.
\newblock In {\em Proceedings of the IEEE/CVF Conference on Computer Vision and
  Pattern Recognition}, pages 3504--3515, 2020.

\bibitem{niessner2013real}
Matthias Nie{\ss}ner, Michael Zollh{\"o}fer, Shahram Izadi, and Marc
  Stamminger.
\newblock Real-time 3d reconstruction at scale using voxel hashing.
\newblock {\em ACM Transactions on Graphics (ToG)}, 32(6):1--11, 2013.

\bibitem{niessner2013hashing}
M. Nie{\ss}ner, M. Zollh\"ofer, S. Izadi, and M. Stamminger.
\newblock Real-time 3d reconstruction at scale using voxel hashing.
\newblock {\em ACM Transactions on Graphics (TOG)}, 2013.

\bibitem{oechsle2019texture}
Michael Oechsle, Lars Mescheder, Michael Niemeyer, Thilo Strauss, and Andreas
  Geiger.
\newblock Texture fields: Learning texture representations in function space.
\newblock In {\em Proceedings of the IEEE/CVF International Conference on
  Computer Vision}, pages 4531--4540, 2019.

\bibitem{Ortiz:etal:iSDF2022}
Joseph Ortiz, Alexander Clegg, Jing Dong, Edgar Sucar, David Novotny, Michael
  Zollhoefer, and Mustafa Mukadam.
\newblock isdf: Real-time neural signed distance fields for robot perception.
\newblock {\em arXiv preprint arXiv:2204.02296}, 2022.

\bibitem{Park_2019_CVPR}
Jeong~Joon Park, Peter Florence, Julian Straub, Richard Newcombe, and Steven
  Lovegrove.
\newblock Deepsdf: Learning continuous signed distance functions for shape
  representation.
\newblock In {\em Proceedings of the IEEE/CVF Conference on Computer Vision and
  Pattern Recognition}, pages 165--174, 2019.

\bibitem{Peng2020ECCV}
Songyou Peng, Michael Niemeyer, Lars Mescheder, Marc Pollefeys, and Andreas
  Geiger.
\newblock Convolutional occupancy networks.
\newblock In {\em European Conference on Computer Vision (ECCV)}, Cham, Aug.
  2020. Springer International Publishing.

\bibitem{Schonberger_2016_CVPR}
Johannes~L. Schonberger and Jan-Michael Frahm.
\newblock Structure-from-motion revisited.
\newblock In {\em Proceedings of the IEEE Conference on Computer Vision and
  Pattern Recognition (CVPR)}, June 2016.

\bibitem{schoenberger2016vote}
Johannes~Lutz Sch\"{o}nberger, True Price, Torsten Sattler, Jan-Michael Frahm,
  and Marc Pollefeys.
\newblock A vote-and-verify strategy for fast spatial verification in image
  retrieval.
\newblock In {\em Asian Conference on Computer Vision (ACCV)}, 2016.

\bibitem{schoenberger2016mvs}
Johannes~Lutz Sch\"{o}nberger, Enliang Zheng, Marc Pollefeys, and Jan-Michael
  Frahm.
\newblock Pixelwise view selection for unstructured multi-view stereo.
\newblock In {\em European Conference on Computer Vision (ECCV)}, 2016.

\bibitem{sitzmann2019siren}
Vincent Sitzmann, Julien~N.P. Martel, Alexander~W. Bergman, David~B. Lindell,
  and Gordon Wetzstein.
\newblock Implicit neural representations with periodic activation functions.
\newblock In {\em arXiv}, 2020.

\bibitem{srns}
Vincent Sitzmann, Michael Zollh{\"o}fer, and Gordon Wetzstein.
\newblock Scene representation networks: Continuous 3d-structure-aware neural
  scene representations.
\newblock In {\em Advances in Neural Information Processing Systems}, 2019.

\bibitem{stier2021vortx}
Noah Stier, Alexander Rich, Pradeep Sen, and Tobias H{\"o}llerer.
\newblock {VoRTX}: Volumetric 3d reconstruction with transformers for voxelwise
  view selection and fusion.
\newblock In {\em 2021 International Conference on 3D Vision (3DV)}, pages
  320--330. IEEE, 2021.

\bibitem{Sucar:etal:ICCV2021}
Edgar Sucar, Shikun Liu, Joseph Ortiz, and Andrew~J Davison.
\newblock imap: Implicit mapping and positioning in real-time.
\newblock In {\em Proceedings of the IEEE/CVF International Conference on
  Computer Vision}, pages 6229--6238, 2021.

\bibitem{sun2021direct}
Cheng Sun, Min Sun, and Hwann-Tzong Chen.
\newblock Direct voxel grid optimization: Super-fast convergence for radiance
  fields reconstruction.
\newblock {\em arXiv preprint arXiv:2111.11215}, 2021.

\bibitem{sun2021neuralrecon}
Jiaming Sun, Yiming Xie, Linghao Chen, Xiaowei Zhou, and Hujun Bao.
\newblock Neuralrecon: Real-time coherent 3d reconstruction from monocular
  video.
\newblock In {\em Proceedings of the IEEE/CVF Conference on Computer Vision and
  Pattern Recognition}, pages 15598--15607, 2021.

\bibitem{takikawa2021nglod}
Towaki Takikawa, Joey Litalien, Kangxue Yin, Karsten Kreis, Charles Loop, Derek
  Nowrouzezahrai, Alec Jacobson, Morgan McGuire, and Sanja Fidler.
\newblock Neural geometric level of detail: Real-time rendering with implicit
  {3D} shapes.
\newblock 2021.

\bibitem{wang2021neus}
Peng Wang, Lingjie Liu, Yuan Liu, Christian Theobalt, Taku Komura, and Wenping
  Wang.
\newblock Neus: Learning neural implicit surfaces by volume rendering for
  multi-view reconstruction.
\newblock {\em NeurIPS}, 2021.

\bibitem{Weder_2020_CVPR}
Silvan Weder, Johannes~L. Sch\"onberger, Marc Pollefeys, and Martin~R. Oswald.
\newblock Routedfusion: Learning real-time depth map fusion.
\newblock In {\em IEEE/CVF Conference on Computer Vision and Pattern
  Recognition (CVPR)}, June 2020.

\bibitem{yariv2021volume}
Lior Yariv, Jiatao Gu, Yoni Kasten, and Yaron Lipman.
\newblock Volume rendering of neural implicit surfaces.
\newblock In {\em Thirty-Fifth Conference on Neural Information Processing
  Systems}, 2021.

\bibitem{yariv2020multiview}
Lior Yariv, Yoni Kasten, Dror Moran, Meirav Galun, Matan Atzmon, Basri Ronen,
  and Yaron Lipman.
\newblock Multiview neural surface reconstruction by disentangling geometry and
  appearance.
\newblock {\em Advances in Neural Information Processing Systems},
  33:2492--2502, 2020.

\bibitem{yu2021plenoxels}
Alex Yu, Sara Fridovich-Keil, Matthew Tancik, Qinhong Chen, Benjamin Recht, and
  Angjoo Kanazawa.
\newblock Plenoxels: Radiance fields without neural networks.
\newblock {\em arXiv preprint arXiv:2112.05131}, 2021.

\bibitem{yu2021plenoctrees}
Alex Yu, Ruilong Li, Matthew Tancik, Hao Li, Ren Ng, and Angjoo Kanazawa.
\newblock Plenoctrees for real-time rendering of neural radiance fields.
\newblock In {\em Proceedings of the IEEE/CVF International Conference on
  Computer Vision}, pages 5752--5761, 2021.

\bibitem{zhang2022nerfusion}
Xiaoshuai Zhang, Sai Bi, Kalyan Sunkavalli, Hao Su, and Zexiang Xu.
\newblock Nerfusion: Fusing radiance fields for large-scale scene
  reconstruction.
\newblock {\em CVPR}, 2022.

\bibitem{zhi2021place}
Shuaifeng Zhi, Tristan Laidlow, Stefan Leutenegger, and Andrew~J Davison.
\newblock In-place scene labelling and understanding with implicit scene
  representation.
\newblock In {\em Proceedings of the IEEE/CVF International Conference on
  Computer Vision}, pages 15838--15847, 2021.

\bibitem{Zhu2022CVPR_niceslam}
Zihan Zhu, Songyou Peng, Viktor Larsson, Weiwei Xu, Hujun Bao, Zhaopeng Cui,
  Martin~R. Oswald, and Marc Pollefeys.
\newblock Nice-slam: Neural implicit scalable encoding for slam.
\newblock In {\em Proceedings of the IEEE/CVF Conference on Computer Vision and
  Pattern Recognition (CVPR)}, 2022.

\end{thebibliography}


\begin{thebibliography}{1}\itemsep=-1pt

\bibitem{azinovic2022neural}
Dejan Azinovi{\'c}, Ricardo Martin-Brualla, Dan~B Goldman, Matthias
  Nie{\ss}ner, and Justus Thies.
\newblock Neural rgb-d surface reconstruction.
\newblock In {\em Proceedings of the IEEE/CVF Conference on Computer Vision and
  Pattern Recognition (CVPR)}, June 2022.

\bibitem{dai2017scannet}
Angela Dai, Angel~X. Chang, Manolis Savva, Maciej Halber, Thomas Funkhouser,
  and Matthias Nie{\ss}ner.
\newblock Scannet: Richly-annotated 3d reconstructions of indoor scenes.
\newblock In {\em Proc. Computer Vision and Pattern Recognition (CVPR), IEEE},
  2017.

\bibitem{dai2017bundlefusion}
Angela Dai, Matthias Nie{\ss}ner, Michael Zollh{\"o}fer, Shahram Izadi, and
  Christian Theobalt.
\newblock Bundlefusion: Real-time globally consistent 3d reconstruction using
  on-the-fly surface reintegration.
\newblock {\em ACM Transactions on Graphics (TOG)}, 36(4):76a, 2017.

\bibitem{Peng2020ECCV}
Songyou Peng, Michael Niemeyer, Lars Mescheder, Marc Pollefeys, and Andreas
  Geiger.
\newblock Convolutional occupancy networks.
\newblock In {\em European Conference on Computer Vision (ECCV)}, Cham, Aug.
  2020. Springer International Publishing.

\bibitem{sitzmann2019siren}
Vincent Sitzmann, Julien~N.P. Martel, Alexander~W. Bergman, David~B. Lindell,
  and Gordon Wetzstein.
\newblock Implicit neural representations with periodic activation functions.
\newblock In {\em arXiv}, 2020.

\bibitem{Zhu2022CVPR_niceslam}
Zihan Zhu, Songyou Peng, Viktor Larsson, Weiwei Xu, Hujun Bao, Zhaopeng Cui,
  Martin~R. Oswald, and Marc Pollefeys.
\newblock Nice-slam: Neural implicit scalable encoding for slam.
\newblock In {\em Proceedings of the IEEE/CVF Conference on Computer Vision and
  Pattern Recognition (CVPR)}, 2022.

\end{thebibliography}
}

\end{document}


\title{Supplementary Material \\GO-Surf: Neural Feature Grid Optimization for \\Fast, High-Fidelity RGB-D Surface Reconstruction}

\author[*]{Jingwen Wang}
\author[*]{Tymoteusz Bleja}
\author[ ]{Lourdes Agapito}
\affil[ ]{Department of Computer Science, University College London}

\maketitle
\thispagestyle{empty}
\renewcommand{\paragraph}[1]{\medskip\noindent\textbf{#1}}

\renewcommand*{\thefootnote}{\fnsymbol{footnote}}
\footnotetext{* The first two authors contributed equally.}
\renewcommand*{\thefootnote}{\arabic{footnote}}
\setcounter{footnote}{0}


\section{Second-order Grid Sampler}
In this section we provide additional details on our implementation of grid sampler that allows double back-propagation in PyTorch. As described in our main paper, thepredicted SDF value $\phi$ is a function of 3D coordinates of the query point $\mathbf{x}$, feature vectors in feature grid $\theta$, and geometry network parameters $\omega$:
\begin{equation}  \label{sdf_full}
    \phi(\mathbf{x}, \theta, \omega) = f_{\omega}(\mathbf{z}(\theta, \mathbf{x}))
\end{equation}
where $\mathbf{z} = \mathcal{V}_{\theta}(\mathbf{x}) \in \mathbb{R}^C$ is the tri-linearly interpolated feature vector at the query point $\mathbf{x}.$ The first-order gradient of the SDF w.r.t. 3D query coordinates is:
\begin{eqnarray}\label{sdf_first_grad}
    \frac{\partial \phi}{\partial \mathbf{x}} & = & \frac{\partial f_{\omega}(\mathbf{z})}{\partial \mathbf{z}} \frac{\partial \mathbf{z}}{\partial \mathbf{x}} \nonumber \\
    & = & \sum_{i=1}^C \frac{\partial f_{\omega}(\mathbf{z})}{\partial z_i} \frac{\partial z_i}{\partial \mathbf{x}}
\end{eqnarray}
Here, to avoid the usage of higher-order tensors in derivation of second-order derivatives we consider the derivatives for each individual feature dimension $z_i$, as in Eq.~\ref{sdf_first_grad}. To regularise this gradient term we need to obtain the gradient Eq.~\ref{sdf_first_grad} w.r.t. $\mathbf{x}$, $\theta$ and $\omega$ respectively, which corresponds to the first three row of the Hessian of SDF:
\begin{eqnarray}
    \frac{\partial^2 \phi}{\partial \mathbf{x}^2} & = & \sum_{i=1}^C \frac{\partial^2 f_{\omega}(\mathbf{z})}{\partial z^2_i}  \bigg( \frac{\partial z_i}{\partial \mathbf{x}} \bigg)^2 + \frac{\partial f_{\omega}(\mathbf{z})}{\partial z_i} \frac{\partial^2 z_i}{\partial \mathbf{x}^2} \label{eq:second_sdf_grad_x}\\
    \frac{\partial^2 \phi}{\partial \mathbf{x} \partial \theta} & = & \sum_{i=1}^C \frac{\partial^2 f_{\omega}(\mathbf{z})}{\partial z^2_i} \frac{\partial z_i}{\partial \mathbf{\theta}} \frac{\partial z_i}{\partial \mathbf{x}} + \frac{\partial f_{\omega}(\mathbf{z})}{\partial z_i} \frac{\partial^2 z_i}{\partial \mathbf{x} \partial \theta} \label{eq:second_sdf_grad_theta}\\
    \frac{\partial^2 \phi}{\partial \mathbf{x} \partial \omega} & = & \sum_{i=1}^C \frac{\partial^2 f_{\omega}(\mathbf{z})}{\partial z_i \partial \omega} \frac{\partial z_i}{\partial \mathbf{x}} \label{eq:second_sdf_grad_omega} 
\end{eqnarray}
We don't need to implement Eq.~\ref{eq:second_sdf_grad_omega} as Pytorch's automatic differentiation package supports double back-propagation through an MLP and Pytorch's \texttt{grid\_sampler} also has first-order gradient implementation. However Eq.~\ref{eq:second_sdf_grad_x} and ~\ref{eq:second_sdf_grad_theta} need to be implemented as they require double back-propagation through the \texttt{grid\_sampler} which is not implemented in PyTorch. 


\paragraph{Derivative Derivation.} Now we will derive the second-order derivatives of the tri-linear interpolation. More specifically, we only need two Hessian blocks: $\frac{\partial^2 z_i}{\partial \mathbf{x}^2}$ and $\frac{\partial^2 z_i}{\partial \mathbf{x} \partial \theta}$. For simplicity we derive with $1\text{-D}$ features, but it generalises to any feature dimensions as the derivation is equivalent to all dimensions. Assume we have a query point $\mathbf{x} = [x, y, z]$ and $8$ feature vectors (points) stacking in a column $\theta = [\theta_{000}, \theta_{001}, \theta_{010}, \theta_{011}, \theta_{100}, \theta_{101}, \theta_{110}, \theta_{111}]^T \in \mathbb{R}^{8}$ that correspond to the eight vertices of the voxel that encloses the point $\mathbf{x}$. Then, the tri-linearly interpolated feature $f$ at point $\mathbf{x}$ is given by:
\begin{figure}[t]
\begin{center}
\includegraphics[width=0.70\linewidth]{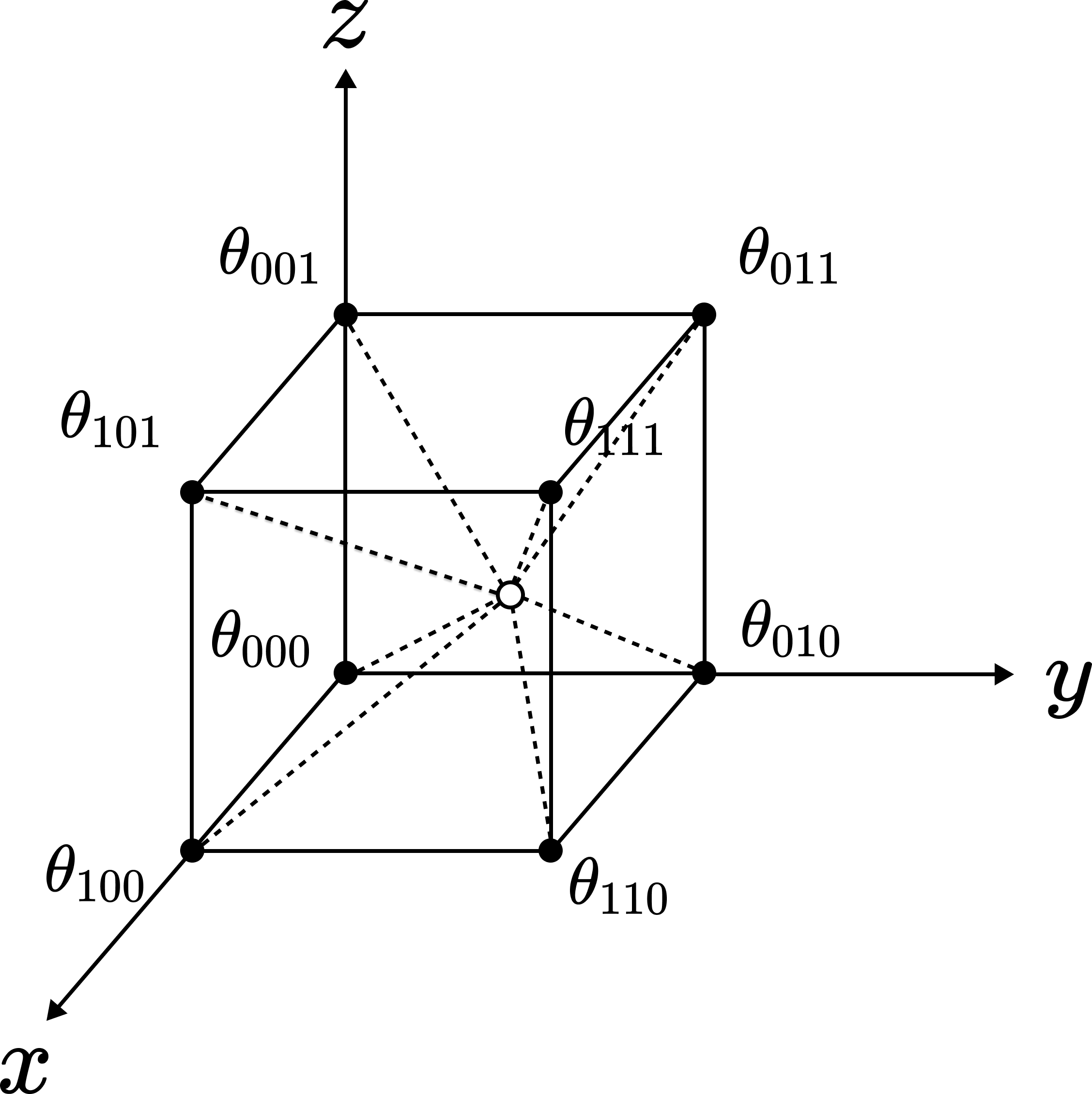}
\end{center}
  \caption{Example of trilinear interpolation in a 3D voxel.}
\label{fig:bilinear_interp}
\end{figure}

\begin{equation} \label{eq:trilinear_interp}
\begin{aligned}
    f = \theta^T \mathbf{w}(\mathbf{x})
\end{aligned}
\end{equation}
where $\mathbf{w}(\mathbf{x}) \in \mathbb{R}^8$ is the coefficient vector of the 8 vertices:
\begin{equation} \label{eq:coefficient_vector}
\begin{aligned}
    \mathbf{w}(\mathbf{x}) = \begin{bmatrix}
        (1-x)(1-y)(1-z) \\
        (1-x)(1-y)z \\
        (1-x)y(1-z) \\
        (1-x)yz \\
        x(1-y)(1-z) \\
        x(1-y)z \\
        xy(1-z) \\
        xyz \\
    \end{bmatrix}
\end{aligned}
\end{equation}
Then the Jacobian of $f$ w.r.t. $\mathbf{x}$ is given by:
\begin{equation} \label{eq:jacobian}
\begin{aligned}
    \frac{\partial f}{\partial \mathbf{x}} = \theta^T \frac{\partial \mathbf{w}(\mathbf{x})}{\partial \mathbf{x}}
\end{aligned}
\end{equation}
where the Jacobian of the coefficient vector $\mathbf{w}$ w.r.t. query point $\mathbf{x}$ is a $8 \times 3$ matrix and is given by:
\begin{eqnarray}
    \frac{\partial\mathbf{w}}{\partial\mathbf{x}} & = & \begin{bmatrix}
        \mathbf{J}_1 & \mathbf{J}_2 & \mathbf{J}_3
    \end{bmatrix} \label{eq:coeff_jacobian} \\
    \mathbf{J}_1 & = & \begin{bmatrix}
        -(1-y)(1-z) \\
        -(1-y)z \\
        -y(1-z) \\
        -yz \\
        (1-y)(1-z) \\
        (1-y)z \\
        y(1-z) \\
        yz \\
    \end{bmatrix} \label{eq:coeff_jacobian_1} \\
        \mathbf{J}_2 & = & \begin{bmatrix}
        -(1-x)(1-z) \\
        -(1-x)z \\
        (1-x)(1-z) \\
        (1-x)z \\
        -x(1-z) \\
        -xz \\
        x(1-z) \\
        xz \\
    \end{bmatrix} \label{eq:coeff_jacobian_2} \\
        \mathbf{J}_3 & = & \begin{bmatrix}
        -(1-x)(1-y) \\
        (1-x)(1-y) \\
        -(1-x)y) \\
        (1-x)y\\
        -x(1-y)\\
        x(1-y) \\
        -xy \\
        xy \\
    \end{bmatrix} \label{eq:coeff_jacobian_3}
\end{eqnarray}
Then the second-order derivative $\frac{\partial^2 f}{\partial \mathbf{x} \partial \theta}$ is given by:
\begin{eqnarray}\label{eq:second_order_theta}
\frac{\partial^2 f}{\partial\mathbf{x}\partial\mathbf{\theta}}
     & = & \frac{\partial}{\partial\theta} \bigg( \frac{\partial f}{\partial\mathbf{x}^T} \bigg) \nonumber \\
    & = & \bigg(\frac{\partial \mathbf{w}}{\partial \mathbf{x}}\bigg)^T
\end{eqnarray}
which is simply the transpose of Eq.~\ref{eq:coeff_jacobian}. And to obtain the other second-order derivative $\frac{\partial^2 f}{\partial \mathbf{x}^2}$, we just need to differentiate Eq.~\ref{eq:jacobian} further w.r.t. the query points:
\begin{eqnarray}
         \frac{\partial^2 f}{\partial \mathbf{x}^2} & = & \frac{\partial}{\partial\mathbf{x}} \bigg( \frac{\partial f}{\partial\mathbf{x}^T} \bigg) \nonumber \\
     & = & \frac{\partial}{\partial\mathbf{x}} \begin{bmatrix}\mathbf{J}^T_1 \theta \\ \mathbf{J}^T_2 \theta \\ \mathbf{J}^T_3 \theta \end{bmatrix} \\
     & = & \begin{bmatrix} h_{11} & h_{12} & h_{13} \\
                         h_{21} & h_{22} & h_{23} \\
                         h_{31} & h_{32} & h_{33} \\
                         \end{bmatrix} \label{eq:second_order_theta_detailed}
\end{eqnarray}
where $\{\mathbf{J}^T_i \theta\}_{i=1}^3$ are inner-products between each column of Eq.~\ref{eq:coeff_jacobian} and $\theta$, so the result is a $3 \times 3$ matrix with each column being the derivative w.r.t. $x$, $y$ and $z$. It is trivial to show that all diagonal elements are zero as each $\mathbf{J}_i$ term only contains the variables of other two dimensions, as in Eq.~\ref{eq:coeff_jacobian_1}, \ref{eq:coeff_jacobian_2} and \ref{eq:coeff_jacobian_3}. The off-diagonal elements can be easily computed as:
\begin{eqnarray}
         h_{12} = h_{21} = (1-z)(\theta_{000} + \theta_{110} - \theta_{010} - \theta_{100}) \nonumber \\
         + \quad z (\theta_{001} + \theta_{111} - \theta_{011} - \theta_{101})
\end{eqnarray}
\begin{eqnarray}
         h_{13} = h_{31} = (1-y)(\theta_{000} + \theta_{101} - \theta_{001} - \theta_{100}) \nonumber \\
         + \quad y (\theta_{010} + \theta_{111} - \theta_{011} - \theta_{110})
\end{eqnarray}
\begin{eqnarray}
         h_{23} = h_{32} = (1-x)(\theta_{000} + \theta_{011} - \theta_{001} - \theta_{010}) \nonumber \\
         + \quad x (\theta_{100} + \theta_{111} - \theta_{101} - \theta_{110})
\end{eqnarray}


\section{Per-scene Quantitative Evaluation}
In this section we provide additional per-scene breakdown of the our quantitative evaluation on the 10 synthetic sequences.

\paragraph{Evaluation Protocol.} For all the methods we run marching cubes at $1\text{cm}$ resolution to extract the meshes for evaluation. We measure accuracy (Acc), completion (Comp), chamfer-$\ell_1$ (C-$\ell_1$), normal consistency (NC) and F-score for evaluation of reconstruction quality. Specifically, all the metrics are computed between point clouds sampled on ground-truth and predicted mesh. Instead of sampling a fixed number of points we sample point cloud at density of $1$ point per $\text{cm}^2$ to take into account the scene scale. The treshold for computing F-score is set to $5\text{ cm}$. 

\paragraph{Mesh Culling.} To prevent the evaluation from falsely penalizing the scene completion ability of our method, surfaces that are not observed in RGB-D images are culled. Following~\cite{azinovic2022neural} we subdivide the meshes such that all the faces have maximum edge length of below $1.5\text{cm}$. A face will be removed if 1. it is not inside any camera frusta, or 2. it is occluded by other geometry, or 3. it has no valid depth measurements. Note for thin geometry sequence, we only apply the first two criteria.

\onecolumn
\begin{longtable}{llccccccc}
        \toprule
        \textbf{Scene}  & \textbf{Method} & \textbf{Acc} $\downarrow$ & \textbf{Comp} $\downarrow$ & \textbf{C-$\ell_1$} $\downarrow$  & \textbf{NC} $\uparrow$  & \textbf{F-score} $\uparrow$ & \textbf{Trans.} $\downarrow$ & \textbf{Rot.} $\downarrow$ \\
        \midrule
        \textbf{Complete kitchen}    & BundleFusion     & 0.0303 & 0.1475 & 0.0889 & 0.8570 & 0.6943 & 0.050 & 0.566 \\
        \multirow{6}{*}[-0.1cm]{\includegraphics[height=0.95in, width=1.5in]{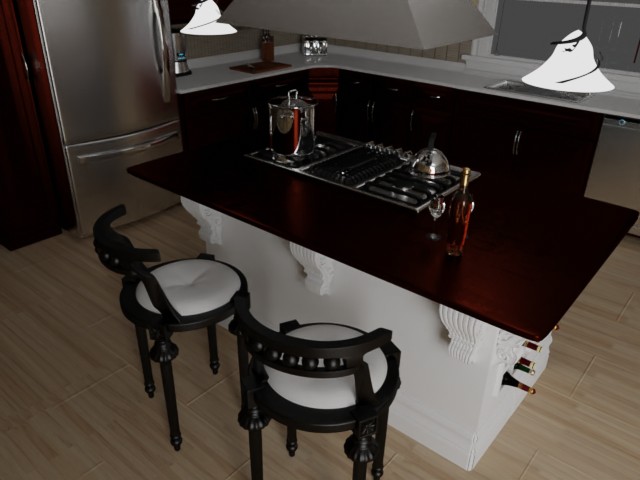}}
        & RoutedFusion & 0.0270 & 0.0854 & 0.0562 & 0.8484 & 0.7939 & - & - \\
        & COLMAP       & 0.0365 & 0.0354 & 0.0360 & 0.9245 & 0.8248 & 0.009 & 0.210  \\
        & ConvOccNets  & 0.0502 & 0.0527 & 0.0514 & 0.8667 & 0.6610 & - & - \\
        & SIREN        & 0.0319 & 0.0700 & 0.0509 & 0.9031 & 0.7415 & - & - \\
        \cmidrule{2-9}       
        & NeuralRGBD   & \textbf{0.0224} & 0.0394 & 0.0309 & 0.9098 & 0.8962 & 0.083 & 0.450 \\
        & Ours         & \textbf{0.0224} & \textbf{0.0258} & \textbf{0.0241} & \textbf{0.9413} & \textbf{0.8998} & \textbf{0.017} & \textbf{0.137}     \\
        \midrule
        
        \textbf{Kitchen}    & BundleFusion     & 0.0253 & 0.0578 & 0.0416 & 0.9112 & 0.7967 & 0.038 & 0.327 \\
        \multirow{6}{*}[-0.1cm]{\includegraphics[height=0.95in, width=1.5in]{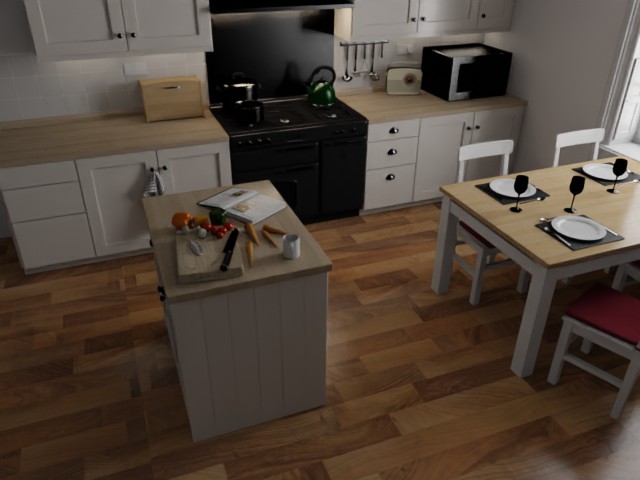}}
        & RoutedFusion & 0.0281 & 0.0362 & 0.0322 & 0.8553 & 0.8484 & - & - \\
        & COLMAP       & 0.0228 & 0.0282 & 0.0255 & 0.9332 & 0.9170 & 0.103 & 0.641  \\
        & ConvOccNets  & 0.0420  & 0.049  & 0.0455 & 0.8752 & 0.6253 & - & - \\
        & SIREN        & 0.0327 & 0.0575 & 0.0451 & 0.8996 & 0.7071 & - & - \\
        \cmidrule{2-9}       
        & NeuralRGBD   & 0.0218 & 0.0297 & 0.0257 & 0.9296 & 0.9005 & 0.030   & \textbf{0.114}  \\
        & Ours         & \textbf{0.0214} & \textbf{0.0271} & \textbf{0.0243 }& \textbf{0.9316} & \textbf{0.9379} & \textbf{0.026} & 0.145     \\
        \midrule   
        
        \textbf{Breakfast room}    & BundleFusion     & 0.0129 & 0.0235 & 0.0182 & 0.9582 & 0.9606 & 0.037 & 0.697 \\
        \multirow{6}{*}[-0.1cm]{\includegraphics[height=0.95in, width=1.5in]{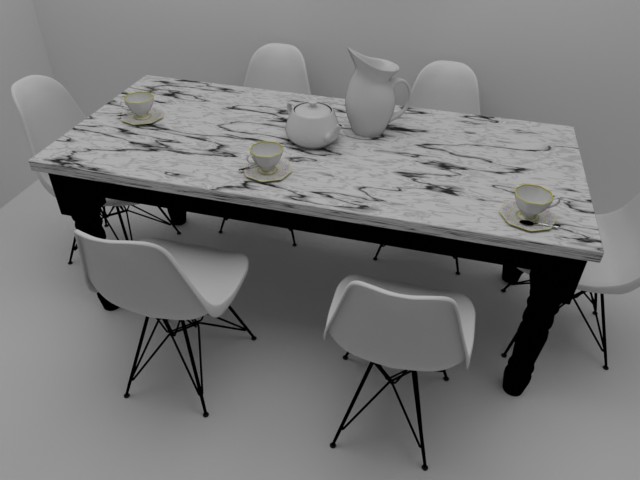}}
        & RoutedFusion & 0.0181 & 0.0202 & 0.0191 & 0.9341 & 0.9758 & - & - \\
        & COLMAP       & 0.0191 & 0.0194 & 0.0192 & 0.9522 & 0.9533 & 0.009 & 0.210  \\
        & ConvOccNets  & 0.0311 & 0.0329 & 0.0320 & 0.8925 & 0.9602 & - & - \\
        & SIREN        & 0.0150 & 0.0454 & 0.0302 & 0.9371 & 0.9230 & - & - \\
        \cmidrule{2-9}       
        & NeuralRGBD   & 0.0145 & 0.0148 & 0.0146 & \textbf{0.9657} & \textbf{0.9898} & \textbf{0.007} & \textbf{0.135} \\
        & Ours         & \textbf{0.0144} & \textbf{0.0136} & \textbf{0.0139} & 0.9629 & 0.9829 & 0.009 & 0.137     \\
        \midrule
        
        \textbf{Morning apartment} & BundleFusion & \textbf{0.0079} & 0.0146 & 0.0112 & 0.8891 & 0.9740 & 0.008 & 0.165 \\
        \multirow{6}{*}[-0.1cm]{\includegraphics[height=0.95in, width=1.5in]{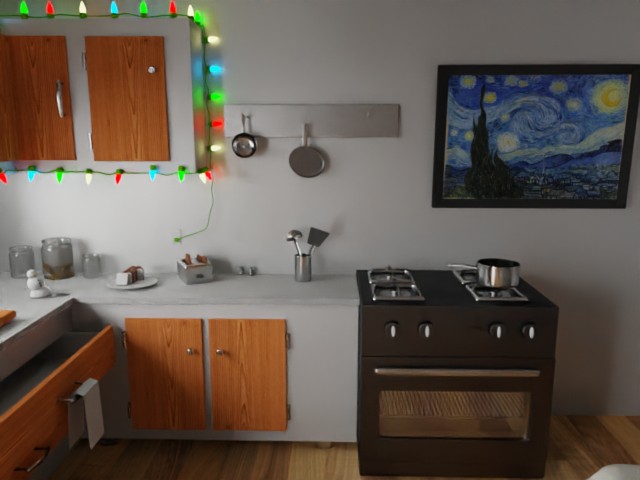}}
        & RoutedFusion & 0.0100   & 0.0143 & 0.0121 & 0.8754 & 0.9795 & - & - \\
        & COLMAP       & 0.0133 & 0.0183 & 0.0158 & 0.8810  & 0.9666 & 0.017 & 0.380  \\
        & ConvOccNets  & 0.0408 & 0.0482 & 0.0445 & 0.8105 & 0.7912 & - & - \\
        & SIREN        & 0.0105 & 0.0146 & 0.0125 & 0.8765 & 0.9718 & - & - \\
        \cmidrule{2-9}       
        & NeuralRGBD   & 0.0087 & \textbf{0.0121} & \textbf{0.0104} & \textbf{0.8918} & \textbf{0.9866} & \textbf{0.005}  & \textbf{0.093} \\
        & Ours         & 0.0095 & 0.0129 & 0.0112 & 0.8874 & 0.9778 & \textbf{0.005} & 0.101     \\
        \midrule
        
        \textbf{Grey white room}    & BundleFusion     & 0.0297 & 0.0456 & 0.0377 & 0.8612 & 0.7537 & 0.056 & 1.891 \\
        \multirow{6}{*}[-0.1cm]{\includegraphics[height=0.95in, width=1.5in]{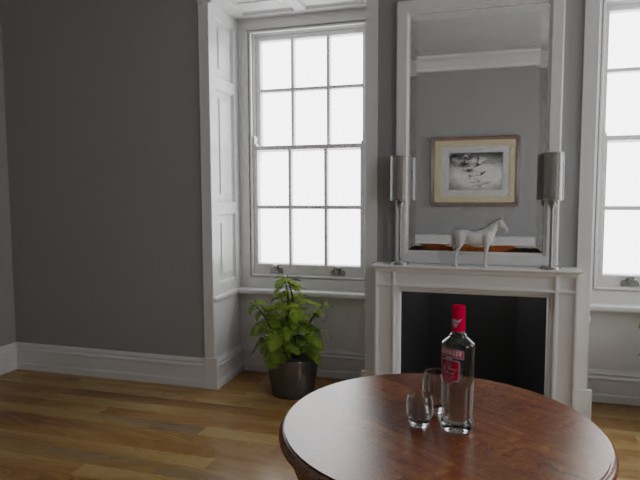}}
        & RoutedFusion & 0.0303 & 0.0347 & 0.0325 & 0.8531 & 0.7908 & - & - \\
        & COLMAP       & 0.0287 & 0.0293 & 0.0290 & 0.9013 & 0.9036 & 0.029 & 0.296 \\
        & ConvOccNets  & 0.0470 & 0.0488 & 0.0479 & 0.8434 & 0.6057 & - & - \\
        & SIREN        & 0.0323 & 0.0335 & 0.0329 & 0.8697 & 0.8142 & - & - \\
        \cmidrule{2-9}       
        & NeuralRGBD   & \textbf{0.0132} & \textbf{0.0151} & \textbf{0.0142} & \textbf{0.9318} & \textbf{0.9923} & 0.014  & \textbf{0.146} \\
        & Ours         & 0.0140 & 0.0158 & 0.0149 & 0.9261 & 0.9895 & \textbf{0.013} & 0.205     \\
        
    
        \bottomrule
    \caption{We compare the quality of our reconstruction on several synthetic scenes for which ground truth data is available. We measure accuracy, completion, chamfer $\ell_1$ distance, normal consistency and F-score for reconstruction quality. The numbers are computed between point clouds sampled with a density of 1 point per cm\textsuperscript{2}.}
\vspace{-0.3cm}
\label{tab:quantitative_per_scene_1}
\end{longtable}
\onecolumn
\begin{longtable}{llccccccc}
        \toprule
        \textbf{Scene}  & \textbf{Method} & \textbf{Acc} $\downarrow$ & \textbf{Comp} $\downarrow$ & \textbf{C-$\ell_1$} $\downarrow$  & \textbf{NC} $\uparrow$  & \textbf{F-score} $\uparrow$ & \textbf{Trans.} $\downarrow$ & \textbf{Rot.} $\downarrow$ \\
        
        \midrule        
        \textbf{White room}    & BundleFusion     & 0.0276 & 0.0918 & 0.0597 & 0.8788 & 0.7286 & 0.045 & 0.375 \\
        \multirow{6}{*}[-0.1cm]{\includegraphics[height=0.95in, width=1.5in]{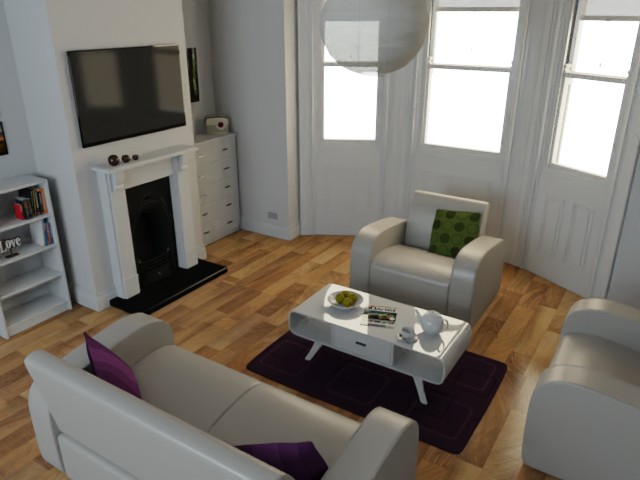}}
        & RoutedFusion & 0.0289 & 0.0430 & 0.0360 & 0.8280 & 0.8222 & - & - \\
        & COLMAP       & 0.0309 & 0.0342 & 0.0325 & 0.9188 & 0.8259 & \textbf{0.018} & 0.167  \\
        & ConvOccNets  & 0.0537 & 0.0583 & 0.0560 & 0.8653 & 0.5012 & - & - \\
        & SIREN        & 0.0276 & 0.0588 & 0.0432 & 0.8992 & 0.7788 & - & - \\
        \cmidrule{2-9}       
        & NeuralRGBD   & \textbf{0.0204} & \textbf{0.0256} & \textbf{0.0230}  & \textbf{0.9297} & \textbf{0.9551} & 0.028  & \textbf{0.146} \\
        & Ours         & 0.0210 & 0.0325 & 0.0268 & 0.9281 & 0.9233 & \textbf{0.024} & 0.157 \\
        
        \midrule
        \textbf{Green room}    & BundleFusion & 0.0118 & 0.0339 & 0.0228 & 0.9254 & 0.9314 & 0.027 & 0.546 \\
        \multirow{6}{*}[-0.1cm]{\includegraphics[height=0.95in, width=1.5in]{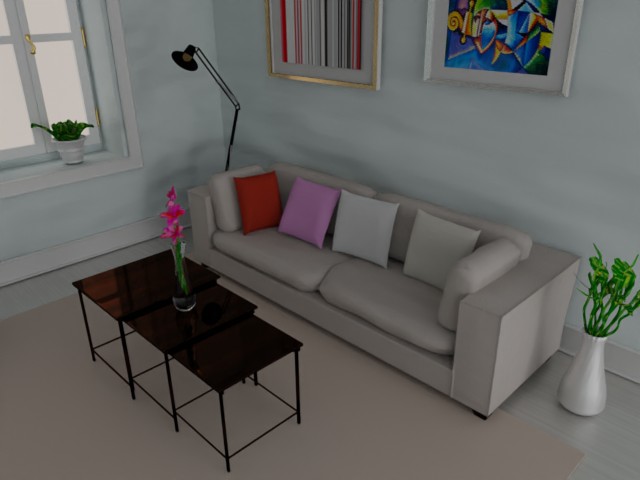}}
        & RoutedFusion & 0.0156 & 0.0193 & 0.0174 & 0.9095 & 0.9735 & -     & -     \\
        & COLMAP       & 0.0159 & 0.0194 & 0.0177 & 0.9270 & 0.9712 & 0.014 & 0.227 \\
        & ConvOccNets  & 0.0548 & 0.0493 & 0.0521 & 0.8600 & 0.7434 & -     & -     \\
        & SIREN        & 0.0183 & 0.0253 & 0.0218 & 0.9143 & 0.9448 & -     & -     \\
        \cmidrule{2-9}       
        & NeuralRGBD   & \textbf{0.0106} & \textbf{0.0142} & \textbf{0.0124} & \textbf{0.9348} & \textbf{0.9913} & \textbf{0.012} & 0.104 \\
        & Ours         & 0.0138 & 0.0169 & 0.0153 & 0.9256 & 0.9838 & 0.014 & \textbf{0.085} \\
        \midrule
        
        \textbf{Staircase}    & BundleFusion & 0.0257 & 0.1146 & 0.0701 & 0.8792 & 0.7108 & 0.039 & 0.643 \\
        \multirow{6}{*}[-0.1cm]{\includegraphics[height=0.95in, width=1.5in]{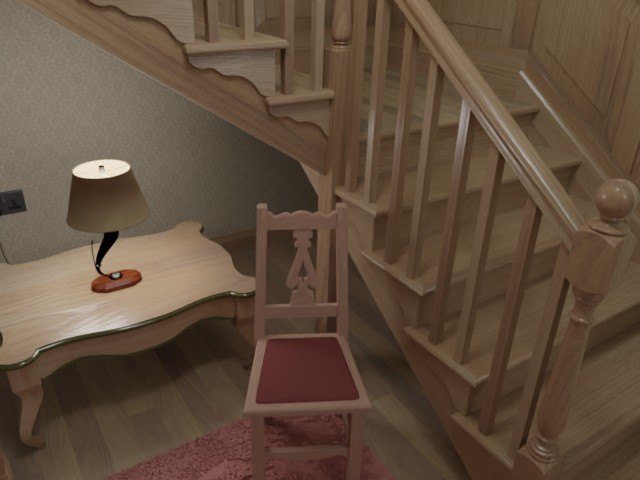}}
        & RoutedFusion & 0.0411 & 0.0512 & 0.0461 & 0.8909 & 0.6896 & -     & -     \\
        & COLMAP       & 0.0454 & 0.058  & 0.0517 & 0.9253 & 0.6875 & 0.043 & 0.305 \\
        & ConvOccNets  & 0.0618 & 0.0562 & 0.059  & 0.8601 & 0.5646 & -     & -     \\
        & SIREN        & 0.0355 & 0.0514 & 0.0434 & 0.9117 & 0.7487 & -     & -     \\
        \cmidrule{2-9}   
        & NeuralRGBD   & \textbf{0.0216} & \textbf{0.0254} & \textbf{0.0235} & 0.9471 & \textbf{0.9333} & 0.016  & 0.123  \\
        & Ours         & 0.0221 & 0.0257 & 0.024  & \textbf{0.9496} & 0.9235 & \textbf{0.015} & 0.144     \\
        \midrule   
        
        \textbf{Thin geometry}    & BundleFusion & 0.0072 & 0.0305 & 0.0188 & 0.9063 & 0.9199 & \textbf{0.009} & 0.126 \\
        \multirow{6}{*}[-0.1cm]{\includegraphics[height=0.95in, width=1.5in]{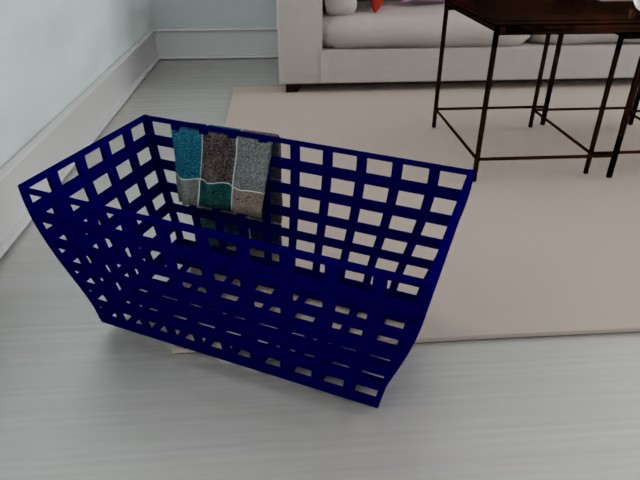}}
        & RoutedFusion & \textbf{0.0070} & 0.0396 & 0.0233 & 0.8243 & 0.8785 & -     & -     \\
        & COLMAP       & 0.0372 & 0.0558 & 0.0465 & 0.8181 & 0.7209 & 0.079 & 2.4   \\
        & ConvOccNets  & 0.0115 & 0.0329 & 0.0222 & 0.8800 & 0.9072 & -     & -     \\
        & SIREN        & 0.0086 & 0.0335 & 0.0210  & 0.8823 & 0.9115 & -     & - \\
        \cmidrule{2-9}       
        & NeuralRGBD   & 0.0079 & \textbf{0.0092} & \textbf{0.0086} & \textbf{0.9077} & \textbf{0.9956} & \textbf{0.010} & \textbf{0.037} \\
        & Ours         & 0.0093 & 0.0121 & 0.0107 & 0.8986 & 0.9817 & 0.011 & 0.146 \\
        \midrule
        
        \textbf{ICL living room}    & BundleFusion & 0.0129 & 0.0214 & 0.0172 & 0.9606 & 0.9694 & 0.022 & 0.382 \\
        \multirow{6}{*}[-0.1cm]{\includegraphics[height=0.95in, width=1.5in]{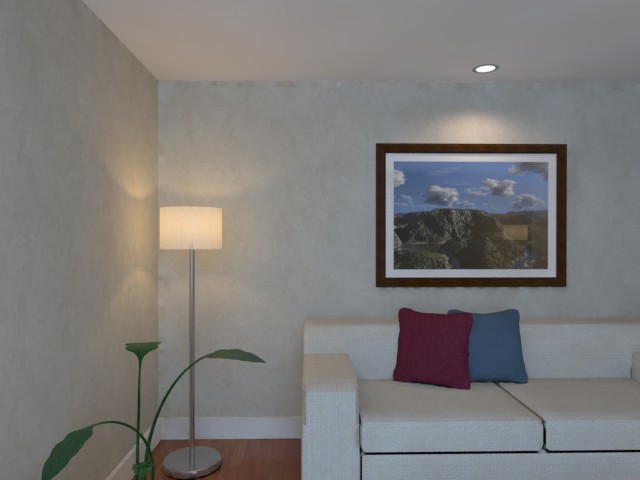}}
        & RoutedFusion & 0.0168 & 0.0201 & 0.0185 & 0.9456 & 0.9841 & -     & -     \\
        & COLMAP       & 0.0209 & 0.0238 & 0.0224 & 0.9528 & 0.9730  & 0.029 & 0.836 \\
        & ConvOccNets  & 0.1049 & 0.0956 & 0.1003 & 0.8535 & 0.5619 & -     & -     \\
        & SIREN        & 0.0167 & 0.0219 & 0.0193 & 0.9555 & 0.9734 & -     & - \\
        \cmidrule{2-9}       
        & NeuralRGBD   & \textbf{0.0095} & \textbf{0.0115} & \textbf{0.0105} & \textbf{0.9689} & \textbf{0.9944} & \textbf{0.007} & \textbf{0.109} \\
        & Ours         & 0.0105 & 0.0127 & 0.0117 & 0.9661 & 0.9909 & \textbf{0.007} & 0.167 \\
    
        \bottomrule
    \caption{We compare the quality of our reconstruction on several synthetic scenes for which ground truth data is available. We measure accuracy, completion, chamfer $\ell_1$ distance, normal consistency and F-score for reconstruction quality. The numbers are computed between point clouds sampled with a density of 1 point per cm\textsuperscript{2}.}
\vspace{-0.3cm}
\label{tab:quantitative_per_scene_2}
\end{longtable}

\twocolumn

\paragraph{Synthetic Dataset.} We evaluate on the same synthetic dataset as in~\cite{azinovic2022neural} which consists of $10$ scenes published under the CC-BY or CC-0 license. For more details please refer to their paper~\cite{azinovic2022neural}.

\section{More Ablation Studies}
In this section, we show additional ablation studies on the effect of RGB loss term, regularisation terms .

\subsection{Effect of RGB Loss Term}

Similar to~\cite{azinovic2022neural}, in the main paper we showed in Fig. 5 that the RGB loss term is able to capture better high-frequency details and recover missing depth regions. In this section we show quantitative evaluation on a synthetic scene with thin geometries that have no depth measurements, and also show more qualitative ablation results on real-world ScanNet scenes.

\begin{figure}[tpb]
    \centering
    \newcolumntype{Y}{>{\centering\arraybackslash}X}
    \begin{tabularx}{0.90\linewidth}{@{}Y@{\,}Y@{}}
        \includegraphics[width=\linewidth]{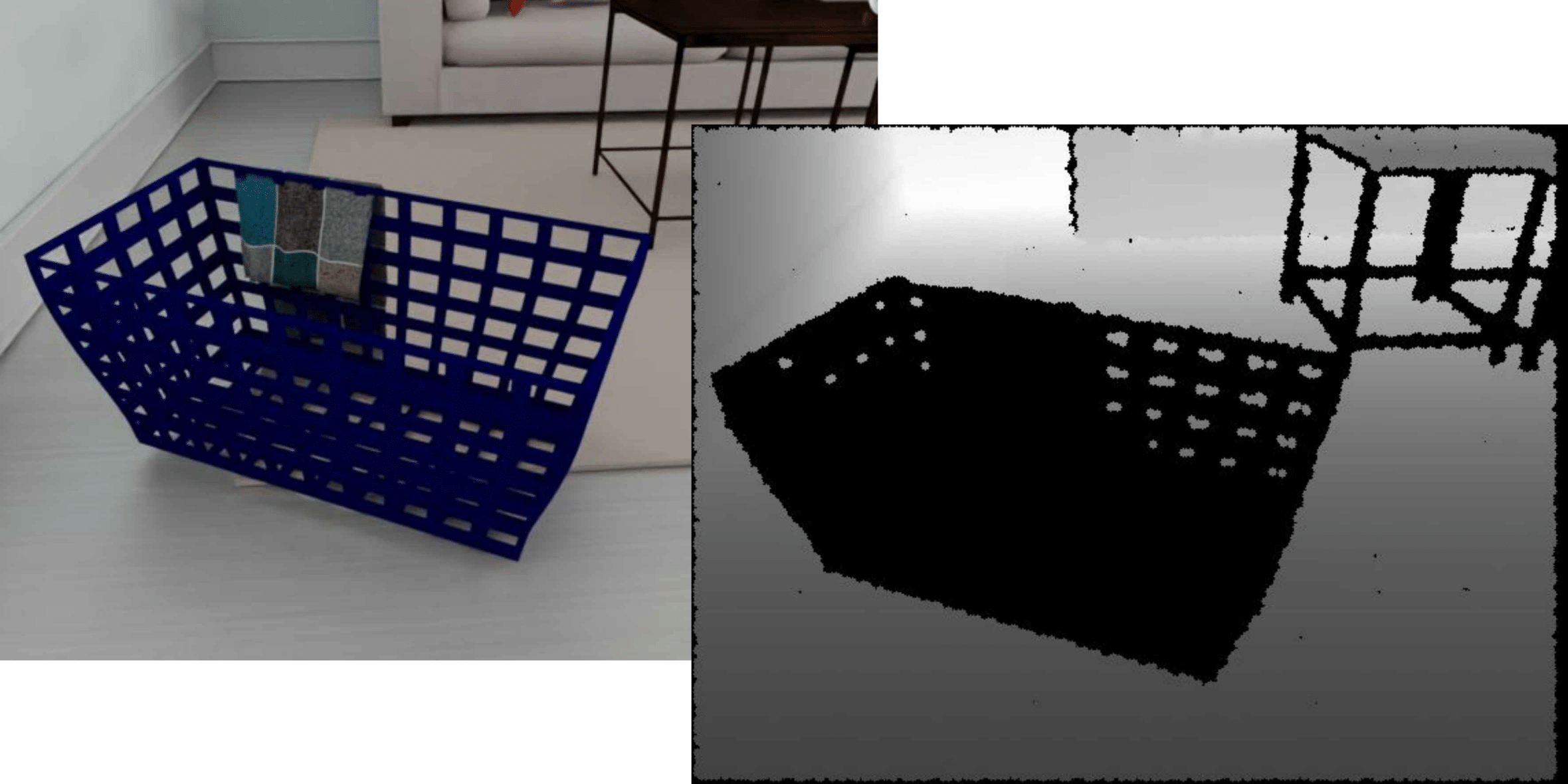}
        & \includegraphics[width=\linewidth]{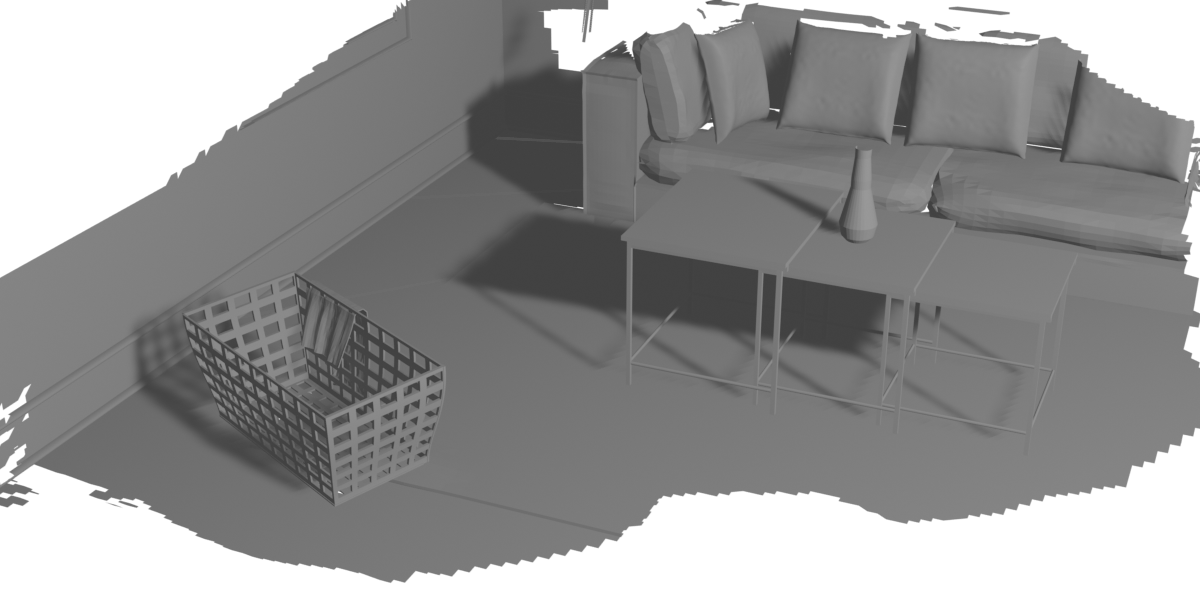} \\
        RGB-D Input & Ground Truth \\
        \includegraphics[width=\linewidth]{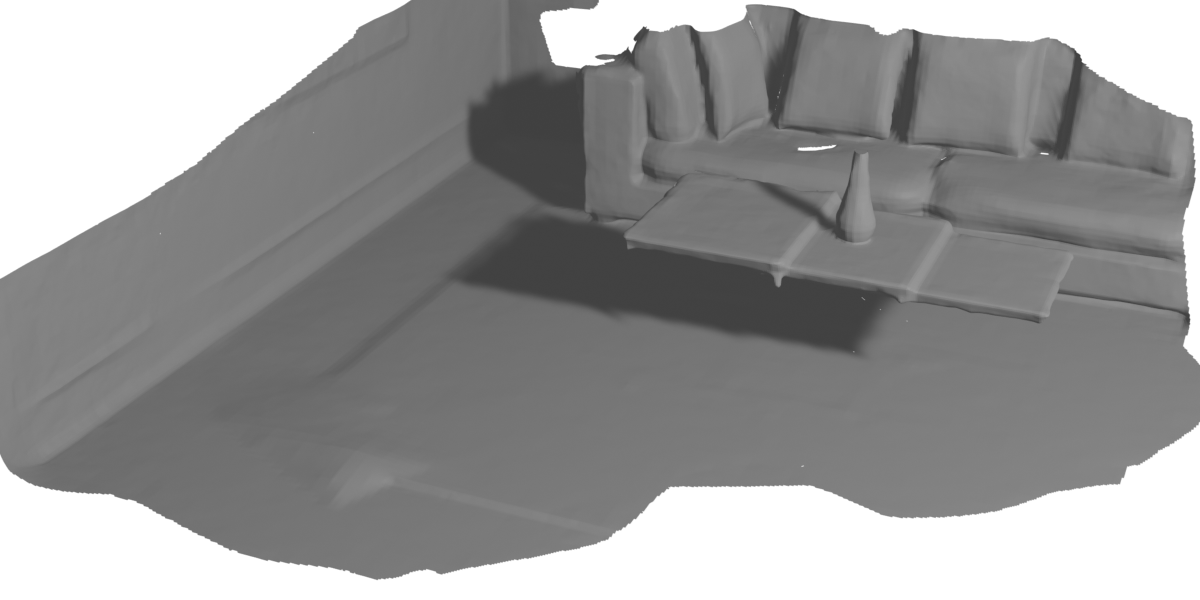}&
        \includegraphics[width=\linewidth]{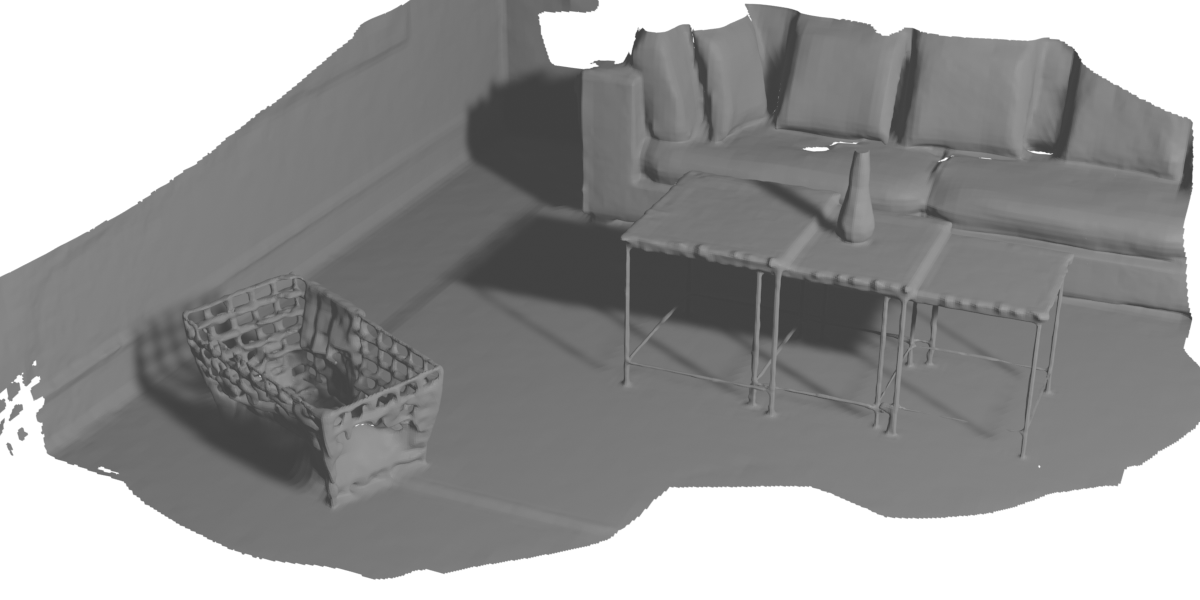}\\
        Ours w/o RGB term &
        Ours-full
    \end{tabularx}
    \caption{Qualitative reconstruction results on synthetic sequence with thin geometries and missing depth measurements.\label{fig:rgb_ablation_thin_geometry} }
    \vspace{-2mm}
\end{figure}

For the experiment on the synthetic scene, we simulate missing depth by removing the depth measurements from the baskets and table legs (top left corner in Fig.~\ref{fig:rgb_ablation_thin_geometry}). Comparison in Fig.~\ref{fig:rgb_ablation_thin_geometry} demonstrates that RGB loss term is able to recover thin structures with missing depth and produce complete reconstruction. Tab.~\ref{tab:rgb_ablation_thin_geometry} also shows our full model achieves significantly better reconstruction quality, especially completeness.

\begin{table}[tp]
	\resizebox{\linewidth}{!}{%
    \centering
    \begin{tabular}{lcccccc}
        \toprule
        \textbf{Method}  & \textbf{Acc.)} $\downarrow$  & \textbf{Com.} $\downarrow$ & \textbf{C-$\ell_1$} $\downarrow$ & \textbf{NC} $\uparrow$  & \textbf{F-score} $\uparrow$ \\
        \midrule
        Ours (no rgb) & \textbf{0.0087}  & 0.0291 & 0.0189 & 0.8967 & 0.9273\\
        Ours (full)   & 0.0093 & \textbf{0.0121} & \textbf{0.0107} & \textbf{0.8986} & \textbf{0.9817} \\
        \bottomrule
    \end{tabular}
    }
    \caption{
    We evaluate the reconstruction results of our model with out RGB loss term and our full model on a synthetic sequence with thin geometries. 
    }
\vspace{-4mm}
\label{tab:rgb_ablation_thin_geometry}
\end{table}

\begin{figure}[tbp]
    \centering
    \newcolumntype{Y}{>{\centering\arraybackslash}X}
    \begin{tabularx}{0.93\linewidth}{@{}Y@{\,}Y@{}}
        \includegraphics[width=\linewidth]{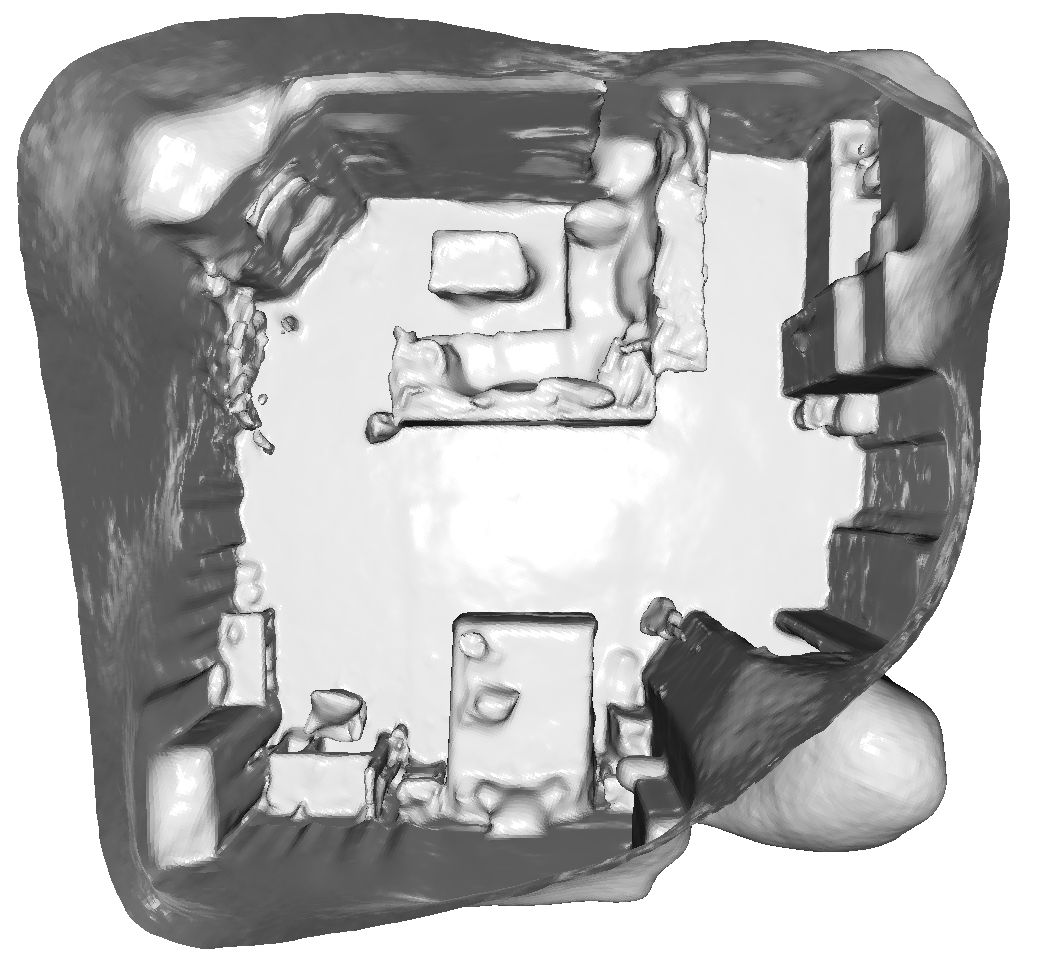}
        & \includegraphics[width=\linewidth]{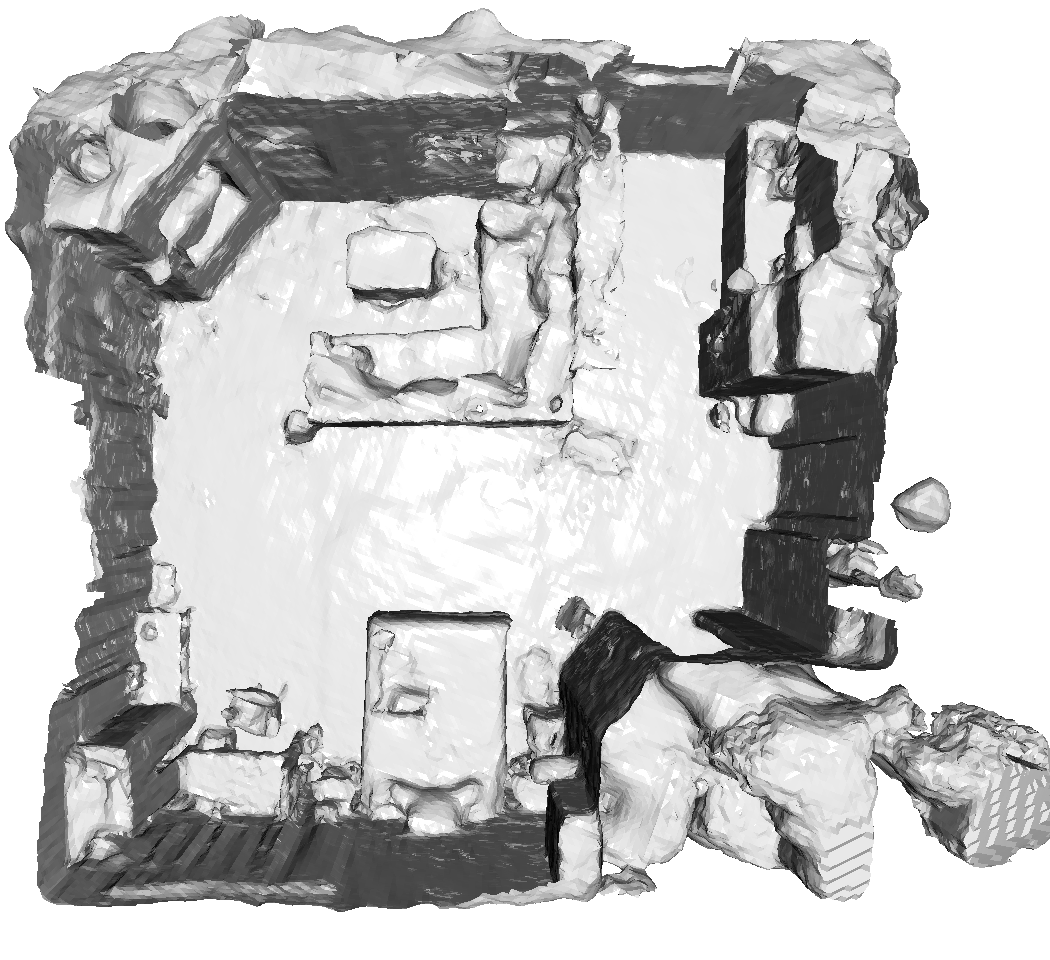} \\
        GO-Surf (Ours) &
        NICE-SLAM~\cite{Zhu2022CVPR_niceslam}
    \end{tabularx}
    \caption{Reconstructed meshes (unculled) of sequential mapping on ScanNet scene0000. Note that for fair comparison NICE-SLAM is running in mapping mode with ground truth camera poses. GO-Surf runs over $\times 30$ faster while achieving much smoother high-quality reconstruction. \label{fig:mapping_result} }
    \vspace{-5mm}
\end{figure}

\subsection{Effect of SDF Regularisation Terms}
In this section we provide additional ablation studies on the two SDF regularisation terms.

\begin{figure*}[tb]
    \centering
    \newcolumntype{Y}{>{\centering\arraybackslash}X}
    \begin{tabularx}{0.99\linewidth}{@{}Y@{\,}Y@{\,}Y@{}}
        \includegraphics[width=\linewidth]{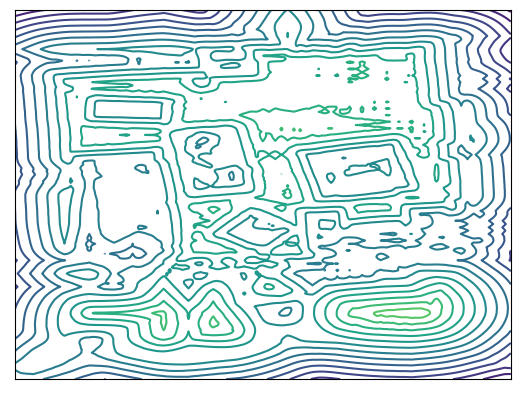}&
        \includegraphics[width=\linewidth]{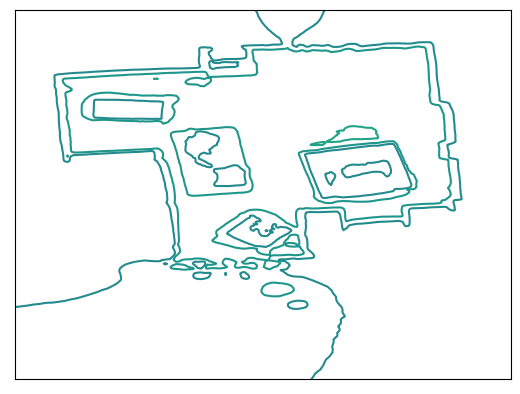}&
        \includegraphics[width=\linewidth]{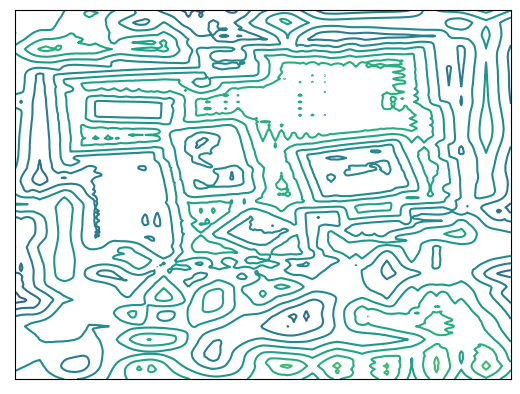}\\
         Full model &
         w/o Eikonal term &
         w/o Geometric initialisation
    \end{tabularx}
    \caption{We found that applying the Eikonal term to some extent propagates the SDF gradient to empty spaces. Without the Eikonal term SDF is reconstructed only within the truncation region. Geometric initialisation reduces the noise in the unobserved regions (e.g. behind walls). \label{fig:ablation_eikonal}}
\end{figure*}

\paragraph{Eikonal Term}
The Eikonal term encourages the model prediction to be a valid signed distance field. We observe that without the Eikonal term the SDF values are being reconstructed only within the truncation region (i.e. very close to the surface). The Eikonal term helps correct SDF values to be propagated outside of the truncation region, although some local artefacts still remain (see Fig.~\ref{fig:ablation_eikonal}) which we suspect are due to the local nature of gradients in the feature grid.

\begin{figure*}[tb]
    \centering
    \newcolumntype{Y}{>{\centering\arraybackslash}X}
    \begin{tabularx}{0.93\linewidth}{@{}Y@{\,}Y@{\,}Y@{\,}Y@{}}
        \includegraphics[width=\linewidth]{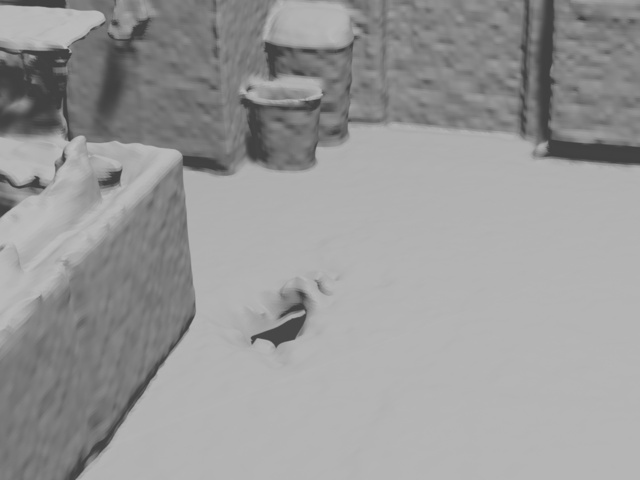}&
        \includegraphics[width=\linewidth]{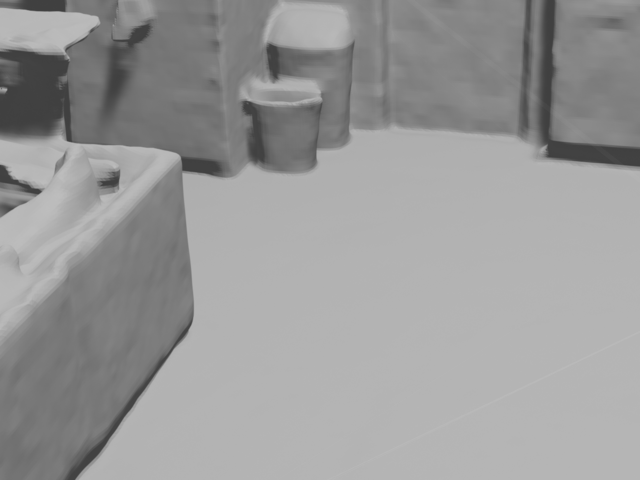}&
        \includegraphics[width=\linewidth]{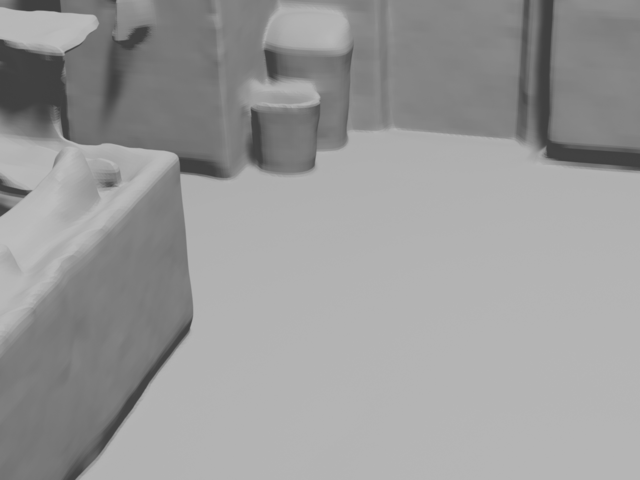}&
        \includegraphics[width=\linewidth]{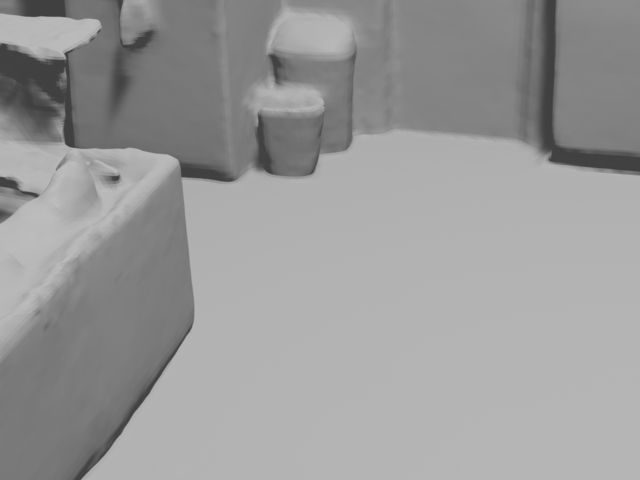}\\
        \includegraphics[width=\linewidth]{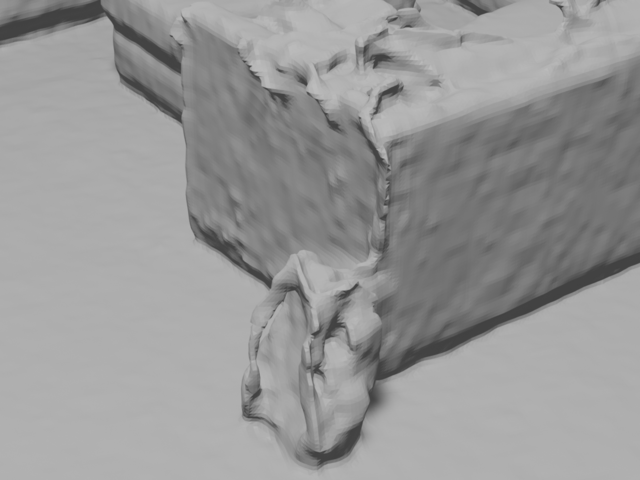}&
        \includegraphics[width=\linewidth]{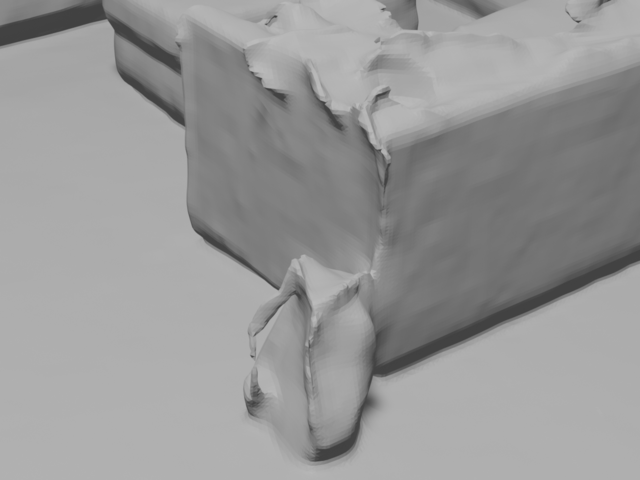}&
        \includegraphics[width=\linewidth]{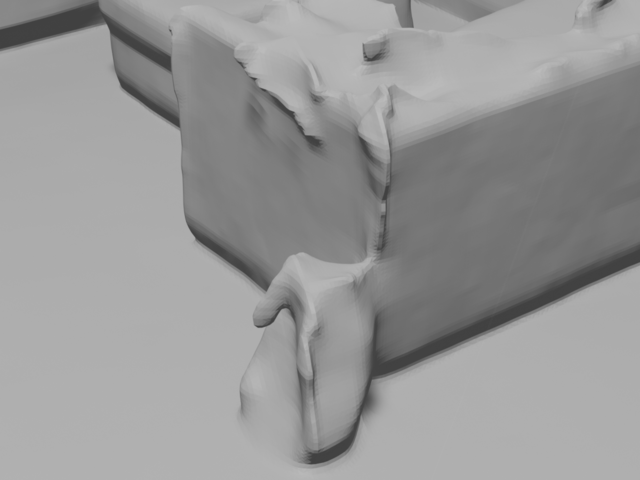}&
        \includegraphics[width=\linewidth]{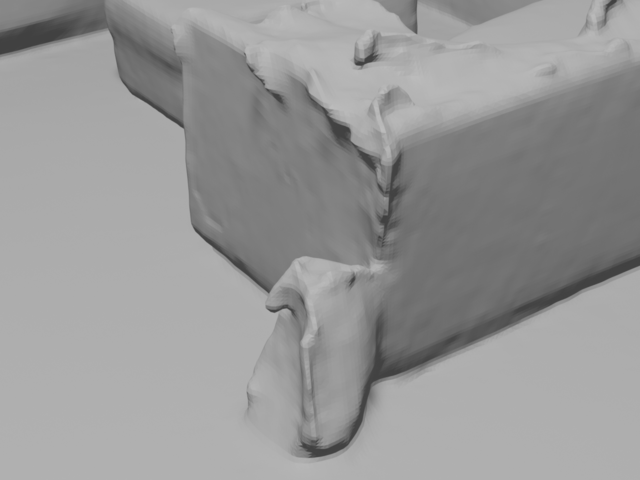}\\
        \includegraphics[width=\linewidth]{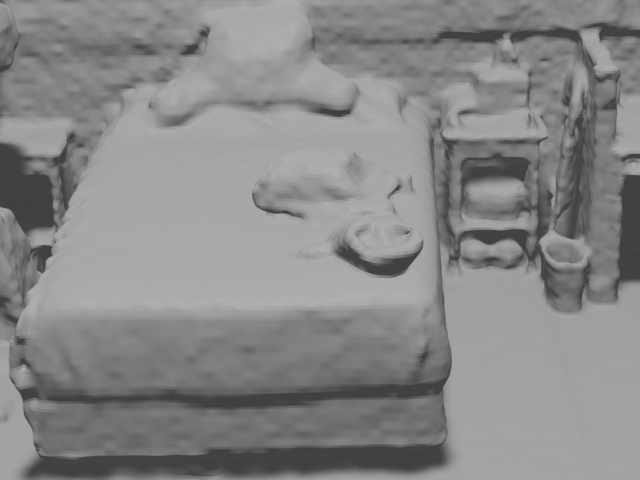}&
        \includegraphics[width=\linewidth]{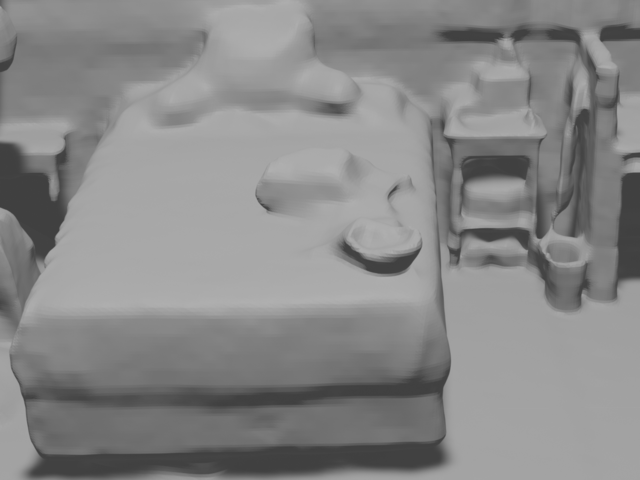}&
        \includegraphics[width=\linewidth]{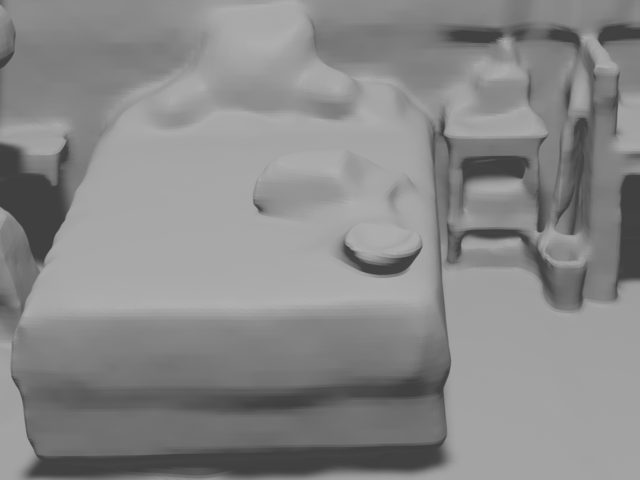}&
        \includegraphics[width=\linewidth]{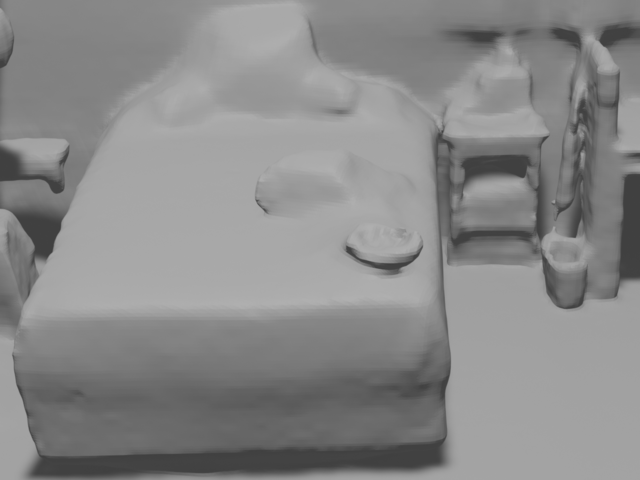}\\
        \includegraphics[width=\linewidth]{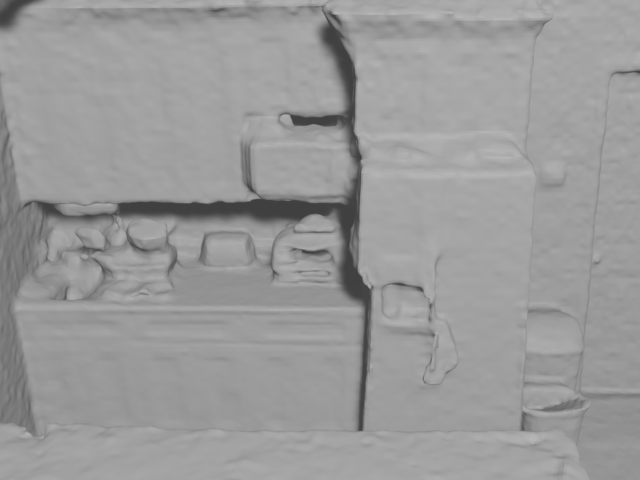}&
        \includegraphics[width=\linewidth]{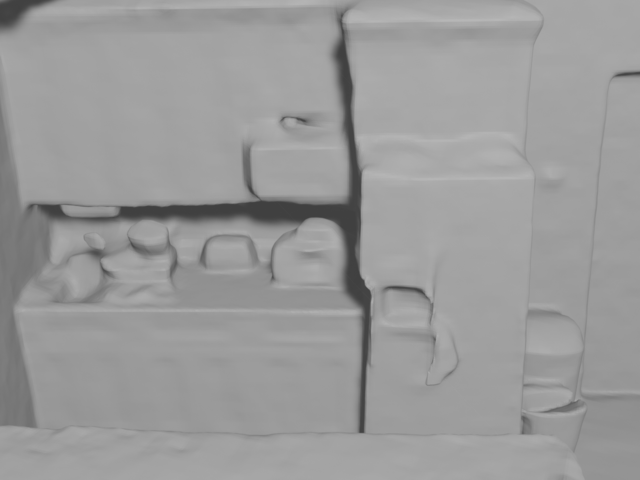}&
        \includegraphics[width=\linewidth]{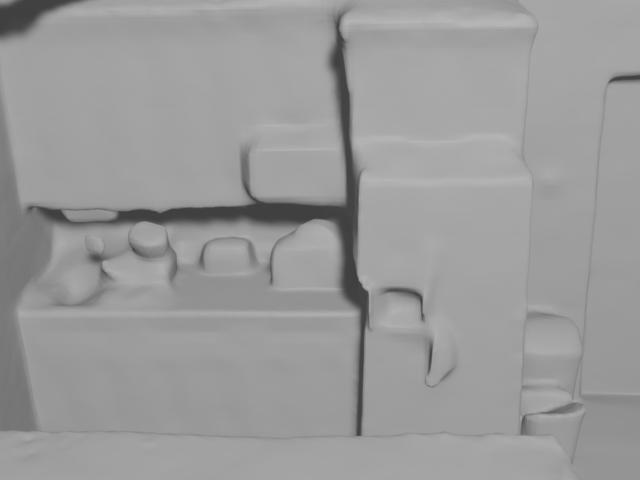}&
        \includegraphics[width=\linewidth]{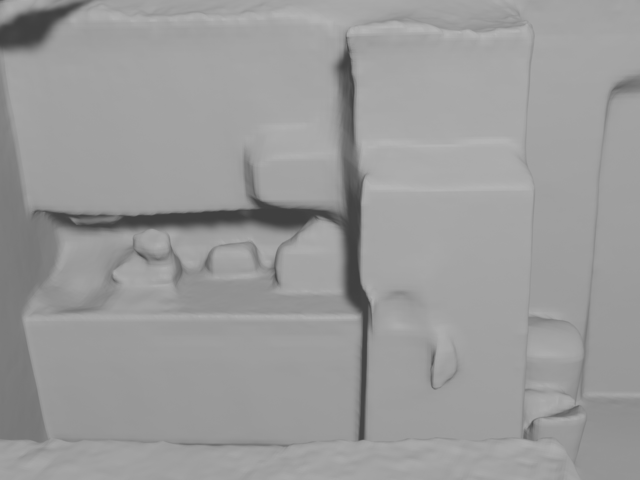}\\
         $\delta_s = 0$ (no reg.) &
         $\delta_s = 5\text{mm}$ &
         $\delta_s = 2 \text{cm}$ &
         $\delta_s = 10 \text{cm}$
    \end{tabularx}
    \caption{Ablation study on effect of smoothness prior. We compare the reconstruction results with different values of $\delta_s$ for normal regularisation term. With $\delta_s = 0$ the results capture more details like the backpack strap and tiny objects on the kitchen table, but also have more high frequency noises. Also the hole shows up near the sofa. Increasing $\delta_s$ to $10\text{cm}$ results in over-smoothed reconstruction. We empirically found $2-5\text{mm}$ is a good trade-off between smoothness and details. \label{fig:smoothness_ablation_scannet}}
\end{figure*}

\paragraph{Normal Smoothness Term}
We found that the normal smoothness term fills the holes in unobserved regions which is particularly useful for fixing discontinuities in large planar structures like walls or floor. It also encourages surface smoothness in all other areas, usually at the cost of some fine details. We investigate different settings of normal smoothness radius (see Fig.~\ref{fig:smoothness_ablation_scannet}).

\section{More Results}
\begin{figure*}[tbp]
  \centering
  \setlength{\tabcolsep}{0.5pt}
  \newcommand{\sz}{0.24}  %

  \begin{tabular}{lcccc}

    \makecell{\rotatebox{90}{BundleFusion~\cite{dai2017bundlefusion}}}  &
    \makecell{\includegraphics[width=\sz\linewidth]{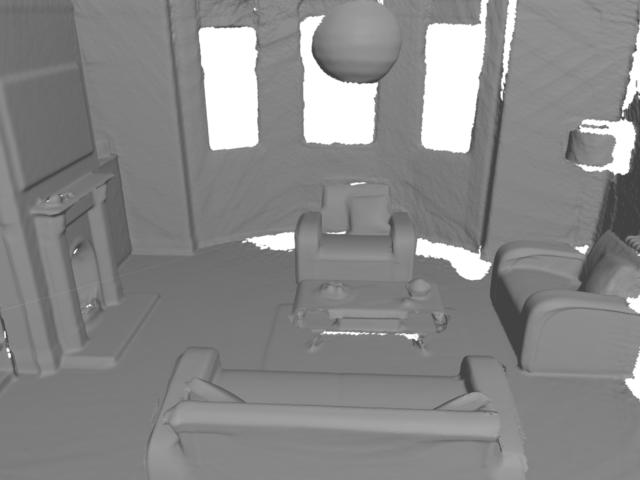}} & 
    \makecell{\includegraphics[width=\sz\linewidth]{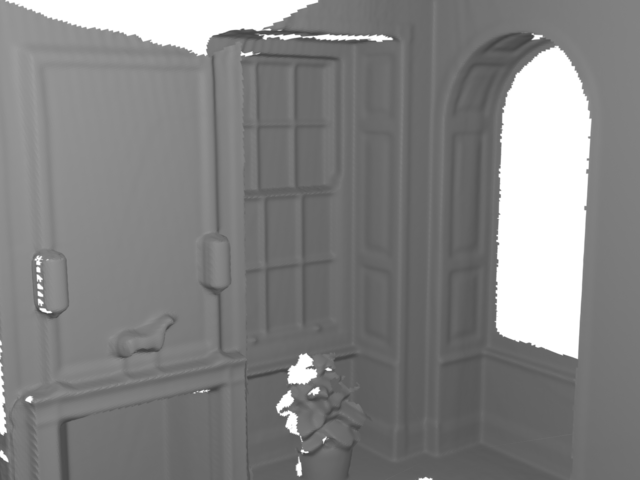}} &
    \makecell{\includegraphics[width=\sz\linewidth]{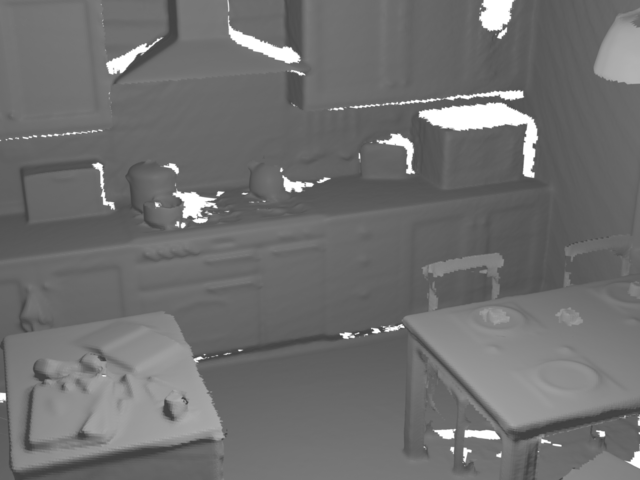}} &
    \makecell{\includegraphics[width=\sz\linewidth]{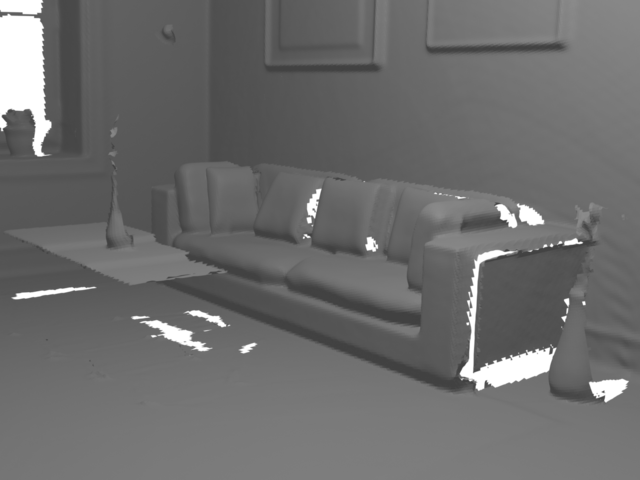}} \\
    
    
    \makecell{\rotatebox{90}{ConvOccNets~\cite{Peng2020ECCV}}}  &
    \makecell{\includegraphics[width=\sz\linewidth]{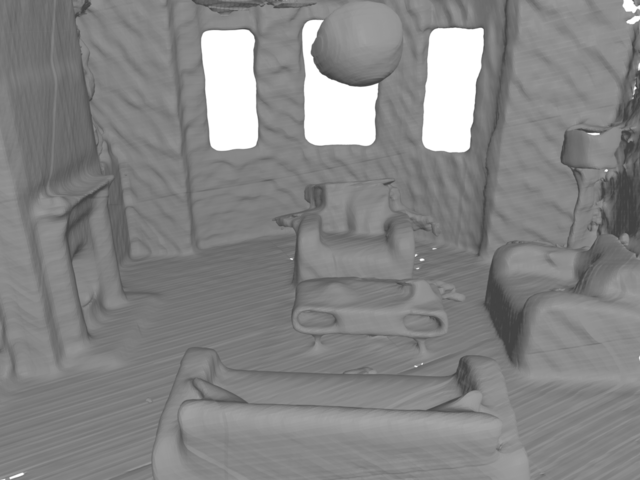}} & 
    \makecell{\includegraphics[width=\sz\linewidth]{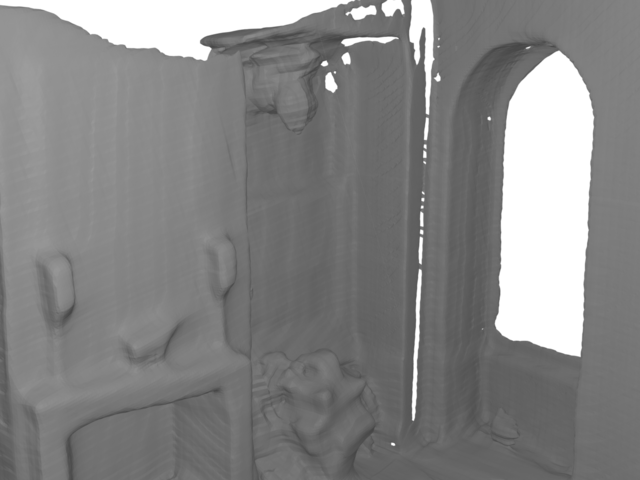}} &
    \makecell{\includegraphics[width=\sz\linewidth]{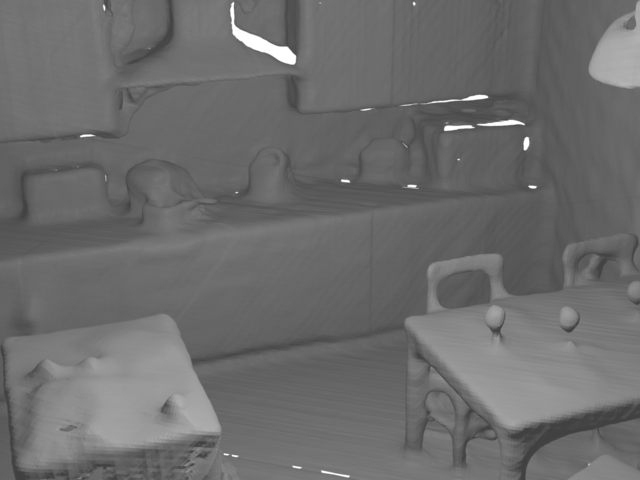}} &
    \makecell{\includegraphics[width=\sz\linewidth]{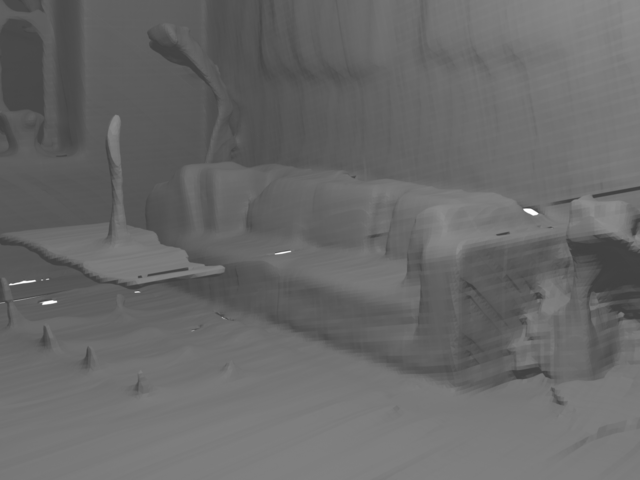}} \\
    
    
    \makecell{\rotatebox{90}{SIREN~\cite{sitzmann2019siren}}} &
    \makecell{\includegraphics[width=\sz\linewidth]{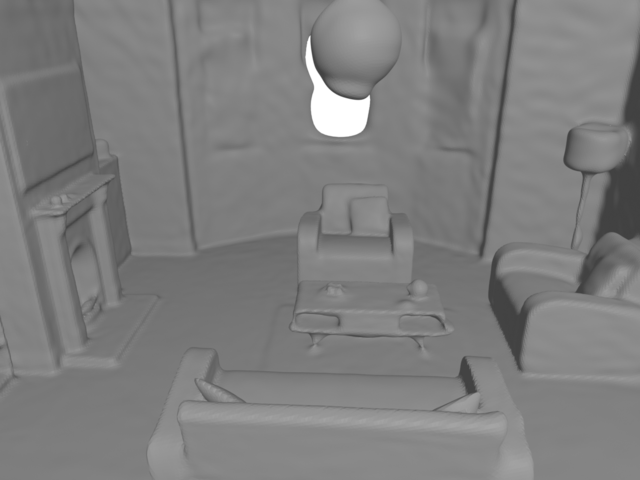}} & 
    \makecell{\includegraphics[width=\sz\linewidth]{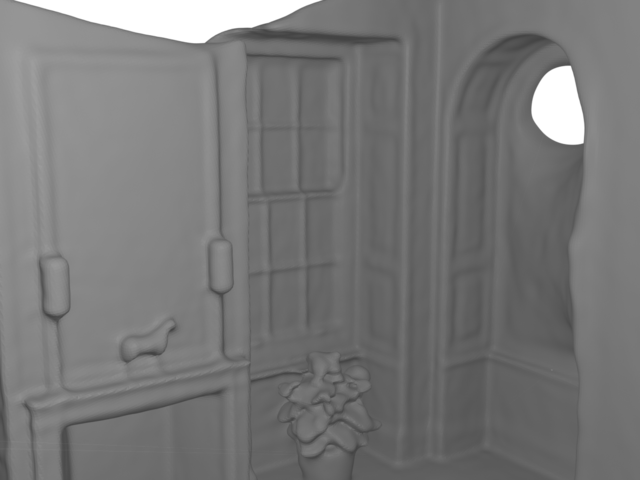}} &
    \makecell{\includegraphics[width=\sz\linewidth]{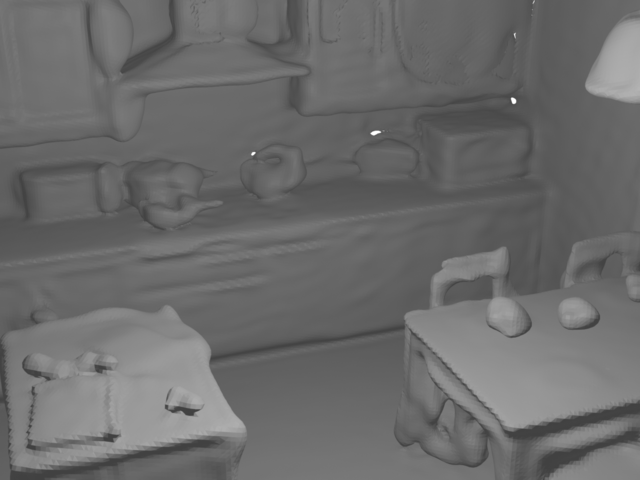}} &
    \makecell{\includegraphics[width=\sz\linewidth]{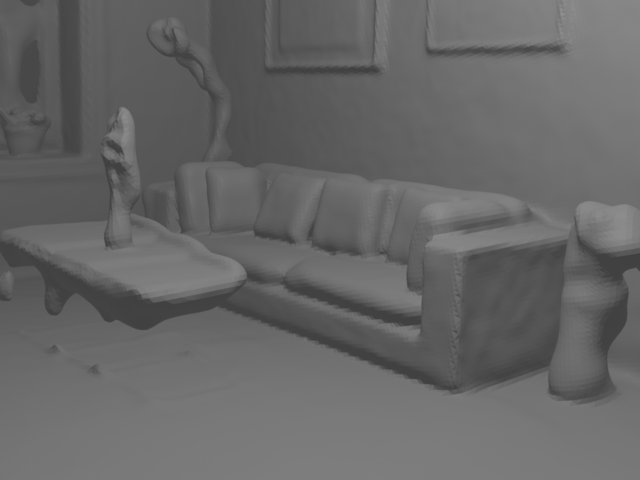}} \\

    \makecell{\rotatebox{90}{NeuralRGBD~\cite{azinovic2022neural}}} &
    \makecell{\includegraphics[width=\sz\linewidth]{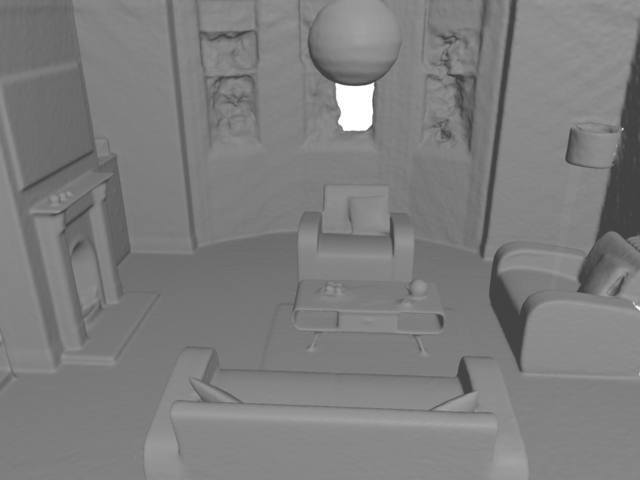}} & 
    \makecell{\includegraphics[width=\sz\linewidth]{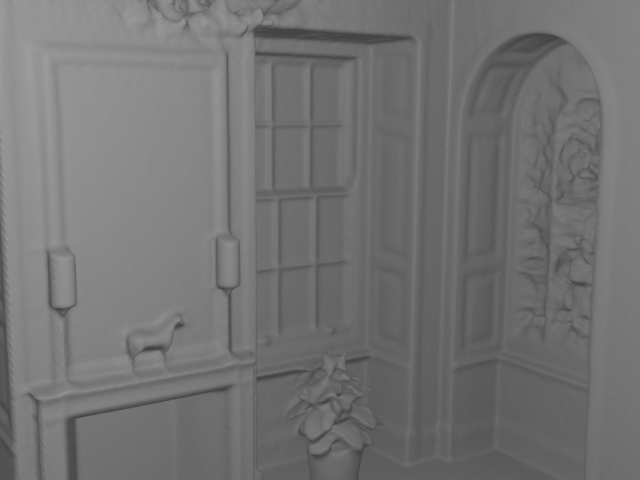}} &
    \makecell{\includegraphics[width=\sz\linewidth]{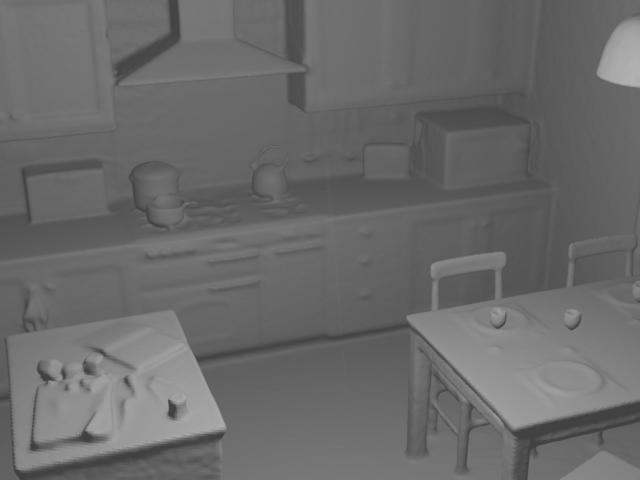}} &
    \makecell{\includegraphics[width=\sz\linewidth]{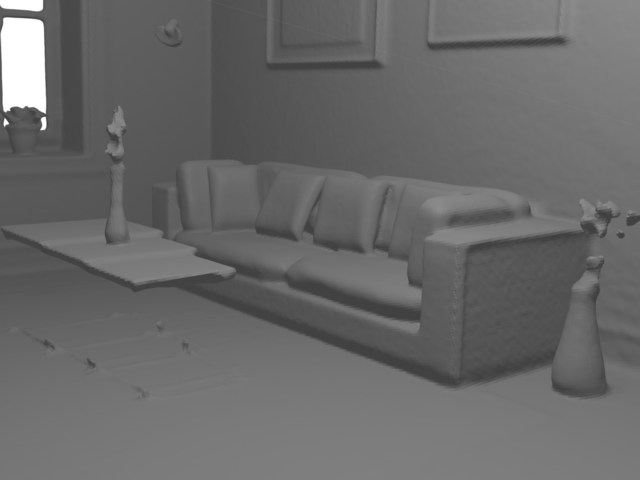}} \\

    \makecell{\rotatebox{90}{GO-Surf (Ours)}} &
    \makecell{\includegraphics[width=\sz\linewidth]{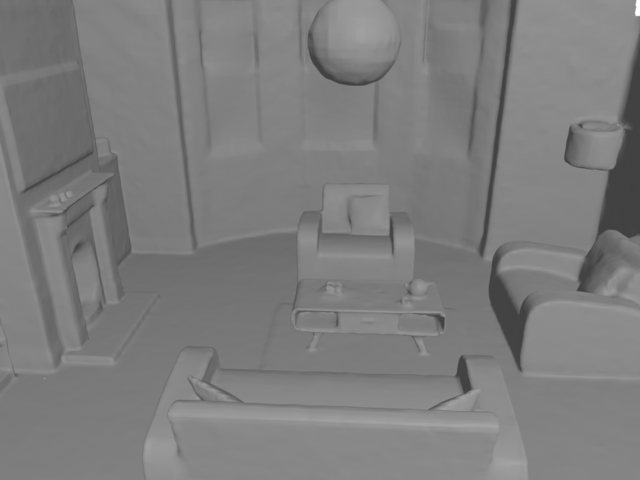}} & 
    \makecell{\includegraphics[width=\sz\linewidth]{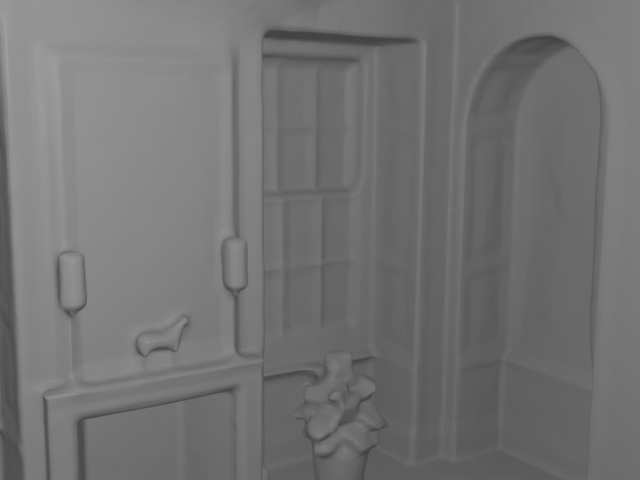}} &
    \makecell{\includegraphics[width=\sz\linewidth]{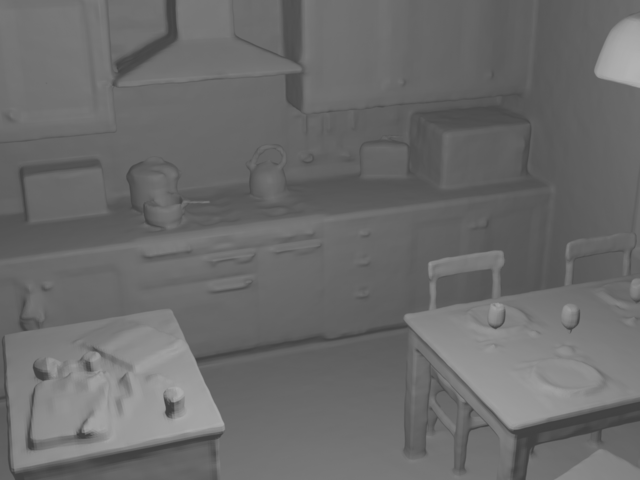}} &
    \makecell{\includegraphics[width=\sz\linewidth]{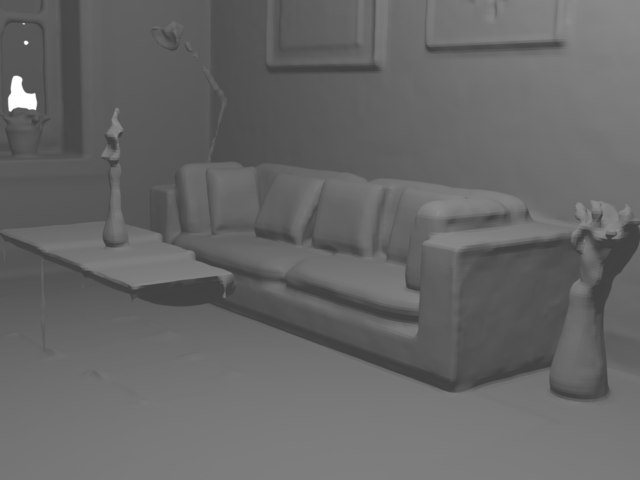}} \\
    
    \makecell{\rotatebox{90}{GT} } &
    \makecell{\includegraphics[width=\sz\linewidth]{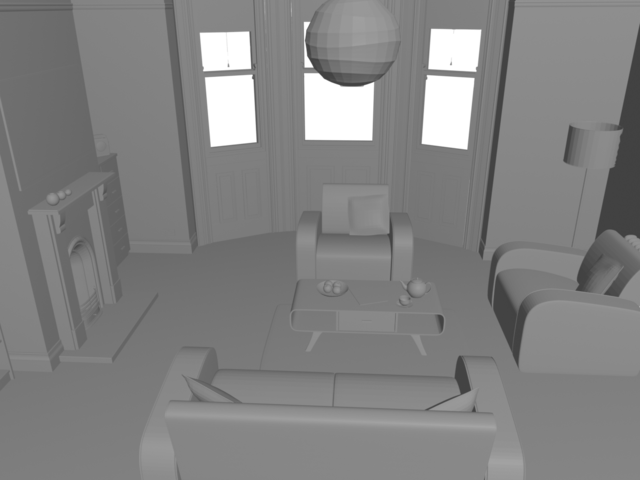}} & 
    \makecell{\includegraphics[width=\sz\linewidth]{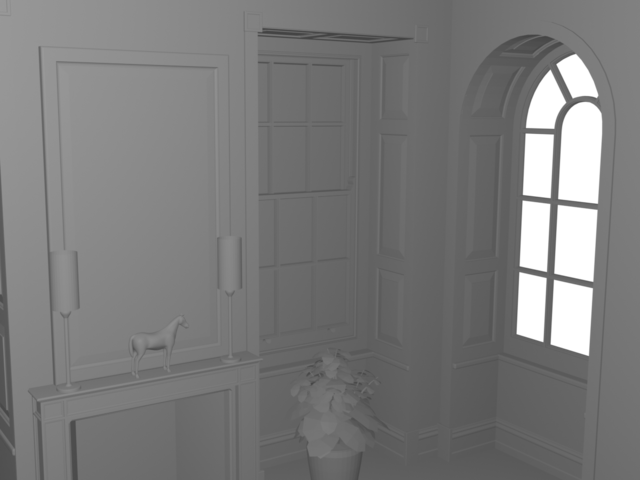}} &
    \makecell{\includegraphics[width=\sz\linewidth]{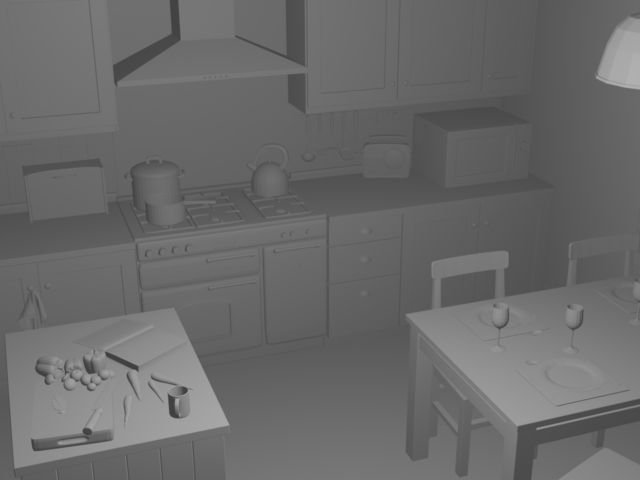}} &
    \makecell{\includegraphics[width=\sz\linewidth]{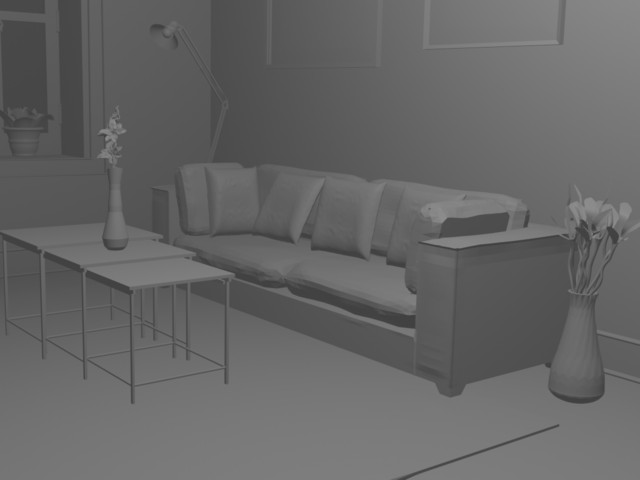}} \\
  \end{tabular}
  \caption{Qualitative comparison on synthetic scenes. GO-Surf produces smoother, and more complete reconstruction without losing tiny details. Both NeuralRGBD and our method have the scene completion ability. However, NeuralRGBD produces much noisier completion results with many artefacts whereas ours is much smoother and looks more natural.
  }
  \label{fig:comparison_synthetic}
\end{figure*}

In this section, we show more qualitative reconstruction results on synthetic scenes. Note that in order to also showcase the scene completion ability of different methods, the results shown here are from unculled meshes. Fig.~\ref{fig:comparison_synthetic} shows the comparison of our method to NeuralRGB-D and other learning-based methods. Overall, our method produces smoother, and more complete reconstruction without losing tiny details. 

\begin{figure*}[tbp]
  \centering
  \setlength{\tabcolsep}{0.5pt}
  \newcommand{\sz}{0.24}  %

  \begin{tabular}{lcccc}

    \makecell{\rotatebox{90}{BundleFusion~\cite{dai2017bundlefusion}}}  &
    \makecell{\includegraphics[width=\sz\linewidth]{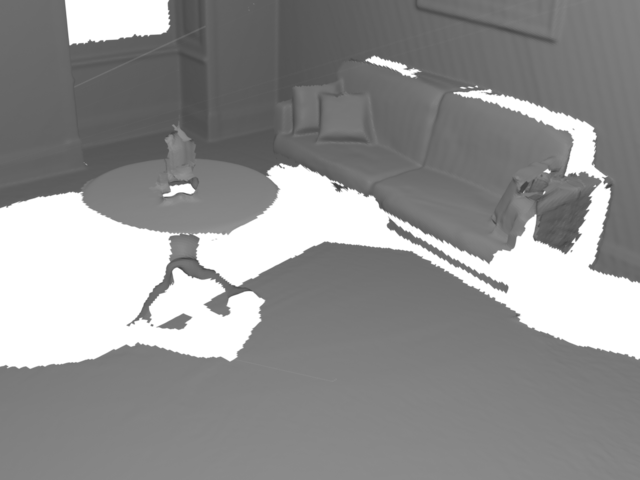}} &
    \makecell{\includegraphics[width=\sz\linewidth]{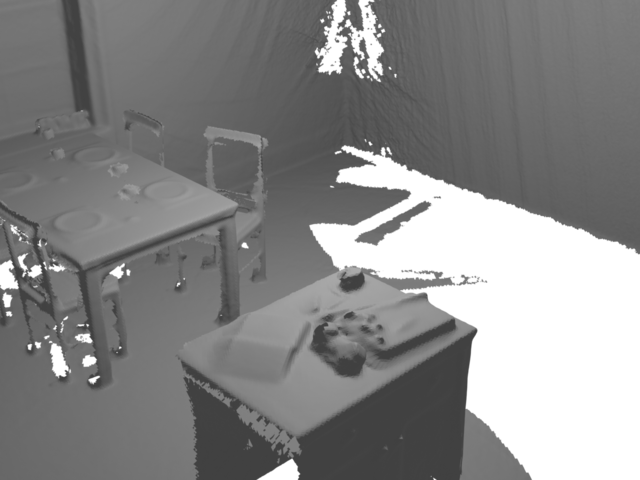}} &
    \makecell{\includegraphics[width=\sz\linewidth]{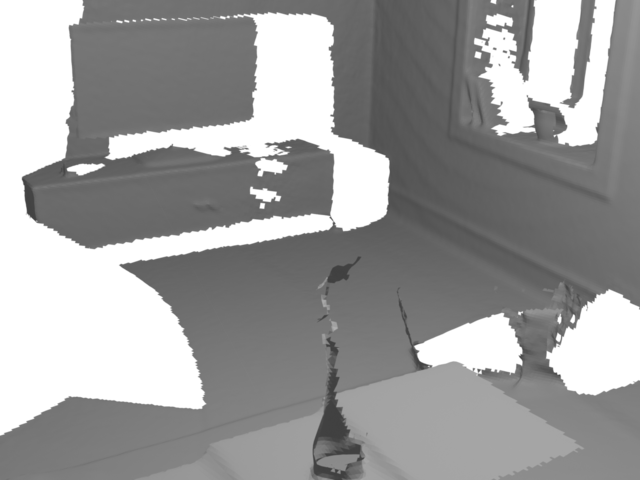}} & 
    \makecell{\includegraphics[width=\sz\linewidth]{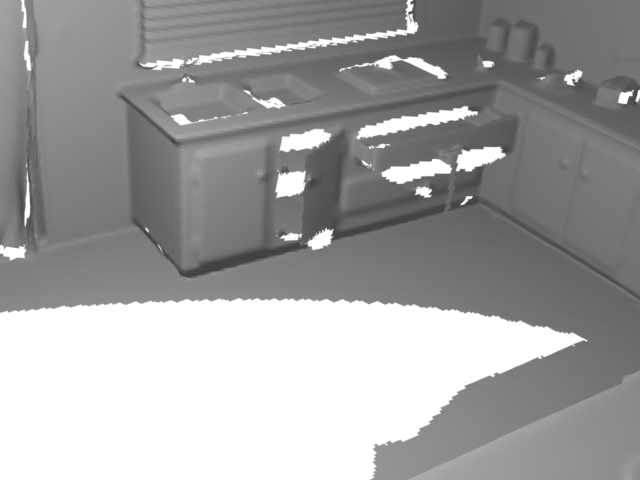}} \\

    \makecell{\rotatebox{90}{NeuralRGBD~\cite{azinovic2022neural}}} &
    \makecell{\includegraphics[width=\sz\linewidth]{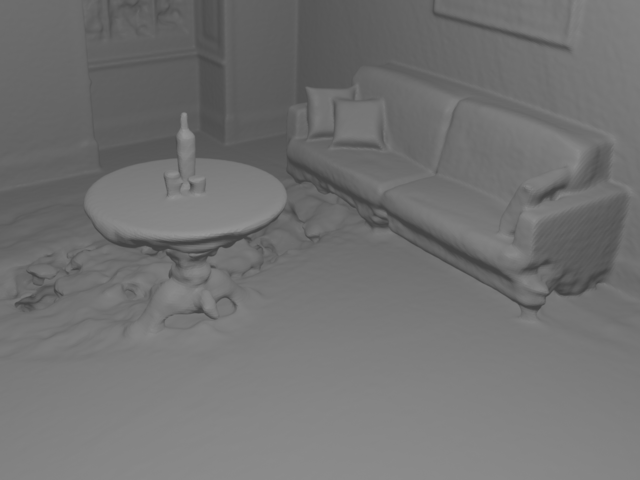}} & 
    \makecell{\includegraphics[width=\sz\linewidth]{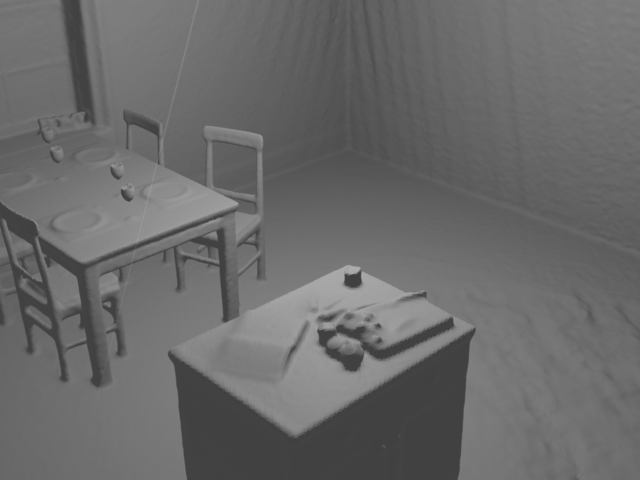}} &
    \makecell{\includegraphics[width=\sz\linewidth]{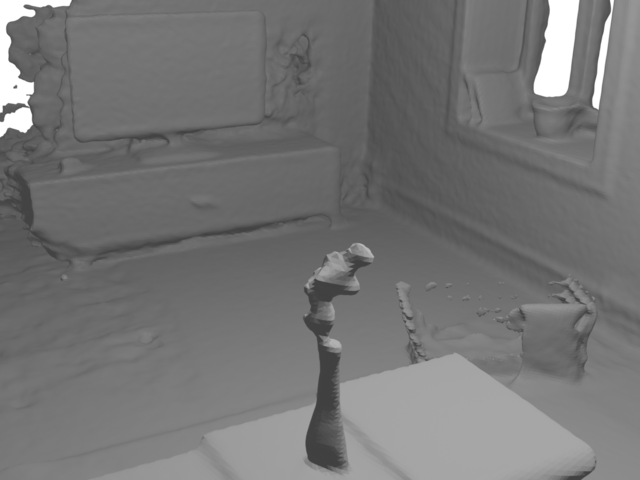}} &
    \makecell{\includegraphics[width=\sz\linewidth]{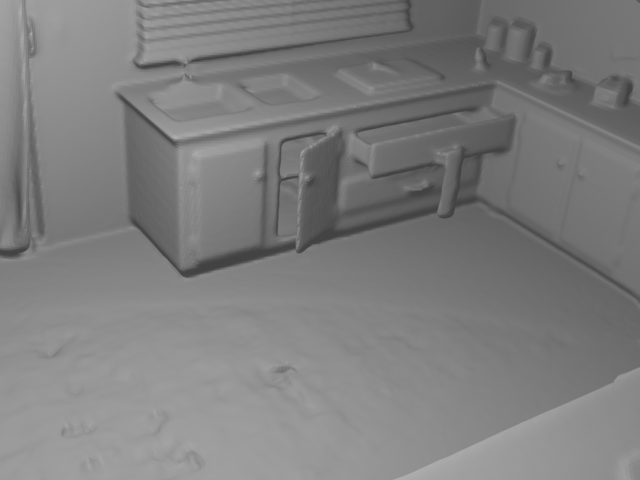}} \\

    \makecell{\rotatebox{90}{GO-Surf (Ours)}} &
    \makecell{\includegraphics[width=\sz\linewidth]{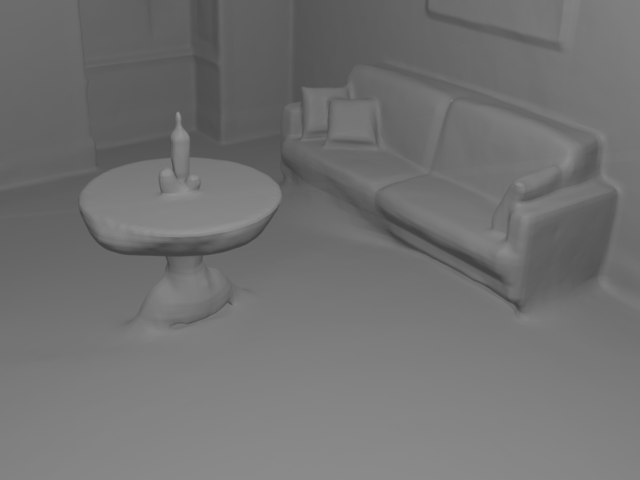}} & 
    \makecell{\includegraphics[width=\sz\linewidth]{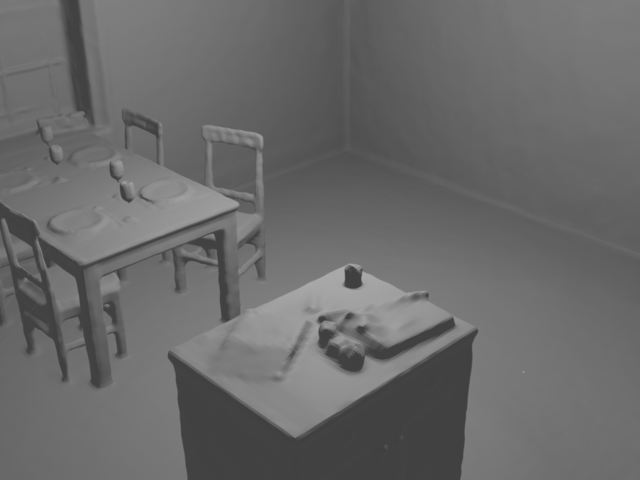}} &
    \makecell{\includegraphics[width=\sz\linewidth]{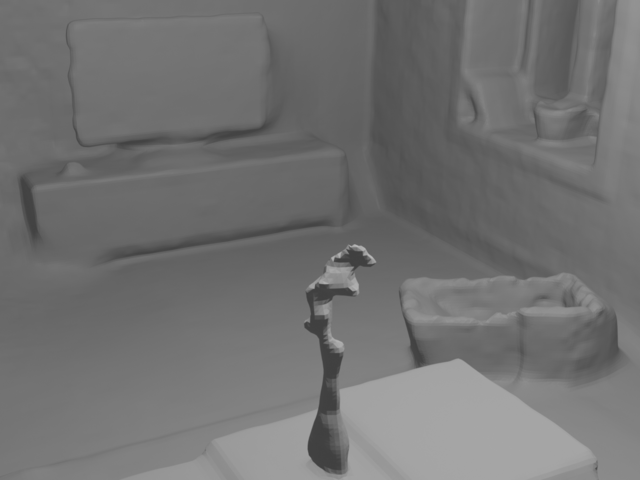}} & \makecell{\includegraphics[width=\sz\linewidth]{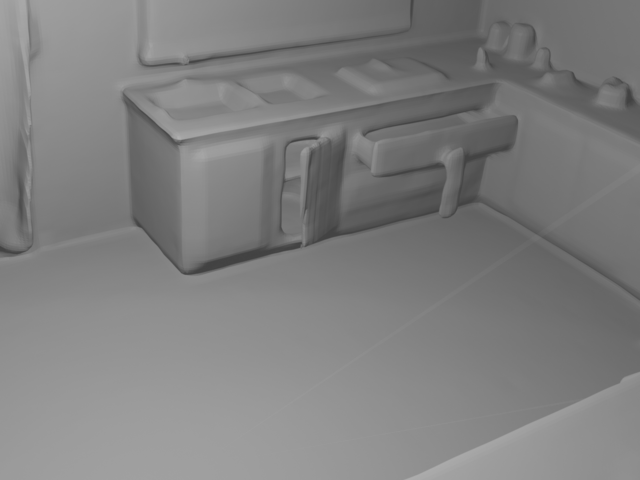}}
    \\
    
    \makecell{\rotatebox{90}{GT} } &
    \makecell{\includegraphics[width=\sz\linewidth]{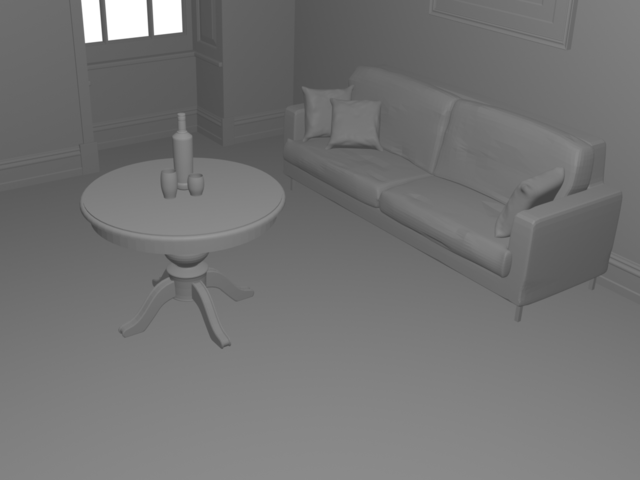}} & 
    \makecell{\includegraphics[width=\sz\linewidth]{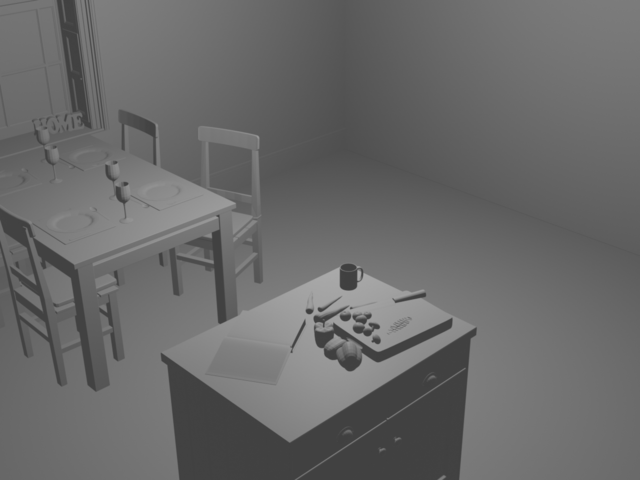}} &
    \makecell{\includegraphics[width=\sz\linewidth]{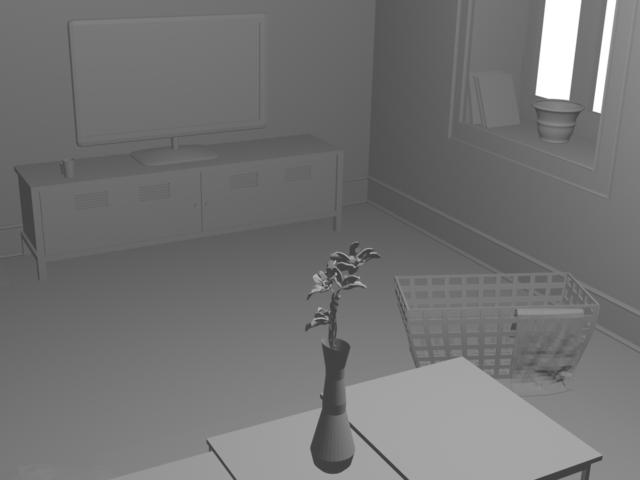}} & \makecell{\includegraphics[width=\sz\linewidth]{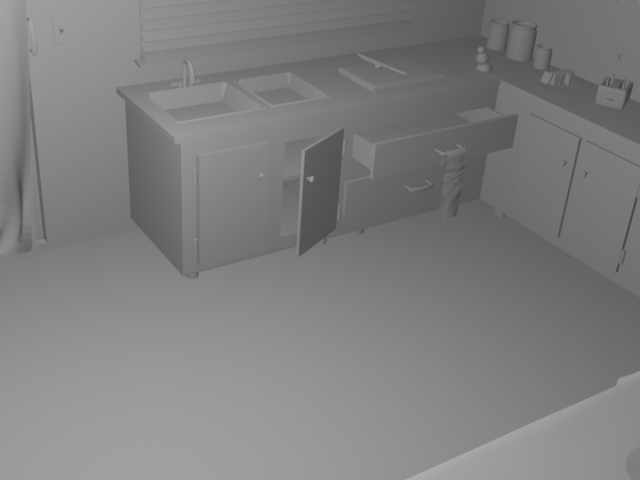}} \\
  \end{tabular}
  \caption{Scene completion comparison. NeuralRGBD produces much noisier completion results with many artefacts whereas ours is much smoother and looks more natural.
  }
  \label{fig:comparison_synthetic_hole_filling}
\end{figure*}

From the first two columns it can be seen that both NeuralRGB-D and our method have the ability to fill in unobserved regions (windows and the top ceiling). However, NeuralRGB-D tends to produce noisier scene completion results with many artefacts whereas ours is much smoother and looks more natural. In Fig~\ref{fig:comparison_synthetic_hole_filling} we provide more results to further showcase our method's advantage over NeuralRGB-D in terms of scene completion.

\begin{table*}[tp!]
    \centering
	\resizebox{0.75\linewidth}{!}{%
    \begin{tabular}{lccccc}
        \toprule
        scene  & scene dim. & voxel dim. & runtime & num params. & model size \\
        \midrule
        complete kitchen & $9.3 \times 10.0 \times 3.3$ & $321 \times 353 \times 129$ & $45 \text{ min}$ & $73.1 \text{ M}$ & $294 \text{ MB}$ \\
        kitchen & $7.0 \times 8.7 \times 3.4$ & $257 \times 321 \times 129$ & $40 \text{ min}$ & $53.2 \text{ M}$ & $214 \text{ MB}$ \\
        breakfast room & $4.1 \times 4.7 \times 3.3$ & $161 \times 193 \times 129$ & $25 \text{ min}$ & $20.1 \text{ M}$ & $81 \text{ MB}$ \\
        morning apartment & $3.5 \times 4.0 \times 2.3$ & $129 \times 161 \times 97$ & $19 \text{ min}$ & $10.1 \text{ M}$ & $41 \text{ MB}$ \\
        grey white room & $5.9 \times 4.4 \times 3.1$ & $225 \times 161 \times 129$ & $24 \text{ min}$ & $23.4 \text{ M}$ & $94 \text{ MB}$ \\
        white room & $5.6 \times 7.9 \times 3.8$ & $193 \times 289 \times 161$ & $39 \text{ min}$ & $44.9 \text{ M}$ & $181 \text{ MB}$ \\
        staircase & $6.8 \times 6.5 \times 3.7$ & $257 \times 225 \times 161$ & $39 \text{ min}$ & $46.6 \text{ M}$ & $188 \text{ MB}$ \\
        green room & $8.0 \times 4.7 \times 3.1$ & $289 \times 193 \times 129$ & $30 \text{ min}$ & $36.0 \text{ M}$ & $145 \text{ MB}$ \\
        thin geometry & $3.4 \times 3.6 \times 1.2$ & $129 \times 129 \times 65$ & $15 \text{ min}$ & $5.4 \text{ M}$ & $28 \text{ MB}$ \\
        ICL living room & $5.3 \times 2.9 \times 5.4$ & $193 \times 193 \times 129$ & $24 \text{ min}$ & $24.1 \text{ M}$ & $97 \text{ MB}$ \\
        \midrule
        scene 0000 & $8.8 \times 9.1 \times 3.4$ & $321 \times 321 \times 129$ & $44 \text{ min}$ & $66.5$ M & $268 \text{ MB}$ \\
        scene 0002 & $4.4 \times 6.0 \times 3.5$ & $161 \times 225 \times 129$ & $28 \text{ min}$ & $23.4$ M & $113 \text{ MB}$ \\
        scene 0005 & $5.8 \times 5.3 \times 2.9$ & $225 \times 193 \times 125$ & $33 \text{ min}$ & $27.1$ M & $113 \text{ MB}$ \\
        scene 0012 & $5.8 \times 5.7 \times 2.9$ & $225 \times 225 \times 129$ & $31 \text{ min}$ & $32.7$ M & $132 \text{ MB}$ \\
        scene 0024 & $7.6 \times 8.4 \times 2.9$ & $289 \times 289 \times 129$ & $40 \text{ min}$ & $53.9$ M & $217 \text{ MB}$ \\
        scene 0050 & $5.9 \times 4.5 \times 3.1$ & $225 \times 161 \times 129$ & $26 \text{ min}$ & $23.4$ M & $94 \text{ MB}$ \\
        \bottomrule
    \end{tabular}
    }
    \caption{
    Runtime and memory requirement on synthetic and ScanNet~\cite{dai2017scannet} scenes. We list scene dimension ($\text{m}^3$), voxel dimension, runtime (min), number of model parameters and model size (MB). Our method requires much less runtime than NeuralRGBD~\cite{azinovic2022neural} ($15-45 \text{ min}$ vs $15-25 \text{ hours}$) at the cost of larger model size (tens to hundreds of MB vs several MB).}
\label{tab:detailed_performance}
\end{table*}

\section{Runtime and Memory Analysis}

In Tab.~\ref{tab:detailed_performance} we provide detailed breakdown of of the runtime and memory usage of our method on all of the scenes that we had run experiments on. Our method requires $15$ to $45$ minutes to converge which is approximately $\times 60$ faster than NeuralRGB-D. However our model size is much larger (tens to hundreds million of parameters vs several million parameters) and scales rapidly as the scene gets larger. This is This is an inherent problem of voxel-like architectures as the number of voxels scales cubically with the scene dimension. Voxel hashing or octree-based sparsification could be adopted to significantly reduce the memory footprint of our system and we intend to explore this direction in future work. 

\section{Sequential Mapping Experiments}
GO-Surf's fast runtime also enables sequential/online mapping at interactive framerate with slightly reduced resolution. In this section we run GO-Surf in a sequential fashion using $3$-level feature grid with voxel sizes of $6\text{cm}$, $24\text{cm}$ and $96\text{cm}$ on ScanNet scene0000 and compare against NICE-SLAM~\cite{Zhu2022CVPR_niceslam} running on the same scene with ground truth camera poses.

With increased voxel size GO-Surf runs in near real-time at $15\text{Hz}$ while NICE-SLAM runs much slower below $0.5\text{Hz}$. Fig~\ref{fig:mapping_result} shows the comparison of reconstruction results. Ours is much smoother and has has less artifacts.

{\small
\bibliographystyle{ieee_fullname}
\bibliography{supp_material}
}